\documentclass[pdflatex,sn-apa]{sn-jnl}% APA Reference Style
\usepackage{graphicx}%
\usepackage{subcaption}%
\usepackage{multirow}%
\usepackage{amsmath,amssymb,amsfonts}%
\usepackage{amsthm}%
\usepackage{mathrsfs}%
\usepackage[title]{appendix}%
\usepackage[dvipsnames,table]{xcolor}%
\usepackage{textcomp}%
\usepackage{manyfoot}%
\usepackage{booktabs}%
\usepackage{algorithm}%
\usepackage{algorithmicx}%
\usepackage{algpseudocode}%
\usepackage{listings}%
\usepackage{anyfontsize}%
\usepackage{float}
\usepackage{longtable}%
\usepackage{tikz}%
\usetikzlibrary{shapes.geometric, fit, calc}

\newsavebox{\resizedtablebox}

\colorlet{algorithm}{black!15}
\colorlet{hindsightrow}{red!15}
\colorlet{testhindsightrow}{red!32}

\begin{document}

\title[TIGER]{TIGER: Time-Series Classification with In-Context-Learning Gated Ensemble of Representations}

\author[]{\fnm{Johann} \sur{Faouzi}}\email{johann.faouzi@ensai.fr}

\affil[]{Univ Rennes, Ensai, CNRS, CREST - UMR 9194, F-35000 Rennes, France}

\abstract{
    A representation family is a distinct way of extracting features from time series.
    Ensemble algorithms that combine several representation families remain the most accurate approach to time series classification.
    Current state-of-the-art ensembles, most notably HIVE-COTE~2.0, pair a bespoke classification algorithm with each representation family and combine their predictions using a fixed, non-adaptive rule.
    We present TIGER (Time-series classification with In-context-learning Gated Ensemble of Representations), which instead applies the same small portfolio of three general-purpose classifiers (Ridge, Extra Trees, and Naive Bayes) to four representations from four distinct families, stacking the resulting twelve base learners' predictions into a meta-feature matrix.
    The final prediction is produced by an adaptive meta-classification rule that chooses, independently for each data set, between a weighted hard majority vote and TabICLv2, a pretrained tabular foundation model used in-context as a meta-classifier, based on the mean number of training samples available per class.
    On a 142-data-set benchmark drawn from the UCR time series classification archive, TIGER obtains the best mean accuracy, balanced accuracy, and F1-score among six compared algorithms, including HIVE-COTE~2.0, and significantly outperforms each of the other five individually.
    TIGER's adaptive rule also meaningfully outperforms either of its two constituent meta-classification methods used alone, and its single hyperparameter, tuned using only a twenty-data-set development subset, is shown to generalize to the full evaluation benchmark.
    We further characterize TIGER's design through an extensive set of ablation experiments and report the design alternatives that we investigated and ultimately discarded.
}

\keywords{time series classification, time series, classification, machine learning, supervised learning, ensemble methods, tabular foundation models}

\maketitle

\section{Introduction}\label{sec1}

We consider time series as an ordered sequence of real-valued observations (often called univariate time series) sampled at equal frequency.
Such data are very common in many real-world applications given their nature.
With the increasing availability of sensors as well as the development of \emph{Internet of things} devices, more time series data and applications become available.
Numerous varied fields are affected, from climate science with weather forecasting \citep{agrawal2012application} to geology with earthquake prediction \citep{amei2012time} to econometrics and finance \citep{perron1989great}.

Time series classification (TSC) is a subfield of machine learning, and more specifically supervised learning, aiming at predicting a discrete label for a time series based on a set of labeled training examples.
Time series classification has been employed in various fields, such as healthcare, speech recognition, and human activity recognition, and for various applications, such as diagnosis and monitoring \citep{middlehurstBakeReduxReview2024}.
The field has been driven for over a decade by large, standardized benchmarks of publicly available data sets, most notably the University of California, Riverside (UCR) time series classification archive \citep{dauUCRTimeSeries2019e}, which has enabled systematic, reproducible comparisons between algorithms.
Many families of algorithms have been developed for time series classification, most of them consisting of a transformation step, which extracts a new representation of the time series more amenable to discrimination between classes, followed by a standard classification algorithm applied to this new representation.
Distance-based, interval-based, shapelet-based, dictionary-based, and convolution-based transformations have all been investigated, each capturing different, partially complementary, notions of similarity between time series.

Since no single family of transformations is uniformly the most discriminative representation across every data set, the current state of the art in time series classification is achieved by ensemble algorithms that combine several families of transformations.
The most notable one is HIVE-COTE~2.0 (HC2) \citep{middlehurstHIVECOTE20New2021a}, which is the fourth iteration of an ensemble design that pairs a bespoke classification algorithm with each transformation family and aggregates their predictions using a probability-weighted combination rule.
While highly accurate, this design has two notable limitations.
First, every one of its constituent classification algorithms is specific to its representation family, hand-picked and separately maintained, which makes both extending the ensemble to new families and reasoning about the aggregation step complex.
Using algorithms with separable transformation and classification steps allows for easily investigating several classification algorithms for each transformation.
Second, its combination rule, the Cross-validation Accuracy Weighted Probabilistic Ensemble (CAWPE) \citep{linesTimeSeriesClassification2018a}, is a fixed, non-learned rule that does not adapt to how much labeled data is actually available for a given data set.
Meanwhile, tabular foundation models, and in particular models based on in-context learning such as TabPFN \citep{hollmannTabPFNTransformerSolves2023} and TabICL \citep{quTabICLTabularFoundation2025}, have recently emerged as a new paradigm for tabular classification, predicting directly from a training set given as context at inference time, without any per-data-set gradient-based training.
Concurrently with our own work, a few studies have combined such in-context-learning models with time-series-specific feature extraction, applying them directly, and unconditionally, as the sole classifier on top of a single family of hand-crafted features \citep{orourkeRocketPFNAccurateTime2026, cuppersInContextTimeSeries2026} or training an analogous model end-to-end on raw time series \citep{kukenTimEEEndtoendTime2026}.
To the best of our knowledge, such models have not yet been investigated as an adaptive meta-classification layer sitting on top of a multifamily stacking ensemble.

We present a novel time series classification algorithm that we call TIGER, which stands for \textbf{T}ime-series classification with \textbf{I}n-context-learning \textbf{G}ated \textbf{E}nsemble of \textbf{R}epresentations.
TIGER performs four transformations from four distinct families (convolution, interval, shapelet, and dictionary).
Unlike HIVE-COTE~2.0, TIGER applies the same small portfolio of three general-purpose classifiers to every family, producing twelve base learners in total.
The predictions of these twelve base learners are stacked into a meta-feature matrix.
The predictions are out-of-fold (OOF) using $k$-fold cross-validation (CV) during training, and held-out during inference.
The final prediction is produced from this meta-feature matrix by one of two meta-classification methods, chosen automatically for each data set based on the mean number of training samples available per class: a weighted hard majority vote when few samples per class are available, or a tabular foundation model, used here as a meta-classifier fit directly on the meta-feature matrix, when more samples per class are available.
We evaluate TIGER on a 142-data-set benchmark drawn from the UCR archive, matching the selection used in a recent large-scale review of the field \citep{middlehurstBakeReduxReview2024}, and compare it to state-of-the-art time series classification algorithms.
Our results show that TIGER is competitive with, and on the full 142-data-set evaluation benchmark significantly outperforms, existing state-of-the-art algorithms, and that the adaptive meta-classification rule meaningfully outperforms either of its two constituent methods used alone.

The remainder of this paper is structured as follows.
Section~\ref{sec2} provides relevant information about existing time series classification algorithms, as well as ensemble methods and tabular foundation models.
In Section~\ref{sec3}, we describe the architecture of TIGER, the reasoning behind its architecture, and which elements were optimized.
Section~\ref{sec4} presents the algorithms that we compare TIGER to, as well as the data sets and metrics used to perform these comparisons.
We provide the results of all the experiments in Section~\ref{sec5}, discuss design alternatives that we tried and discarded in Section~\ref{sec6}, and conclude in Section~\ref{sec7}.

\section{Related work}\label{sec2}

In this section, we provide information about time series classification algorithms, ensemble methods, and tabular foundation models.
We categorize the transformation-based algorithms based on their approach and present the more relevant ones to our study, as providing a systematic review is out of scope of this study.
For a recent, comprehensive review, we refer the readers to \citep{middlehurstBakeReduxReview2024}.

\subsection{Interval-based approaches}

Interval-based methods generate intervals, extract the subseries corresponding to these intervals, and compute statistics from these subseries.
The Time Series Forest (TSF) algorithm \citep{dengTimeSeriesForest2013} computes three statistics for each subseries (the mean, the standard deviation, and the slope) and applies a random forest \citep{breimanRandomForests2001a} on top of this transformation.
The Canonical Interval Forest (CIF) algorithm \citep{middlehurstCanonicalIntervalForest2020} extends TSF with more features being extracted, and the Diverse Representation Canonical Interval Forest (DrCIF) algorithm \citep{middlehurstHIVECOTE20New2021a} further extends CIF with two additional time series representations: periodograms and first-order differences.
The QUANT algorithm \citep{dempsterQuantMinimalistInterval2024} instead uses fixed dyadic intervals and extracts quantiles rather than the aforementioned features, but includes four time series representations: raw time series, first- and second-order differences, and Fourier coefficients.
The classification algorithm used in conjunction is extremely randomized trees \citep{geurtsExtremelyRandomizedTrees2006a}.
PULSAR \citep{cabelloPULSARAdvancingIntervalBased2025} instead computes simple local statistics from subseries of varied length and dilation over several representations of the series, hierarchically pools these statistics across multiple partition depths using randomly chosen pooling operators, and applies supervised feature selection to keep only the most discriminative aggregates before training an ensemble classifier.
PULSAR closes most of the accuracy gap to HC2, while being more interpretable.

\subsection{Shapelet-based approaches}

Shapelets are subseries extracted from the training time series and used to differentiate time series, most commonly using the minimum Euclidean distance between the shapelet and all the subseries, of the same length as the shapelet, extracted from a time series.
Shapelets were first investigated in \citep{yeTimeSeriesShapelets2011}, and the Shapelet Transform Classifier (STC) \citep{hillsClassificationTimeSeries2014a} later investigated all the possible shapelets from the training set before selecting the most discriminative ones, followed by an ensemble of classifiers.
The Random Dilated Shapelet Transform (RDST) algorithm \citep{guillaumeRandomDilatedShapelet2022} adds two novel elements to existing shapelet-based approaches: the use of dilation (the entries of a shapelet are not necessarily consecutive entries of a training time series, with every other $d$ entries being used) and the addition of two new features computed for each shapelet, the position of the minimum distance and the number of occurrences of the shapelet.

\subsection{Dictionary-based approaches}

Dictionary-based methods, similarly to shapelet-based approaches, extract subseries from a time series, but instead of computing a distance, each subseries is turned into a short sequence of discrete symbols (a word), and the frequencies of all the words extracted from a time series are computed to obtain its new representation.
There exist two main symbolic representations: the Symbolic Aggregate approXimation (SAX) \citep{linExperiencingSAXNovel2007a}, in the time domain, and the Symbolic Fourier Approximation (SFA) \citep{schaferSFASymbolicFourier2012}, in the frequency domain.
The Bag-of-SFA-Symbols (BOSS) algorithm \citep{schaferBOSSConcernedTime2015} extracts subseries using overlapping windows, transforms each subseries into a word using SFA, and computes the word frequencies, in conjunction with a nearest-neighbor classifier using a dedicated distance.
The Word Extraction for Time Series Classification (WEASEL) algorithm \citep{schaferFastAccurateTime2017a} involves several changes compared to BOSS, notably supervised selection of the Fourier coefficients and of the SFA quantization, the use of bigrams and multiple window lengths, and a logistic regression classifier fit on a supervised subset of words.
WEASEL was later refined into WEASEL~2.0 \citep{schaferWEASEL20Random2023}, adding a novel dilation mapping, using less supervised selection methods in SFA, and replacing logistic regression with a Ridge classifier.

\subsection{Convolution-based approaches}

Convolution is a mathematical operation with applications in numerous fields such as signal and image processing.
In machine learning, a convolution is characterized by a kernel, with its weights, bias, padding, and dilation.
Instead of learning the kernels for a specific classification task, it has been proposed to use random kernels for time series classification, a strategy introduced with the Random Convolutional Kernel Transform (ROCKET) algorithm \citep{dempsterROCKETExceptionallyFast2020}, which generates numerous random kernels, applies the convolution for each of them, extracts aggregate features (the maximum value and the proportion of positive values), and trains a Ridge classifier on these extracted features.
ROCKET was later extended by MiniROCKET \citep{dempsterMiniRocketVeryFast2021a}, which involves much less randomness in the generation of the kernels and is faster with very similar performance, and by MultiROCKET \citep{tanMultiRocketMultiplePooling2022}, which adopts the improvements of MiniROCKET, extracts three additional features from the convolutional output, and applies half of the convolutions to the first-order difference of the time series.
A model combining both dictionary- and convolution-based approaches, Hydra \citep{dempsterHydraCompetingConvolutional2023}, aggregates kernels into groups, records the best matching kernel among each group, and computes their frequencies.
Concatenating Hydra's features with MultiROCKET's own, a combination referred to as MultiROCKET-Hydra (MR-Hydra), was the top-ranked convolution-based classifier in a recent, large-scale review of the field \citep{middlehurstBakeReduxReview2024}.
More recently, the Kernel Grouping for Multiple Transformations and Pooling operators (KG-MTP) algorithm \citep{wangKernelGroupingTime2025} groups convolutional kernels and combines several transformations and pooling operators to extract a richer convolution-based feature set.

\subsection{Ensemble approaches}\label{sec2sub6}

Given that no single family of transformations described above is uniformly the most discriminative representation across every data set, ensemble algorithms combining several such families have consistently achieved state-of-the-art predictive performance on the UCR archive.

\begin{figure}[tbp]
    \centering
    \resizebox{0.92\linewidth}{!}{
    \begin{tikzpicture}[
        node distance=1.4cm,
        every node/.style={font=\small},
    ]
        \tikzstyle{io} = [
            rectangle,
            rounded corners,
            minimum width=2.6cm,
            minimum height=1.1cm,
            text width=2.6cm,
            text centered,
            align=center,
            draw=black,
        ]

        \tikzstyle{arrow} = [thick, ->, >=stealth, rounded corners=6pt]
        \tikzstyle{trunk} = [thick, rounded corners=6pt]
        \tikzstyle{stagelabel} = [font=\bfseries\small, text=black!60, text width=3.8cm, align=center]
        \tikzstyle{stagebox} = [draw=black!55, dashed, rounded corners=6pt, thick]

        \coordinate (labelx) at (-10.6, 0);
        \coordinate (leftEdge) at (-8.3, 0);
        \coordinate (rightEdge) at (8.3, 0);

        \node (input) [io, fill=Apricot!30] at (0, 0) {Time series};

        \node (stc) [io, fill=Bittersweet!30, minimum width=3.9cm, text width=3.6cm] at (-6.3, -2.6) {STC\\ \emph{(shapelet transform + Rotation Forest)}};
        \node (tde) [io, fill=Bittersweet!30, minimum width=3.9cm, text width=3.6cm] at (-2.1, -2.6) {TDE\\ \emph{(dictionary ensemble)}};
        \node (arsenal) [io, fill=Bittersweet!30, minimum width=3.9cm, text width=3.6cm] at (2.1, -2.6) {Arsenal\\ \emph{(ROCKET + Ridge ensemble)}};
        \node (drcif) [io, fill=Bittersweet!30, minimum width=3.9cm, text width=3.6cm] at (6.3, -2.6) {DrCIF\\ \emph{(interval forest)}};

        \node (cawpe) [io, fill=Yellow!30, minimum width=5.8cm, text width=5.8cm] at (0, -5.4) {CAWPE\\ \emph{(accuracy-weighted soft-majority voting)}};

        \node (output) [io, fill=Cyan!30] at (0, -7.6) {Final prediction};

        \node[stagelabel] at (labelx |- stc) {Family-specific\\transform + classifier};
        \node[stagelabel] at (labelx |- cawpe) {Combination\\(CAWPE)};

        \coordinate (fam-wl) at (leftEdge |- stc);
        \coordinate (fam-wr) at (rightEdge |- stc);
        \coordinate (comb-wl) at (leftEdge |- cawpe);
        \coordinate (comb-wr) at (rightEdge |- cawpe);

        \draw [thick] (input.south) -- ++(0,-0.5) coordinate (input-branch);
        \draw [arrow] (input-branch) -| (stc.north);
        \draw [arrow] (input-branch) -| (tde.north);
        \draw [arrow] (input-branch) -| (arsenal.north);
        \draw [arrow] (input-branch) -| (drcif.north);

        \coordinate (mergeY) at ($(tde.south)+(0,-1.0)$);
        \coordinate (mainTrunk) at (cawpe |- mergeY);
        \foreach \n in {stc,tde,arsenal,drcif} {
            \draw [trunk] (\n.south) |- (mainTrunk);
        }
        \draw [arrow] (mainTrunk) -- (cawpe);

        \draw [arrow] (cawpe) -- (output);

        \node[stagebox, fit=(stc)(tde)(arsenal)(drcif)(fam-wl)(fam-wr), inner sep=6pt] {};
        \node[stagebox, fit=(cawpe)(comb-wl)(comb-wr), inner sep=6pt] {};
    \end{tikzpicture}}

    \caption{
        Overview of HIVE-COTE~2.0's architecture.
    }
    \label{fig:hc2}
\end{figure}
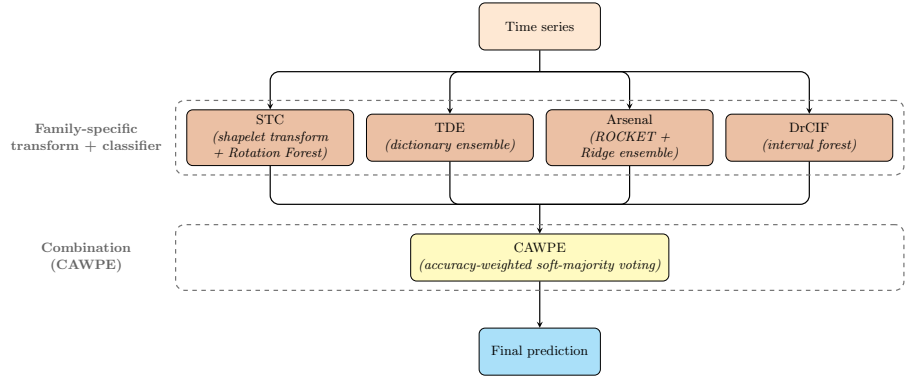

The first such hybrid ensemble, the Collective of Transformation-based Ensembles (COTE, later renamed Flat-COTE) \citep{bagnallTimeSeriesClassification2015}, pools 35 individual classifiers built on four representations (shapelets, elastic distances computed directly on the raw series, autocorrelation, and the power spectrum) into a single flat ensemble, combined by a weighted vote using each classifier's own cross-validated training accuracy.
The Hierarchical Vote Collective of Transformation-based Ensembles (HIVE-COTE, subsequently referred to as HIVE-COTE$\alpha$) \citep{linesTimeSeriesClassification2018a} restructures this idea into a two-level hierarchy: five representation-specific ensembles (the Elastic Ensemble for distance-based features, Time Series Forest for interval-based features, BOSS for dictionary-based features, a shapelet transform ensemble, and the Random Interval Spectral Ensemble for spectral features) are each trained independently, and their predicted class probabilities are combined using the Cross-validation Accuracy Weighted Probabilistic Ensemble (CAWPE) rule: each constituent's probabilities are averaged, weighted by that constituent's own cross-validated training accuracy raised to a fixed power (four, empirically found to work best).
HIVE-COTE version 1.0 \citep{bagnallUsagePerformance2020} simplifies this original design for usability and speed rather than accuracy: the computationally expensive Elastic Ensemble is dropped entirely, the dictionary and shapelet ensembles are each replaced with a faster, more configurable variant, and the whole ensemble is made contractable, so that it can be given a maximum training time budget and will build as large an ensemble as that budget allows.
HIVE-COTE~2.0 \citep{middlehurstHIVECOTE20New2021a} further updates all four constituent modules to more recent, more accurate representatives of their respective families (the Shapelet Transform Classifier (STC), the Temporal Dictionary Ensemble (TDE), the Diverse Representation Canonical Interval Forest (DrCIF), and the Arsenal, a new convolution-based ensemble of ROCKET classifiers) while keeping the same CAWPE combination rule.
It also estimates each module's own accuracy (the CAWPE weight) via an adapted form of out-of-bag error rather than a separate cross-validation, since every module is itself a bagging ensemble.
HIVE-COTE~2.0 remains one of the most accurate time series classification algorithms as of a recent, large-scale review of the field \citep{middlehurstBakeReduxReview2024}.
Figure~\ref{fig:hc2} summarizes its architecture.

Other hybrid designs combine representations differently from the HIVE-COTE lineage.
TS-CHIEF \citep{shifazTSCHIEF2020} is a homogeneous ensemble of decision trees in which each node considers randomly initialized splitting criteria drawn from several representations directly (dictionary-based splits inspired by BOSS, distance-based splits inspired by the Elastic Ensemble, and interval-based splits inspired by RISE) rather than combining separately trained, representation-specific classifiers.
Proximity Forest \citep{lucasProximityForest2019} is a related, purely distance-based tree ensemble: at each node, one of eleven elastic distance measures (the same repertoire used by the Elastic Ensemble) is selected together with a small set of randomly chosen class exemplars, and each series is routed toward whichever exemplar it is closest to under that distance.
RIST \citep{middlehurstExtractingFeatures2023} departs from ensembling altogether: it concatenates interval-based features (from DrCIF) and shapelet-based features (from RDST), each computed on the raw series and on several derived representations, into a single feature vector classified by one Extra Trees model, trading the accuracy benefit of ensembling for a simpler, single-model pipeline.

\subsection{Foundation models and in-context learning}

A foundation model is a single model, pretrained once on a large corpus of synthetic or real data, that can then be applied to a new, unseen task or data set with little or no further gradient-based training.
A growing family of such models uses in-context learning: the new data set's own training rows are given to the model as context at inference time, and predictions for new rows are produced directly by a single forward pass, without any per-data-set optimization.

\subsubsection{Tabular foundation models}

TabPFN \citep{hollmannTabPFNTransformerSolves2023} introduced in-context learning for tabular classification, using a transformer pretrained on synthetic data sets generated from a prior over structural causal models, and was originally restricted to small data sets.
A second version, TabPFNv2 \citep{hollmannAccuratePredictionsSmall2025}, improved accuracy and scaled to larger data sets, and a subsequent update, TabPFN-2.5 \citep{grinsztajnTabPFN25AdvancingState2025}, extended this further still.
TabICL \citep{quTabICLTabularFoundation2025} scaled this same approach to substantially larger data sets, and its successor, TabICLv2 \citep{quTabICLv2BetterFaster2026}, further improves accuracy, speed, and scalability, while remaining fully open.

\subsubsection{General-purpose time series foundation models}

Separately, a number of foundation models have been developed directly for time series data, most of them primarily targeting forecasting.
MOMENT \citep{goswamiMOMENTFamilyOpen2024} is a family of transformer encoders pretrained by masked time series modeling on a large, heterogeneous public corpus (\emph{the Time Series Pile}), and is evaluated, among other tasks, on classification in low-supervision regimes.
We refer the reader to \citep{kottapalliFoundationModelsTime2025} for a comprehensive, up-to-date survey of this broader landscape.

\subsubsection{Time series classification foundation models}

A more recent line of work develops foundation models specifically for time series classification, rather than treating it as one of several downstream tasks of a forecasting-oriented model.
CauKer \citep{xieCauKerClassificationTime2025} is a synthetic-data generator for pretraining such models, combining Gaussian process kernel composition with structural causal models to produce diverse, labeled synthetic time series at scale, and is shown to yield predictable scaling behavior when used to pretrain classification foundation models.
Mantis \citep{feofanovMantisLightweightFoundation2025} is one such model: a lightweight transformer pretrained with self-supervised contrastive learning on CauKer-generated synthetic data, using a multi-view tokenizer that combines instance-normalized, differenced, and statistical patch encodings.
UniShape \citep{liuUniShapeUnifiedShapeAware2026} instead pretrains a shape-aware adapter that aggregates multiscale discriminative subsequences into class tokens, together with a prototype-based objective, explicitly targeting the shapelet-like, locally discriminative patterns that generic forecasting-oriented pretraining does not emphasize.

\subsubsection{In-context learning directly on time-series-derived features}\label{sec2icl}

A small but growing body of concurrent work combines time-series-specific feature extraction with a pretrained tabular in-context-learning model, rather than pretraining a dedicated time series foundation model from scratch.
RocketPFN \citep{orourkeRocketPFNAccurateTime2026} applies ROCKET's random convolutional features and feeds them directly to TabPFN for zero-shot classification, matching HC2 on the UCR archive with no per-data-set training.
MASHT \citep{cuppersInContextTimeSeries2026} follows the same recipe with MultiROCKET and Hydra features fed to a pretrained tabular in-context-learning model.
TS2TabPFN \citep{dacostamerlinTS2TabPFNTimeSeries2026} follows a similar recipe with a choice of catch22, tsfresh, or MultiROCKET features fed to TabPFN~2.5, for both classification and extrinsic regression.
TimEE \citep{kukenTimEEEndtoendTime2026} goes further and removes the feature-extraction step entirely, training a single transformer end-to-end, on purely synthetic data, to map a labeled time series context directly to class predictions via in-context learning.

\section{Methodology}\label{sec3}

In this section, we detail the proposed ensemble classification algorithm and specify which hyperparameters were optimized and the values for hyperparameters that were not optimized.

\subsection{Time series classification algorithm}\label{sec3sub1}

\begin{figure}[tbp]
    \centering
    \resizebox{0.95\linewidth}{!}{
    \begin{tikzpicture}[
        node distance=1.4cm,
        every node/.style={font=\small},
    ]
        \tikzstyle{io} = [
            rectangle,
            rounded corners,
            minimum width=2.6cm,
            minimum height=1.1cm,
            text width=2.6cm,
            text centered,
            align=center,
            draw=black,
        ]
        \tikzstyle{classifierbox} = [
            rectangle,
            rounded corners=2pt,
            minimum height=0.85cm,
            text width=3.0cm,
            text centered,
            align=center,
            draw=black,
            font=\scriptsize,
            rotate=90,
        ]
        \tikzstyle{decision} = [
            diamond,
            aspect=2.2,
            text width=2.8cm,
            text centered,
            align=center,
            draw=black,
            fill=Dandelion!30,
        ]
        \tikzstyle{arrow} = [thick, ->, >=stealth, rounded corners=6pt]
        \tikzstyle{trunk} = [thick, rounded corners=6pt]
        \tikzstyle{stagelabel} = [font=\bfseries\small, text=black!60, text width=2.4cm, align=center]
        \tikzstyle{stagebox} = [draw=black!55, dashed, rounded corners=6pt, thick]

        \node[stagelabel] at (-9.9, -2.8) {Feature\\extraction};

        \node (input) [io, fill=Apricot!30] at (0, 0) {Time series};

        \node (conv) [io, fill=Aquamarine!30, minimum width=3.9cm, text width=3.6cm] at (-6.3, -2.8) {KG-MTP\\ \emph{(convolution features)}};
        \node (interval) [io, fill=Aquamarine!30, minimum width=3.9cm, text width=3.6cm] at (-2.1, -2.8) {QUANT\\ \emph{(interval features)}};
        \node (shapelet) [io, fill=Aquamarine!30, minimum width=3.9cm, text width=3.6cm] at (2.1, -2.8) {RDST\\ \emph{(shapelet features)}};
        \node (dictionary) [io, fill=Aquamarine!30, minimum width=3.9cm, text width=3.6cm] at (6.3, -2.8) {WEASEL 2.0\\ \emph{(dictionary features)}};

        \node (convR)  [classifierbox, fill=Bittersweet!30] at (-7.3, -6.2) {Ridge};
        \node (convE)  [classifierbox, fill=Bittersweet!30] at (-6.3, -6.2) {Extra Trees};
        \node (convN)  [classifierbox, fill=Bittersweet!30] at (-5.3, -6.2) {Gaussian Naive Bayes};

        \node (intR)   [classifierbox, fill=Bittersweet!30] at (-3.1, -6.2) {Ridge};
        \node (intE)   [classifierbox, fill=Bittersweet!30] at (-2.1, -6.2) {Extra Trees};
        \node (intN)   [classifierbox, fill=Bittersweet!30] at (-1.1, -6.2) {Gaussian Naive Bayes};

        \node (shpR)   [classifierbox, fill=Bittersweet!30] at (1.1, -6.2) {Ridge};
        \node (shpE)   [classifierbox, fill=Bittersweet!30] at (2.1, -6.2) {Extra Trees};
        \node (shpN)   [classifierbox, fill=Bittersweet!30] at (3.1, -6.2) {Gaussian Naive Bayes};

        \node (dicR)   [classifierbox, fill=Bittersweet!30] at (5.3, -6.2) {Ridge};
        \node (dicE)   [classifierbox, fill=Bittersweet!30] at (6.3, -6.2) {Extra Trees};
        \node (dicN)   [classifierbox, fill=Bittersweet!30] at (7.3, -6.2) {Multinomial Naive Bayes};

        \node (stacked) [io, fill=WildStrawberry!30, minimum width=11.4cm, text width=11cm] at (0, -9.7) {
            Stacked meta-features (12 base learners)\\[4pt]
            {\raggedright
            $\bullet$ Training: out-of-fold, via $k$-fold cross-validation\\
            $\bullet$ Inference: held-out, from a single model fit on the entire training set\par}
        };

        \node (decision) [decision] at (0, -12.5) {Mean samples\\ per class $\geq$ 60.83?};

        \node (cawpe) [io, fill=Yellow!30, minimum width=3.4cm, text width=3.1cm] at (-5.5, -14.9) {Weighted hard\\ majority vote};
        \node (tabicl) [io, fill=Yellow!30, minimum width=3.4cm, text width=3.1cm] at (5.5, -14.9) {TabICLv2\\ meta-classifier};

        \node (output) [io, fill=Cyan!30] at (0, -17.3) {Final prediction};

        \coordinate (feat-wl) at (-8.3, -2.8);
        \coordinate (feat-wr) at (8.3, -2.8);
        \coordinate (base-wl) at (-8.3, -6.2);
        \coordinate (base-wr) at (8.3, -6.2);
        \coordinate (meta-wl) at (-8.3, -15.9);
        \coordinate (meta-wr) at (8.3, -15.9);
        \coordinate (labelx) at (-9.9, 0);

        \draw [thick] (input.south) -- ++(0,-0.5) coordinate (input-branch);
        \draw [arrow] (input-branch) -| (conv.north);
        \draw [arrow] (input-branch) -| (interval.north);
        \draw [arrow] (input-branch) -| (shapelet.north);
        \draw [arrow] (input-branch) -| (dictionary.north);

        \foreach \f/\a/\b/\c in {conv/convR/convE/convN, interval/intR/intE/intN, shapelet/shpR/shpE/shpN, dictionary/dicR/dicE/dicN} {
            \draw [thick] (\f.south) -- ++(0,-0.5) coordinate (\f-branch);
            \draw [arrow] (\f-branch) -| (\a.east);
            \draw [arrow] (\f-branch) -| (\b.east);
            \draw [arrow] (\f-branch) -| (\c.east);
        }

        \coordinate (mergeY) at ($(convE.west)+(0,-0.6)$);
        \coordinate (mainTrunk) at (stacked |- mergeY);
        \foreach \n in {convR,convE,convN,intR,intE,intN,shpR,shpE,shpN,dicR,dicE,dicN} {
            \draw [trunk] (\n.west) |- (mainTrunk);
        }
        \draw [arrow] (mainTrunk) -- (stacked);

        \draw [arrow] (stacked) -- (decision);
        \draw [arrow] (decision) -- node[above left, xshift=-2pt] {No} (cawpe);
        \draw [arrow] (decision) -- node[above right, xshift=2pt] {Yes} (tabicl);

        \draw [arrow] (cawpe) -- (output);
        \draw [arrow] (tabicl) -- (output);

        \coordinate (base-mid) at ($(convE.east)!0.5!(stacked.south)$);
        \coordinate (base-label-y) at (labelx |- base-mid);
        \node[stagelabel] at (base-label-y) {Base\\classification};

        \coordinate (meta-mid) at ($(decision.north)!0.5!(output.south)$);
        \coordinate (meta-label-y) at (labelx |- meta-mid);
        \node[stagelabel] at (meta-label-y) {Meta-classification\\(aggregation)};

        \node[stagebox, fit=(conv)(interval)(shapelet)(dictionary)(feat-wl)(feat-wr), inner sep=6pt] {};
        \node[stagebox, fit=(convR)(convE)(convN)(intR)(intE)(intN)(shpR)(shpE)(shpN)(dicR)(dicE)(dicN)(stacked)(base-wl)(base-wr), inner sep=6pt] {};
        \node[stagebox, fit=(decision)(cawpe)(tabicl)(output)(meta-wl)(meta-wr), inner sep=6pt] {};
    \end{tikzpicture}}

    \caption{
        Architecture of TIGER.
    }
    \label{fig:pipeline}
\end{figure}
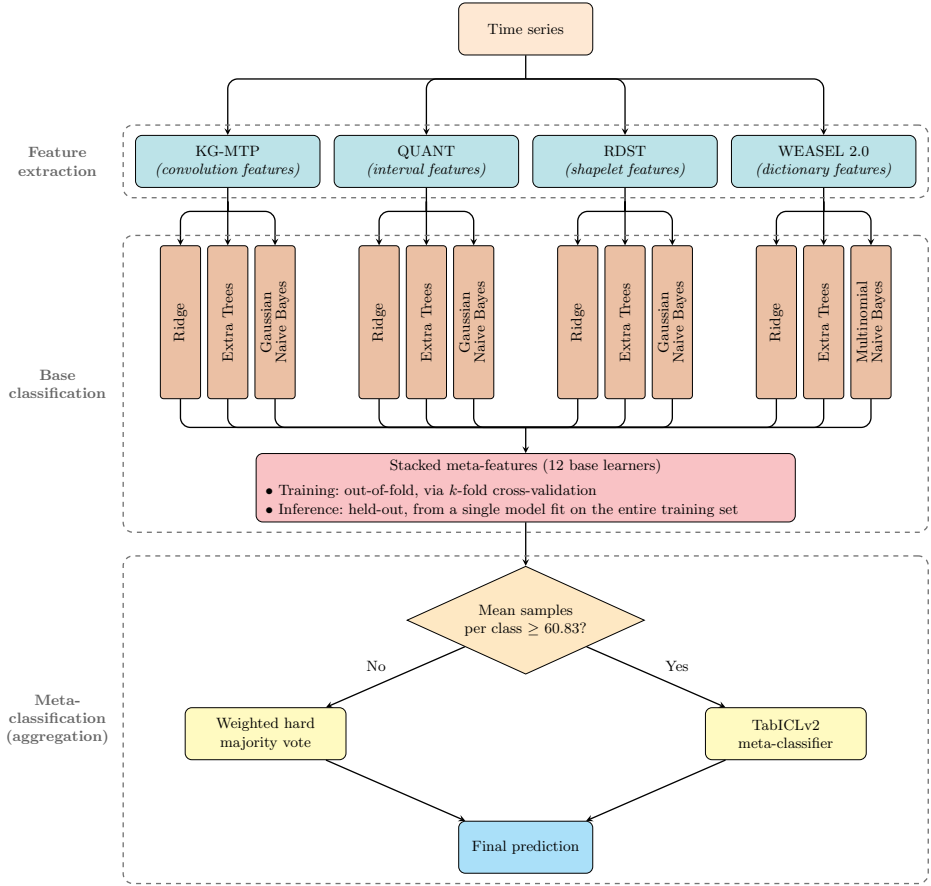

TIGER is a three-stage stacking ensemble.
In the first stage, four transformations from four distinct families widely used in the time series classification literature (convolution, interval, shapelet, and dictionary) are performed.
In the second stage, three diverse base classifiers are applied on top of the four transformations, giving twelve base learners in total.
In the third stage, the predictions of these twelve base learners are combined by a meta-classifier to produce the final prediction.
Unlike HIVE-COTE~2.0 \citep{middlehurstHIVECOTE20New2021a}, which fits one bespoke classifier per representation family and combines the four resulting predictions directly, TIGER decouples representation from classification: every family is combined with the same small portfolio of general-purpose classifiers, and a dedicated meta-classification stage learns (or is told, depending on the amount of available data) how to combine their outputs.
Figure~\ref{fig:pipeline} shows the final architecture of TIGER, with its three main steps: feature extraction, base classification, and meta-classification (aggregation).
In the remainder of this subsection, we further describe each stage.

\subsubsection{Feature extraction}

Time series classification algorithms can be grouped into families based on the type of representation from which they extract features.
TIGER draws on four such families (convolution-, interval-, shapelet-, and dictionary-based) chosen because each has a recent, well-established algorithm that is both fast and among the most accurate in its family: KG-MTP \citep{wangKernelGroupingTime2025} for convolution-based features, QUANT \citep{dempsterQuantMinimalistInterval2024} for interval-based features, RDST \citep{guillaumeRandomDilatedShapelet2022} for shapelet-based features, and WEASEL~2.0 \citep{schaferWEASEL20Random2023} for dictionary-based features.
For each family, TIGER reuses exactly the feature extraction (transformation) step of the corresponding algorithm, with every hyperparameter left at its published default value.
TIGER does not reuse these algorithms' own classification step: instead, every family's raw extracted features are passed, unmodified and without any dimensionality reduction, to the same portfolio of base classifiers described below.

\subsubsection{Base classification}

For each of the four families, we fit three classifiers directly on the family's raw extracted features: a Ridge classifier \citep{hoerlRidgeRegressionApplications1970, hoerlRidgeRegressionBiased1970}, an Extra Trees classifier \citep{geurtsExtremelyRandomizedTrees2006a}, and a Naive Bayes classifier.
The Naive Bayes variant is chosen per family to match the sign and type of that family's raw features.
Gaussian Naive Bayes is used for the convolution, interval, and shapelet families, whose features are real-valued (convolution and interval) or continuous but not genuine count data (shapelet).
Multinomial Naive Bayes is used for the dictionary family, whose WEASEL~2.0 features are non-negative integer bag-of-words counts.
This gives twelve base learners in total.

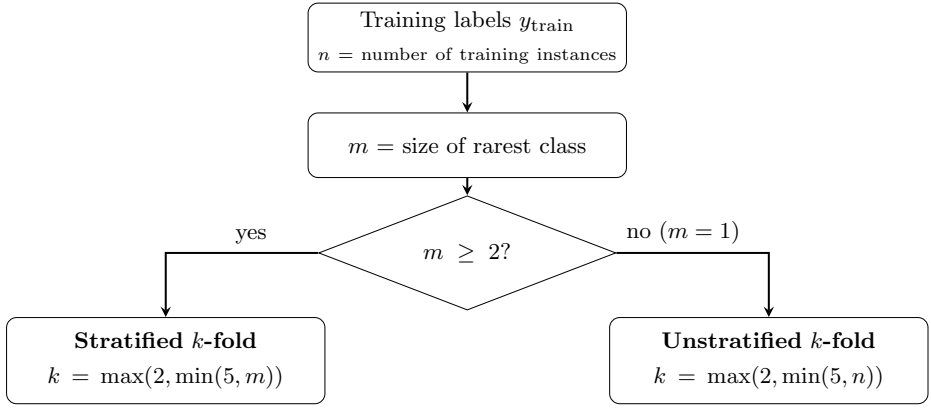
\begin{figure}[tbp]
    \centering
    \resizebox{0.95\linewidth}{!}{
    \begin{tikzpicture}[
        node distance=1.2cm and 2.0cm,
        every node/.style={font=\small},
    ]
        \tikzstyle{io} = [
            rectangle, rounded corners, minimum width=4.4cm, minimum height=0.9cm,
            text width=4.2cm, text centered, align=center, draw=black,
        ]
        \tikzstyle{decision} = [
            diamond, aspect=2.6, text width=2.6cm, text centered, align=center,
            draw=black,
        ]
        \tikzstyle{outcome} = [
            rectangle, rounded corners, minimum width=4.4cm, minimum height=1.2cm,
            text width=4.2cm, text centered, align=center, draw=black,
        ]
        \tikzstyle{arrow} = [thick,->,>=stealth]

        \node (input) [io] at (0,0) {Training labels $y_{\text{train}}$\\[2pt] \footnotesize $n$ = number of training instances};
        \node (m) [io] at (0,-1.5) {$m$ = size of rarest class};
        \node (decision) [decision] at (0,-3.) {$m \geq 2$?};

        \node (yes) [outcome] at (-4.2,-4.5) {
            \textbf{Stratified $k$-fold}\\[3pt]
            $k = \max(2, \min(5, m))$
        };
        \node (no) [outcome] at (4.2,-4.5) {
            \textbf{Unstratified $k$-fold}\\[3pt]
            $k = \max(2, \min(5, n))$
        };

        \draw[arrow] (input) -- (m);
        \draw[arrow] (m) -- (decision);
        \draw[arrow] (decision.west) -| node[pos=0.22,above,font=\small] {yes} (yes.north);
        \draw[arrow] (decision.east) -| node[pos=0.22,above,font=\small] {no ($m=1$)} (no.north);
    \end{tikzpicture}
    }
    \caption{
        Adaptive cross-validation splitting strategy for out-of-fold predictions.
    }
    \label{fig:cv-fold-selection}
\end{figure}

For every base learner, we compute two sets of predictions.
First, out-of-fold predictions on the training set are obtained via $k$-fold cross-validation, so that every training instance is scored by a model that never saw it during fitting.
The fold count $k$ and the choice between stratified and unstratified splitting are both determined adaptively from the training set's class distribution, as summarized in Figure~\ref{fig:cv-fold-selection}.
Concretely, letting $m$ denote the size of the training set's rarest class: whenever $m \geq 2$, we use a label-stratified $k$-fold split with $k = \max(2, \min(5, m))$, that is five folds by default, or as many folds as the rarest class allows (but never fewer than two) when it has fewer than five instances, so that every fold's test set still contains at least one instance of every class.
Whenever instead $m = 1$ (some class has only a single training instance), stratification is impossible: a class cannot appear in more than one fold's test set under a stratified split with $k \geq 2$, so a singleton class cannot be stratified across two or more folds at all.
We fall back to a plain unstratified $k$-fold split with $k = \max(2, \min(5, n))$, where $n$ is the total number of training instances.
Here nothing guarantees that the singleton class appears in every fold's training data.
This fallback is not merely a hypothetical edge case: some of the smallest data sets in our benchmark have as few as twenty training instances in total.

Second, held-out predictions on the test set are obtained from a single model fit on the entire training set.
The OOF predictions are what the meta-classification stage is trained on, while the test predictions are what the meta-classifier is ultimately applied to.
The two are never mixed, so that no information from the test set leaks into the meta-classification stage's own training procedure.

\subsubsection{Meta-classification (aggregation)}

The twelve base learners' (OOF for training, held-out for inference) predictions are concatenated into a single wide meta-feature matrix, one row per sample.
Class probabilities are used directly wherever a base learner provides them (Extra Trees, Naive Bayes), and Ridge's own (uncalibrated) decision function margins are used directly as meta-features otherwise.

TIGER's final, deployed meta-classification stage is an adaptive rule that selects, for each data set, between two methods, based on the mean number of training samples per class relative to a threshold $\tau = 60.83$ (Figure~\ref{fig:pipeline}).
Below the threshold, a weighted hard majority vote is used, in which every base learner casts a single vote for its own predicted class, weighted by its own out-of-fold accuracy raised to the power four.
At or above the threshold, a tabular in-context-learning foundation model, TabICLv2 \citep{quTabICLv2BetterFaster2026}, is fit directly on the meta-feature matrix instead, without any per-data-set gradient-based training.

\subsection{Computational complexity}\label{sec3sub2}

We now compare the asymptotic computational complexity of TIGER's own pipeline to that of HIVE-COTE~2.0, TIGER's strongest literature competitor.
Let $n$ denote a data set's number of training instances, $c$ its number of classes, and $t$ its series length.
Both algorithms are ensembles of several representation-specific components, so we give each component's own complexity in turn, as a function of $n$, $t$, $c$, and that component's own internal hyperparameters, before comparing the two totals.
These are asymptotic orders of growth, not measured wall-clock time: Section~\ref{sec5subsec3} reports the latter and interprets it in light of this analysis.
Several of the literature algorithms below are, in practice, run under a fixed wall-clock time budget (a ``contract'') rather than to asymptotic completion, which caps their worst-case cost at the expense of a potentially less thorough search as the archive grows.

TIGER's four feature-extraction steps are each near-linear in $n$ and $t$.
KG-MTP's convolution transform costs $O(n\kappa t)$ for $\kappa$ kernels, each a sliding-window operation over a length-$t$ series.
QUANT's interval transform has recently been shown to have an exact computational complexity of $\Theta(n \, t \, e \log t)$, where $e = \min(\delta, \lfloor \log_2 t \rfloor + 1)$ and $\delta$ is QUANT's own (fixed) recursion depth hyperparameter, simplifying to $\Theta(n \, t \log t)$ for any series longer than $2^{\delta - 1}$ (at least 32 time steps under QUANT's default $\delta = 6$) \citep{faouzi2026momentquantminimalistintervalmethod}, since extracting exact quantiles from its dyadic interval structure requires sorting each interval.
RDST's shapelet transform costs $O(nst)$ for $s$ randomly sampled shapelets, each compared against every training series by a single $O(t)$ sliding dilated-distance computation, since $s$ is fixed independently of $n$ and $t$ rather than grown by exhaustively enumerating candidate shapelets.
WEASEL~2.0's dictionary transform costs $O(nt)$ for its sliding-window symbolic (SFA) approximation.
Each of the resulting four feature matrices, of width $d_f$ for family $f \in \{\text{conv}, \text{int}, \text{shp}, \text{dict}\}$, is then passed to the same three base classifiers.
Ridge costs $O(n \, d_f \min(n, d_f))$ for its SVD-based closed-form fit, Extra Trees costs $O(b_{\text{ET}} \, n \, d_f \log n)$ for $b_{\text{ET}}$ trees, and Naive Bayes costs $O(n \, d_f)$.
Every base learner is fit $k \leq 5$ times for its out-of-fold predictions, a small constant multiplier rather than an additional asymptotic term.
The meta-classification stage adds either $O(nc)$, for weighted hard majority vote's vote tally, or TabICLv2's own in-context inference cost, depending on which the adaptive rule selects.
TabICLv2's own architecture costs $O(n^2 + np^2)$ for a table of $n$ rows and $p$ columns, quadratic in both the number of rows and the number of columns \citep{quTabICLv2BetterFaster2026}.
TIGER's meta-feature matrix has $p = 12c$ columns (twelve base learners times $c$ classes), so on many-class data sets, where $c$ pushes $p$ past TabICLv2's own pretrained operating range (2 to 100 columns), TIGER applies a column-grouping mechanism, splitting the meta-feature matrix into $\lceil p / 100 \rceil$ groups of at most 100 columns each, running TabICLv2 independently on each group, and averaging the resulting probabilities.
Since each group's own width is capped at a constant, grouping bounds the column-quadratic term to $O(n)$ per group, but multiplies the row-quadratic term by the number of groups, giving a total meta-classification cost of $O\!\left(n^2 \lceil p/100 \rceil\right)$ that grows, in the worst case, with the number of classes rather than staying fixed.

HIVE-COTE~2.0 instead pairs each of its four representation families with its own large, bespoke ensemble classifier, so classifier-side cost, rather than a small constant multiplier, is typically what dominates each module's total.
STC's own candidate shapelets are drawn from every subseries of every training series, giving $O(nt^2)$ candidates (one per series, start position, and length), each compared against all $n$ series at a further $O(nt^2)$ cost using a sliding-window distance, for a brute-force total of $O(n^2t^4)$ \citep{yeTimeSeriesShapelets2011}, markedly worse than RDST's $O(nst)$ precisely because $s$ is fixed a priori while STC's candidate pool grows with both $n$ and $t$.
This shapelet-search cost already dominates the further, smaller cost added by STC's own Rotation Forest classifier, an ensemble of decision trees each fit on a PCA-rotated random feature subset.
TDE searches a combinatorial space of SFA hyperparameter configurations (word length, alphabet size, normalization), guided by a Gaussian process model of that space, so its total cost scales with the number of configurations $g$ explored, each one requiring both an $O(nt)$ dictionary transform and an $O(n^2)$ nearest-neighbor classification step over all pairwise distances.
DrCIF fits a large ensemble of interval forests, each tree built over its own randomly sampled intervals across several derived representations of the series (raw, first-order differences, periodogram), so its cost scales as $O(b_{\text{DrCIF}} \, i \, nt)$ for $b_{\text{DrCIF}}$ trees and $i$ intervals per tree.
Arsenal, HIVE-COTE~2.0's convolution-based module, has the same near-linear structure as KG-MTP, $O(n \kappa t)$ for an ensemble of several such transforms.
CAWPE's own combination step, like weighted hard majority vote's, costs only $O(nc)$.

Table~\ref{tab:complexity} summarizes both breakdowns side by side.
The comparison mirrors the two architectures' own design principles: TIGER decouples representation from classification and pairs four already fast, near-linear transforms with the same small portfolio of cheap, search-free classifiers, so its total cost per data set grows close to linearly with $n$ and $t$ except for TabICLv2's own row-quadratic term, itself used only when the adaptive rule selects it and further inflated by column-grouping on many-class data sets.
HIVE-COTE~2.0 instead fuses each transform with its own bespoke ensemble classifier, and at least two of these four classifiers, STC's exhaustive shapelet search and TDE's Gaussian-process-guided hyperparameter search combined with nearest-neighbor classification, carry a substantially worse nominal dependence on $n$ or $t$ than any of TIGER's own components, an asymmetry only partly hidden by HIVE-COTE~2.0's own practice of contracting these searches to a fixed time budget rather than a fundamentally cheaper algorithm.

\begin{table}[tbp]
\centering
\sbox{\resizedtablebox}{\resizebox{\linewidth}{!}{
\begin{tabular}{lll}
\toprule
Module & Approximate complexity & Dominant cost \\
\midrule
\rowcolor{algorithm} \multicolumn{3}{l}{\textit{TIGER}} \\
KG-MTP (convolution transform) & $O(n\kappa t)$ & transform \\
QUANT (interval transform) & $\Theta(nte\log t)$ & transform \\
RDST (shapelet transform) & $O(nst)$ & transform \\
WEASEL~2.0 (dictionary transform) & $O(nt)$ & transform \\
Ridge (per family) & $O(n \, d_f \min(n, d_f))$ & classification \\
Extra Trees (per family) & $O(b_{\text{ET}} \, n \, d_f \log n)$ & classification \\
Naive Bayes (per family) & $O(n \, d_f)$ & classification \\
Weighted hard majority vote (when selected) & $O(nc)$ & aggregation \\
TabICLv2 (when selected) & $O(n^2 \lceil p/100 \rceil)$ & aggregation \\
\midrule
\rowcolor{algorithm} \multicolumn{3}{l}{\textit{HIVE-COTE~2.0}} \\
STC (shapelet search + Rotation Forest) & up to $O(n^2t^4)$ & shapelet search \\
TDE (dictionary ensemble + hyperparameter search) & $O(g(nt + n^2))$ & search + classification \\
DrCIF (interval forest) & $O(b_{\text{DrCIF}} \, i \, nt)$ & classification \\
Arsenal (convolution ensemble) & $O(n\kappa t)$ & transform \\
CAWPE combination & $O(nc)$ & aggregation \\
\bottomrule
\end{tabular}
}}
\caption{
    Asymptotic computational complexity of TIGER's and HIVE-COTE~2.0's modules.
}
\label{tab:complexity}
\usebox{\resizedtablebox}
\end{table}

\subsection{Hyperparameter values and optimization}\label{sec3sub3}

Among the three stages of the TIGER pipeline, feature extraction is left entirely at its published default hyperparameter values, and base classification almost entirely so, for the reasons given below.
One choice is instead tuned from data and actually shapes the deployed pipeline: the meta-classification stage's own candidate method and threshold, detailed below.
The out-of-fold estimator used in base classification, by contrast, is cross-validation throughout the deployed pipeline.

For feature extraction, every family's transformation step is left at the published default hyperparameter values of the corresponding literature algorithm (KG-MTP, QUANT, RDST, WEASEL~2.0), since each of these algorithms has already been shown to be a strong, well-tuned time series classifier in its own right with these default values.
Retuning them here, on top of TIGER's own base classification and meta-classification stages, would multiply the size of an already large hyperparameter search without a clear expectation of benefit.

For base classification, the Ridge classifier's own regularization strength is tuned automatically via efficient leave-one-out cross-validation, as is standard for this classifier in the time series classification literature.
The regularization strength is chosen over a per-family grid of candidate values matched, where a literature precedent exists, to that family's own reference implementation (KG-MTP, RDST, WEASEL~2.0), and extrapolated otherwise for the interval family, whose own QUANT design uses Extra Trees rather than Ridge.
Naive Bayes is left at its standard default hyperparameter values, being a classifier with essentially no hyperparameters to tune.
Extra Trees, by contrast, departs from \emph{scikit-learn}'s default configuration: every family uses 200 trees and the entropy splitting criterion (scikit-learn's own defaults are 100 trees and Gini impurity), and the interval family's \texttt{max\_features} is additionally set to $0.1$ of its own raw feature count, matching QUANT's implementation.
The other three families keep \emph{scikit-learn}'s default \texttt{max\_features} value, that is the square root of the number of features.

For the meta-classification stage, rather than committing to a single meta-classifier a priori, we considered six deployable candidate methods for combining the twelve base learners' predictions into a final one.
The first is a plurality (hard majority) vote over the base learners' own predicted classes.
The second is a multinomial logistic regression fit on the meta-feature matrix, with its regularization strength tuned by its own internal cross-validation.
The third is a gradient-boosted tree ensemble, LightGBM \citep{keLightGBMHighlyEfficient2017}, fit on the same meta-feature matrix.
The fourth is CAWPE \citep{linesTimeSeriesClassification2018a}, HIVE-COTE~2.0's own accuracy-weighted probability combination rule, in which every base learner's calibrated probabilities are averaged with weight equal to its own OOF accuracy raised to the power four.
The fifth is a weighted hard majority vote, the hard-voting analogue of CAWPE, in which every base learner casts a single vote for its own predicted class, weighted by the same accuracy-based weight, rather than averaging probabilities.
The sixth is a tabular in-context-learning foundation model, TabICLv2 \citep{quTabICLv2BetterFaster2026}, fit directly on the meta-feature matrix without any per-data-set gradient-based training.
Each candidate's own hyperparameters (the logistic regression's regularization strength, and CAWPE and weighted hard majority vote's shared accuracy-weighting exponent) are left at standard default values (respectively, internal cross-validation and the exponent of four already used by HIVE-COTE~2.0 \citep{middlehurstHIVECOTE20New2021a}), and TabICLv2 is used as a frozen, pretrained foundation model, with no per-data-set fine-tuning, so that this comparison is not itself confounded by an additional layer of hyperparameter tuning.
The result of this hyperparameter optimization procedure is what leads to TIGER's final, deployed meta-classification stage: an adaptive rule that selects, for each data set, between weighted hard majority vote and TabICLv2 based on the number of training samples available per class relative to a threshold $\tau = 60.83$ (Figure~\ref{fig:pipeline}).
The setup of this hyperparameter optimization procedure is described in Section~\ref{sec4sub1} and Section~\ref{sec4sub2}, and the results in Section~\ref{sec5subsec1}.

\section{Experimental setup}\label{sec4}

\subsection{Data sets and algorithms}\label{sec4sub1}

We use the UCR time series classification archive \citep{dauUCRTimeSeries2019e} to assess the performance of TIGER.
More precisely, we use the same 142 univariate time series classification data sets selected in a recent, large-scale review of the field \citep{middlehurstBakeReduxReview2024}.
All the reported results are the average of 30 resamples per data set, using the standard UCR-archive resampling procedure \citep{middlehurstBakeReduxReview2024}: resample 0 is each data set's original, published train/test split, and resamples 1--29 pool the training and test sets together and re-split them according to the same fixed resample indices used throughout the field.
Results are thus directly comparable with previously published per-resample scores.

To perform hyperparameter and architecture optimization, we use a development subset of twenty data sets among the 142 data sets, listed in Table~\ref{tab:devsubset}.
For this optimization, we use only the first five of each development data set's thirty resamples, rather than all thirty, to limit computational cost.
We also only use the training sets: the same cross-validation procedure (Figure~\ref{fig:cv-fold-selection}) is performed on each of these training sets to obtain the out-of-fold predictions that this tuning is based on.

We evaluate TIGER on all 142 data sets, including the twenty development data sets, when comparing to other algorithms, since TIGER's own architecture and hyperparameter were chosen using only cross-validated training performance on the development subset, never any test-set information.
Using all available data sets, rather than artificially excluding the development subset, is the conventional practice in the time series classification literature \citep{middlehurstBakeReduxReview2024}.
Only five of each development data set's thirty resamples ever entered the optimization procedure, but this should not be read as making the remaining twenty-five resamples equivalent to an independent, never-seen data set.
Table~\ref{tab:devsubset}'s training coverage column, the proportion of each data set's pooled training-plus-test instances that appear in at least one of the five training sets used for optimization, shows that for most development data sets these five training sets already span the large majority of that data set's own samples, so few individual samples in the remaining resamples are genuinely novel to the optimization procedure.
More fundamentally, even a resample built entirely from instances absent from all five training sets is still drawn from the same underlying data set, whose overall characteristics, not only its specific training instances, already shaped which out-of-fold estimator and threshold value were chosen.
All the decisions are thus made using only the development subset, and are then evaluated out-of-sample on the full 142-data-set benchmark.

As an additional, more conservative check, we also report results restricted to the 122 data sets that exclude the development subset, in order to show that our conclusions are not an artifact of any residual, indirect influence of the development subset on TIGER's design.

\begin{table}[htbp]
\centering
\sbox{\resizedtablebox}{\resizebox{\linewidth}{!}{%
\begin{tabular}{lrrrrrr}
\toprule
Data set & Training & Test & Classes & Length & Mean samples/class & Training coverage \\
\midrule
AsphaltPavementTypeUni & 1055 & 1056 & 3 & 2371 & 351.7 & 97.3\% \\
CinCECGTorso & 40 & 1380 & 4 & 1639 & 10.0 & 13.5\% \\
Computers & 250 & 250 & 2 & 720 & 125.0 & 96.8\% \\
Covid3Month\_disc & 140 & 61 & 3 & 84 & 46.7 & 100.0\% \\
EOGHorizontalSignal & 362 & 362 & 12 & 1250 & 30.2 & 96.5\% \\
EOGVerticalSignal & 362 & 362 & 12 & 1250 & 30.2 & 96.5\% \\
FacesUCR & 200 & 2050 & 14 & 131 & 14.3 & 37.1\% \\
FloodModeling1\_disc & 471 & 202 & 2 & 266 & 235.5 & 100.0\% \\
FloodModeling2\_disc & 466 & 201 & 2 & 266 & 233.0 & 99.4\% \\
FloodModeling3\_disc & 429 & 184 & 2 & 266 & 214.5 & 99.2\% \\
GestureMidAirD3 & 208 & 130 & 26 & 360 & 8.0 & 98.5\% \\
Haptics & 155 & 308 & 5 & 1092 & 31.0 & 88.3\% \\
LargeKitchenAppliances & 375 & 375 & 3 & 720 & 125.0 & 96.9\% \\
MoteStrain & 20 & 1252 & 2 & 84 & 10.0 & 7.7\% \\
OliveOil & 30 & 30 & 4 & 570 & 7.5 & 100.0\% \\
PigAirwayPressure & 104 & 208 & 52 & 2000 & 2.0 & 88.5\% \\
PigCVP & 104 & 208 & 52 & 2000 & 2.0 & 88.5\% \\
ProximalPhalanxOutlineAgeGroup & 400 & 205 & 3 & 80 & 133.3 & 99.2\% \\
ScreenType & 375 & 375 & 3 & 720 & 125.0 & 96.9\% \\
SemgHandMovementCh2 & 450 & 450 & 6 & 1500 & 75.0 & 95.9\% \\
\bottomrule
\end{tabular}%
}}
\caption{
    The twenty data sets forming TIGER's development subset.
}
\label{tab:devsubset}
\usebox{\resizedtablebox}
\end{table}

We compare TIGER to five other time series classification algorithms.
For our main comparisons, we include HIVE-COTE~2.0 \citep{middlehurstHIVECOTE20New2021a}, currently one of the most accurate published time series classification algorithms, as well as the four algorithms whose transformation steps TIGER itself reuses: QUANT \citep{dempsterQuantMinimalistInterval2024}, RDST \citep{guillaumeRandomDilatedShapelet2022}, WEASEL~2.0 \citep{schaferWEASEL20Random2023}, and KG-MTP \citep{wangKernelGroupingTime2025}, to assess how much TIGER's ensembling and stacking design adds on top of each individual transformation family used on its own.
For HC2, QUANT, RDST, and WEASEL~2.0, we use the results reported in \citep{middlehurstBakeReduxReview2024}, obtained from the official \emph{tsml-eval} results repository.\footnote{\url{https://github.com/time-series-machine-learning/tsml-eval/tree/main/tsml_eval/publications/y2023/tsc_bakeoff}}
For KG-MTP, we reproduce the results ourselves using the original authors' published implementation, since it was not included in this bake-off study.

\subsection{Metrics and comparisons}\label{sec4sub2}

To evaluate the algorithms, we use classification accuracy (ACC) as our primary metric, consistent with the great majority of the time series classification literature and with the reference results we compare against \citep{middlehurstBakeReduxReview2024}.
For our main comparison against other time series classification algorithms, we additionally report four secondary metrics, following the same convention as \citep{middlehurstBakeReduxReview2024}: balanced accuracy (BALACC), area under the receiver operating characteristic curve (AUROC), negative log-likelihood (NLL), and F1-score (F1).

To compare algorithms across several data sets, we use a modified version of the critical difference diagram \citep{demsarStatisticalComparisonsClassifiers2006a}, in which the post-hoc Nemenyi test is replaced with pairwise Wilcoxon signed-rank tests, with cliques formed using the Holm correction, as recommended in \citep{garciaExtensionStatisticalComparisons2008, benavoliShouldWeReally2016}.
We use a significance level of $0.05$ for all hypothesis tests, with one exception: the critical difference diagrams' clique formation uses a Holm-corrected value of $\alpha = 0.1$, as it is done in other time series classification studies.
In the critical difference diagrams, we compare the average ranks of several algorithms, in terms of accuracy, over a set of data sets, and cliques, represented by solid lines, indicate no significant difference between the corresponding algorithms.
We also report, for pairwise comparisons, the number of wins, losses, and ties, along with the mean accuracy difference and the $p$-value of a Wilcoxon signed-rank test.

\subsection{Experiments, implementation, and reproducibility}\label{sec4sub3}

All the experiments were run on a dual-socket server with two Intel(R) Xeon(R) Gold 5218R CPUs (20 cores / 40 threads each, 80 logical CPUs in total) and 540~GB of RAM, running Debian GNU/Linux 13 (trixie), using Python 3.13.5.
No hardware acceleration, such as graphical processing units (GPU), was used: every algorithm was run on CPU, including TabICLv2.
We used the following Python packages to implement TIGER itself, i.e., to run its final, deployed pipeline, whose meta-classification stage selects only between weighted hard majority vote and TabICLv2:
\begin{itemize}
    \item \emph{aeon} (1.6.0) \citep{middlehurstAeonPythonToolkit2024} is a popular Python package dedicated to time series analysis using machine learning.
    We used its implementations of the RDST and WEASEL~2.0 transformation steps.
    \item \emph{Numba} (0.63.1) \citep{lamNumbaLLVMbasedPython2015} is an open source just-in-time compiler that translates a subset of Python and NumPy code into fast machine code, used by our own implementations of the QUANT and KG-MTP transformation steps, following \citep{dempsterQuantMinimalistInterval2024} and \citep{wangKernelGroupingTime2025} respectively.
    Neither is (yet) available in \emph{aeon} as a NumPy/Numba implementation: KG-MTP has no \emph{aeon} implementation at all, and \emph{aeon}'s own QUANT implementation is a direct port of the original PyTorch code, which we have shown our own implementation to be faster than \citep{faouzi2026momentquantminimalistintervalmethod}.
    \item \emph{NumPy} (2.3.5) \citep{harrisArrayProgrammingNumPy2020} is a standard Python package for manipulating $n$-dimensional arrays, used throughout our own implementation, including our own implementations of the QUANT and KG-MTP transformation steps.
    \item \emph{scikit-learn} (1.7.2) \citep{pedregosaScikitlearnMachineLearning2011} is a popular Python package dedicated to machine learning, providing implementations of many algorithms as well as utility tools.
    We used its implementations of the Ridge, Extra Trees, and Naive Bayes classifiers, as well as its cross-validation utilities.
    \item \emph{SciPy} (1.17.1) \citep{virtanenSciPy10Fundamental2020} is used directly by our own implementation of the KG-MTP transformation step, for its implementation of the Hilbert transform.
    \item \emph{tabicl} (2.2.0) is the official implementation of TabICLv2 \citep{quTabICLv2BetterFaster2026}, used, wrapped with a grouping mechanism to respect its pretrained operating range, whenever the adaptive rule selects it as the meta-classifier.
\end{itemize}

Additionally, to perform all the experiments, including running the hyperparameter-optimization comparison across all six candidate meta-classification methods, the comparisons against other algorithms, computing the metrics described in Section~\ref{sec4sub2}, and generating the figures and tables in this paper, we also used the following Python packages:
\begin{itemize}
    \item \emph{LightGBM} (4.7.0) \citep{keLightGBMHighlyEfficient2017} provides the gradient-boosted tree implementation used as one of the six candidate meta-classification methods compared in Section~\ref{sec5subsec1}.
    \item \emph{Matplotlib} (3.11.2) \citep{hunterMatplotlib2DGraphics2007} is a comprehensive library for creating static, animated, and interactive visualizations in Python, and was used to generate the figures.
    \item \emph{pandas} (2.3.3) \citep{mckinney-proc-scipy-2010} is a fast, powerful, flexible and easy to use open source data analysis and manipulation tool.
    \item \emph{tabpfn-extensions} (0.6.2), providing the error-correcting-output-codes wrapper for TabICLv2's many-class classification.
\end{itemize}

The source code is publicly available on a GitHub repository.\footnote{\url{https://github.com/johannfaouzi/TIGER}}
We provide detailed instructions so that our results can be easily reproduced by anyone.

\section{Results}\label{sec5}

We present our results in several subsections.
Section~\ref{sec5subsec1} describes how the meta-classification stage of TIGER was chosen, and validates the resulting adaptive rule out-of-sample on the full 142-data-set benchmark.
In Section~\ref{sec5subsec2}, we provide the results of the comparisons between TIGER and existing time series classification algorithms.
Section~\ref{sec5subsec3} discusses runtimes, Section~\ref{sec5subsec4} presents our ablation experiments, and Section~\ref{sec5subsec5} reports further analyses of TIGER's behavior.

\subsection{Hyperparameter optimization}\label{sec5subsec1}

We compare six candidate meta-classification methods on the twenty-data-set development subset: hard majority vote, logistic regression, LightGBM, CAWPE, weighted hard majority vote, and TabICLv2.
We additionally report two reference points: the single best base learner, chosen independently per data set by cross-validated training accuracy, and TIGER's own adaptive rule (``auto-selected''), which chooses between weighted hard majority vote and TabICLv2 using the threshold $\tau$ derived in Section~\ref{sec3sub3}.

Table~\ref{tab:dev-subset-results} reports the mean accuracy (over five resamples) of each of the six candidate methods on every one of the twenty development data sets individually, sorted by the mean number of training samples per class, and makes explicit the pattern that motivates the threshold $\tau$.
Weighted hard majority vote obtains the best mean accuracy among the eleven data sets with fewer than $\tau$ samples per class (0.799, against 0.763 for TabICLv2 and 0.787 for the next-best fixed alternative, hard majority vote), while TabICLv2 obtains the best mean accuracy among the nine data sets at or above $\tau$ (0.898, against 0.872 for weighted hard majority vote).

TabICLv2's failures are concentrated, and severe, at the very smallest sample sizes.
The two most extreme cases are PigAirwayPressure and PigCVP, both 52-class data sets with only two training samples per class: TabICLv2 obtains a mean accuracy of only 0.543 on PigAirwayPressure, 22.7 accuracy points below weighted hard majority vote's 0.770, and 0.809 on PigCVP, 10.2 points below weighted hard majority vote's 0.911.
Both are also comfortably beaten by the parameter-free hard majority vote (0.764 and 0.911, respectively).
Every other candidate method that fits parameters on the meta-feature matrix fails just as badly or worse on these two data sets.
Logistic regression obtains 0.394 and 0.834, and LightGBM 0.473 and 0.784, on PigAirwayPressure and PigCVP respectively.
These results indicate that this failure mode is not specific to TabICLv2's own in-context learning mechanism, but a more general symptom of asking any method to learn a combination rule for 52 classes from only two labeled meta-training examples per class.
The two vote-based methods, which fit no parameters at the meta-classification stage at all, are the only ones that remain robust in this regime.
Outside these two extreme, many-class, few-shot data sets, TabICLv2 already competes closely with, or occasionally beats, weighted hard majority vote even below $\tau$ (e.g., EOGVerticalSignal, Haptics, OliveOil).
The threshold $\tau = 60.83$ trades away these smaller, more marginal losses in favor of avoiding the occasional catastrophic ones illustrated by PigAirwayPressure and PigCVP.

\begin{table}[tbp]
\centering
\sbox{\resizedtablebox}{\resizebox{\linewidth}{!}{%
\begin{tabular}{lrcccccc}
\toprule
Data set & Mean samples/class & HMV & LR & LightGBM & CAWPE & WHMV & TabICLv2 \\
\midrule
PigCVP & 2 & \textbf{0.911} & 0.834 & 0.784 & 0.898 & \textbf{0.911} & 0.809 \\
PigAirwayPressure & 2 & 0.764 & 0.394 & 0.473 & 0.537 & \textbf{0.770} & 0.543 \\
OliveOil & 7.5 & 0.880 & 0.900 & 0.893 & \textbf{0.907} & 0.887 & 0.900 \\
GestureMidAirD3 & 8 & 0.526 & 0.474 & 0.452 & 0.528 & \textbf{0.537} & 0.508 \\
CinCECGTorso & 10 & 0.990 & 0.976 & 0.970 & 0.990 & \textbf{0.995} & 0.989 \\
MoteStrain & 10 & 0.937 & 0.942 & 0.931 & 0.941 & \textbf{0.943} & 0.940 \\
FacesUCR & 14.3 & 0.940 & 0.947 & 0.947 & 0.949 & \textbf{0.960} & 0.959 \\
EOGHorizontalSignal & 30.2 & 0.787 & 0.789 & 0.806 & 0.802 & \textbf{0.818} & 0.816 \\
EOGVerticalSignal & 30.2 & 0.721 & 0.749 & 0.738 & 0.745 & 0.755 & \textbf{0.757} \\
Haptics & 31 & 0.554 & 0.547 & 0.550 & 0.538 & 0.564 & \textbf{0.566} \\
Covid3Month\_disc & 46.7 & \textbf{0.649} & 0.561 & 0.574 & 0.626 & 0.646 & 0.607 \\
\midrule
SemgHandMovementCh2 & 75 & 0.767 & 0.885 & 0.876 & 0.841 & 0.850 & \textbf{0.906} \\
Computers & 125 & 0.812 & 0.855 & \textbf{0.865} & 0.818 & 0.829 & 0.864 \\
ScreenType & 125 & 0.588 & 0.684 & 0.678 & 0.664 & 0.676 & \textbf{0.701} \\
LargeKitchenAppliances & 125 & 0.903 & 0.922 & 0.926 & 0.920 & 0.924 & \textbf{0.933} \\
ProximalPhalanxOutlineAgeGroup & 133.3 & \textbf{0.859} & 0.853 & 0.848 & 0.845 & \textbf{0.859} & 0.855 \\
FloodModeling3\_disc & 214.5 & 0.862 & 0.932 & \textbf{0.936} & 0.897 & 0.900 & \textbf{0.936} \\
FloodModeling2\_disc & 233 & 0.952 & 0.980 & 0.984 & 0.957 & 0.961 & \textbf{0.987} \\
FloodModeling1\_disc & 235.5 & 0.923 & 0.950 & 0.949 & 0.936 & 0.947 & \textbf{0.955} \\
AsphaltPavementTypeUni & 351.7 & 0.898 & 0.941 & 0.934 & 0.898 & 0.907 & \textbf{0.943} \\
\bottomrule
\end{tabular}%
}}
\caption{
    Mean accuracy of six candidate meta-classification methods per development data set (HMV = hard majority vote, LR = logistic regression, WHMV = weighted hard majority vote).
    The best value per row is bold, and the rule marks $\tau = 60.83$.
}
\label{tab:dev-subset-results}
\usebox{\resizedtablebox}
\end{table}

\begin{figure}[tbp]
    \centering
    \includegraphics[width=0.85\textwidth]{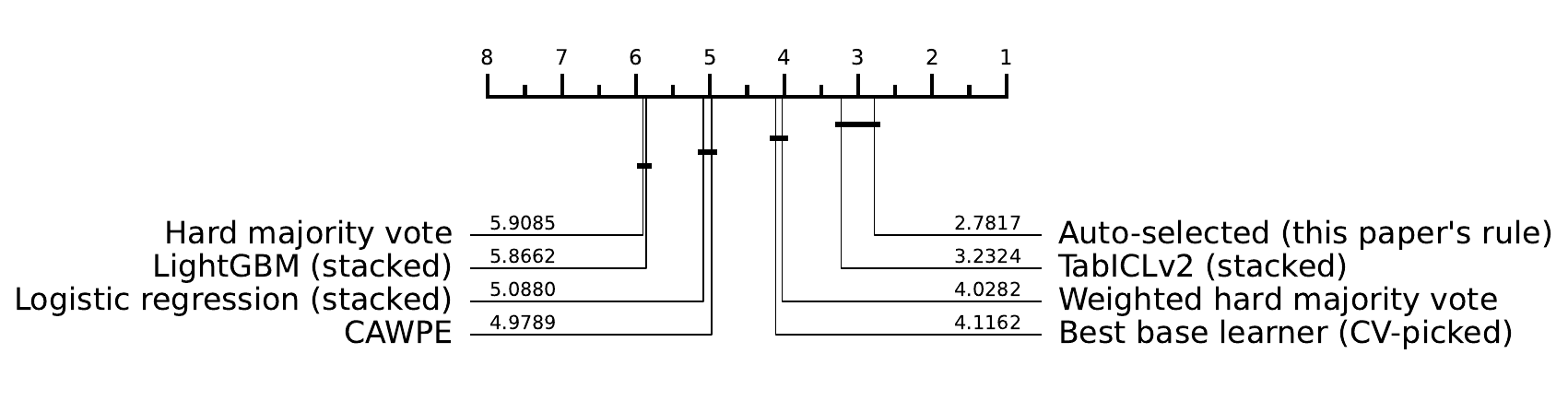}
    \caption{
        Critical difference diagram for the six candidate meta-classification methods, the best base learner (picked using cross-validation), and TIGER's adaptive rule, over the full 142-data-set benchmark, in terms of mean accuracy.
    }
    \label{fig:cd-methods}
\end{figure}

Figure~\ref{fig:cd-methods} shows the critical difference diagram comparing the mean ranks (in terms of accuracy) of all eight methods on the full 142-data-set benchmark.
The auto-selected rule obtains the best mean rank (2.78), essentially tied with TabICLv2 alone (3.23) and forming a clique with it, meaningfully ahead of every fixed alternative: weighted hard majority vote and the best base learner form a second clique around rank 4.0-4.1, CAWPE and logistic regression a third around rank 5.0-5.1, and LightGBM and hard majority vote the weakest clique, around rank 5.9.
This confirms that no single fixed meta-classification method dominates uniformly, and that TIGER's adaptive rule successfully identifies, on a per-data-set basis, which of its two constituent methods to trust.

\begin{figure}[tbp]
    \centering
    \includegraphics[width=\textwidth]{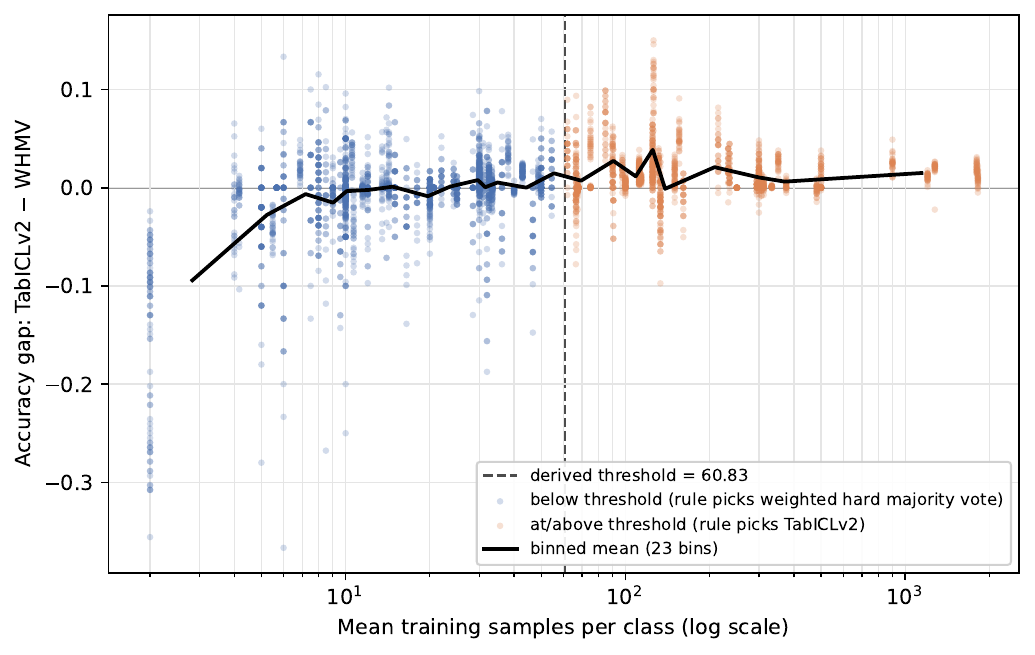}
    \caption{
        TabICLv2 minus weighted-hard-majority-vote accuracy gap against training samples per class, with threshold $\tau = 60.83$.
    }
    \label{fig:threshold}
\end{figure}

We further validate that the threshold $\tau = 60.83$, derived using only the twenty-data-set development subset, generalizes to the full evaluation benchmark.
Figure~\ref{fig:threshold} plots, for every (data set, resample) row of the 142-data-set benchmark, the accuracy gap between TabICLv2 and weighted hard majority vote against the number of training samples per class, together with the binned mean gap.
Below $\tau$, the binned mean gap is consistently and, for the smallest sample sizes, substantially negative (as low as $-0.095$ for the very smallest bin), meaning that weighted hard majority vote is the safer choice, exactly as the adaptive rule prescribes.
At and above $\tau$, the binned mean gap hovers close to zero and is occasionally positive, meaning that the two methods are roughly interchangeable on average, with TabICLv2 never substantially behind.

A concrete illustration of the adaptive rule's benefit is given by the ten data sets on which TabICLv2 alone loses the most, in accuracy, relative to WHMV.
Table~\ref{tab:tabicl-losses} provides the results.
Every one of these ten data sets has fewer than $\tau$ training samples per class, so the adaptive rule redirects every one of them to weighted hard majority vote.
Doing so raises the mean accuracy over these ten data sets from 0.671 to 0.748.
On the remaining 132 data sets, auto-selected's mean accuracy is within 0.0004 of TabICLv2 alone, essentially unaffected.
Appendix~\ref{secA} provides the characteristics of the 142 data sets, while Appendix~\ref{secB} provides the mean accuracy scores (over the thirty resamples) of the six meta-classification methods on the 142 data sets.

\begin{table}[tbp]
\caption{
    The ten data sets on which TabICLv2 loses the most, in accuracy, compared to WHMV.
}
\label{tab:tabicl-losses}
\centering
\begin{tabular}{lcccccc}
\toprule
Data set & Mean samples/class & TabICLv2 & WHMV & $\Delta$ \\
\midrule
PigAirwayPressure & 2.0 & 0.602 & 0.865 & -0.264 \\
PigCVP & 2.0 & 0.823 & 0.928 & -0.105 \\
Beef & 6.0 & 0.702 & 0.778 & -0.076 \\
PigArtPressure & 2.0 & 0.892 & 0.963 & -0.071 \\
Rock & 5.0 & 0.839 & 0.883 & -0.043 \\
WordSynonyms & 10.7 & 0.722 & 0.765 & -0.043 \\
Herring & 32.0 & 0.594 & 0.635 & -0.041 \\
DodgerLoopDay & 9.6 & 0.584 & 0.625 & -0.041 \\
Phoneme & 5.5 & 0.357 & 0.397 & -0.040 \\
Covid3Month\_disc & 46.7 & 0.599 & 0.638 & -0.039 \\
\bottomrule
\end{tabular}
\end{table}

\subsection{Comparisons to other time series classification algorithms}\label{sec5subsec2}

We compare TIGER to the five algorithms described in Section~\ref{sec4sub1}: HC2, QUANT, RDST, WEASEL~2.0, and KG-MTP.
Our main comparison uses all 142 data sets of the evaluation benchmark, including the twenty data sets of TIGER's own development subset.
We additionally report the same comparison restricted to the 122 held-out data sets (the 142 data sets excluding the development subset) as a more conservative check, in case of any residual, indirect influence of the development subset on TIGER's own design.

Table~\ref{tab:multi-metric-142} presents, for the full 142-data-set benchmark, all five metrics described in Section~\ref{sec4sub2} (ACC, BALACC, AUROC, NLL, and F1) for TIGER and the five literature algorithms, with algorithms ordered by accuracy.
TIGER obtains both the best mean value and the best mean rank on accuracy, balanced accuracy, and F1-score (0.884, 0.863, and 0.863, respectively).
The AUROC and NLL need to be interpreted in context: RDST and WEASEL~2.0 use Ridge classifiers, which provide a predicted class and a per-class score (signed distance to the hyperplane) but no calibrated probability estimate.
Following the same convention as \citep{middlehurstBakeReduxReview2024}, we substitute a one-hot probability vector for the missing estimate rather than the raw decision scores, and this substitution is why both metrics come out poorly here: NLL is unbounded whenever the prediction is wrong, and AUROC discards most of its ranking signal once only the single predicted class is scored instead of the full per-class ordering.
TIGER is only partly exempt: its meta-classification stage adaptively defers to weighted hard majority vote, a hard-voting rule with no native probabilistic output of its own, on 85 of the 142 data sets, and to TabICLv2, which does output genuine class probabilities, on the remaining 57.
For AUROC and NLL, we assign a one-hot probability vector (1 for the predicted class, 0 for every other class) whenever weighted hard majority vote is selected, and use TabICLv2's own predicted class probabilities whenever it is selected instead.
Under this convention, TIGER's AUROC (0.933) and NLL (2.775) rank third among the six algorithms, behind HC2 (0.962 and 0.403) and QUANT (0.956 and 0.533).
This reflects the one-hot fallback's own limitation on the 85 data sets where it applies, rather than a genuine calibration weakness, since a one-hot probability vector is either perfectly confident and correct or perfectly confident and wrong, with no genuine predictive uncertainty.
Appendix~\ref{secC} isolates the 57 data sets on which TIGER always selects TabICLv2, where its reported probabilities are genuine throughout: there, TIGER obtains the best AUROC and NLL as well, ahead of HC2.
Appendix~\ref{secD} reports the mean score (over the thirty resamples) of TIGER for each five metrics and for each of the 142 data sets.

\begin{figure}[tbp]
    \centering
    \includegraphics[width=0.8\textwidth]{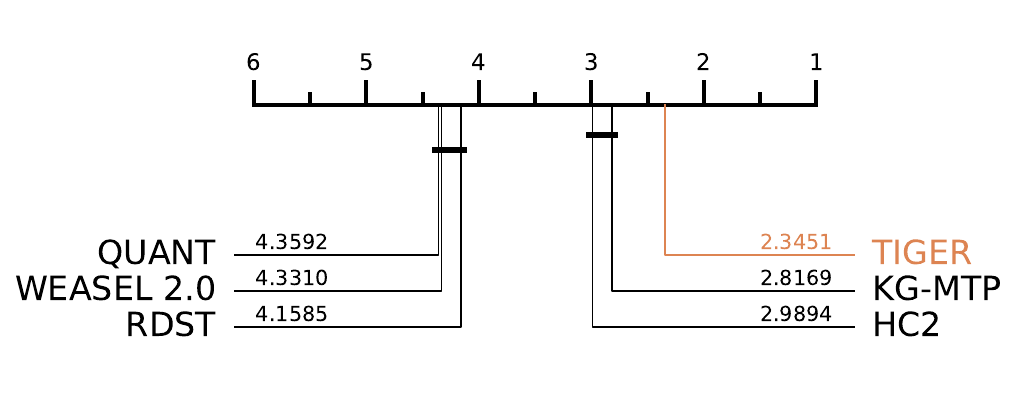}
    \caption{
        Critical-difference diagram for TIGER and the five literature algorithms on accuracy, over the full 142-data-set benchmark.
    }
    \label{fig:cd-literature}
\end{figure}

Pairwise Wilcoxon signed-rank tests on accuracy show that TIGER significantly outperforms every one of the five literature algorithms individually.
As expected, given that TIGER builds directly on top of their own transformation steps, the margin is largest against the three single-transformation algorithms: QUANT (mean difference $+0.029$, $p = 3.6 \times 10^{-18}$), WEASEL~2.0 ($+0.026$, $p = 5.1 \times 10^{-15}$), and RDST ($+0.021$, $p = 7.1 \times 10^{-14}$).
Against the two other ensemble-style algorithms, the margin is smaller but, evaluated on the full 142-data-set benchmark, still statistically significant: TIGER beats KG-MTP on 83 data sets and loses on 53, with 6 ties (mean difference $+0.0075$, $p = 1.5 \times 10^{-3}$), and beats HC2 on 80 data sets and loses on 55, with 7 ties (mean difference $+0.0080$, $p = 1.7 \times 10^{-4}$).
Figure~\ref{fig:cd-literature} summarizes these pairwise comparisons as a critical-difference diagram over mean accuracy rank across the full 142-data-set benchmark: TIGER's rank forms its own clique, the two other ensemble-style algorithms (KG-MTP, HC2) form a second, and the three single-transformation algorithms (RDST, WEASEL~2.0, QUANT) form a third, consistent with the pairwise test results above and with their accuracy ranking in Table~\ref{tab:multi-metric-142}.

Table~\ref{tab:multi-metric-122-heldout} repeats this comparison on the 122 held-out data sets.
The accuracy ranking of the six algorithms is unchanged, and TIGER again obtains the best accuracy, balanced accuracy, and F1-score, with AUROC and NLL again ranking third behind HC2 and QUANT for the same reason as above.
The accuracy margin narrows slightly against HC2 (mean difference $+0.0072$, $p = 5.4 \times 10^{-4}$) and against KG-MTP (mean difference $+0.0049$, $p = 3.8 \times 10^{-2}$), but both comparisons remain significant at the $0.05$ level.
We conclude that TIGER is competitive with, and on the full evaluation benchmark significantly outperforms, the current state of the art, and that this conclusion is not an artifact of the development subset used to design TIGER itself.

\begin{table}[tbp]
\caption{
    Summary performance measures of TIGER and five literature algorithms on the full 142-data-set benchmark, ordered by accuracy.
    Ranks are in parentheses, the best value per column is bold, and ``--'' marks a metric not reported.
}
\label{tab:multi-metric-142}
\centering
\begin{tabular}{lccccc}
\toprule
Algorithm & ACC & BALACC & AUROC & NLL & F1 \\
\midrule
TIGER & \textbf{0.884 (1)} & \textbf{0.863 (1)} & 0.933 (3) & 2.775 (3) & \textbf{0.863 (1)} \\
KG-MTP & 0.876 (2) & 0.855 (2) & 0.906 (4) & 4.457 (4) & 0.857 (2) \\
HC2 & 0.876 (3) & 0.848 (3) & \textbf{0.962 (1)} & \textbf{0.403 (1)} & -- \\
RDST & 0.863 (4) & 0.836 (4) & 0.893 (5) & 4.943 (5) & 0.835 (3) \\
WEASEL~2.0 & 0.858 (5) & 0.830 (6) & 0.888 (6) & 5.114 (6) & 0.825 (5) \\
QUANT & 0.855 (6) & 0.830 (5) & 0.956 (2) & 0.533 (2) & 0.831 (4) \\
\bottomrule
\end{tabular}
\end{table}

\begin{table}[tbp]
\caption{
    Summary performance measures on the 122 held-out data sets, ordered by accuracy.
    Ranks are in parentheses, the best value per column is bold, and ``--'' marks a metric not reported.
}
\label{tab:multi-metric-122-heldout}
\centering
\begin{tabular}{lccccc}
\toprule
Algorithm & ACC & BALACC & AUROC & NLL & F1 \\
\midrule
TIGER & \textbf{0.889 (1)} & \textbf{0.867 (1)} & 0.935 (3) & 2.606 (3) & \textbf{0.867 (1)} \\
KG-MTP & 0.884 (2) & 0.862 (2) & 0.911 (4) & 4.186 (4) & 0.865 (2) \\
HC2 & 0.882 (3) & 0.855 (3) & \textbf{0.963 (1)} & \textbf{0.387 (1)} & -- \\
RDST & 0.872 (4) & 0.847 (4) & 0.902 (5) & 4.613 (5) & 0.849 (3) \\
WEASEL~2.0 & 0.871 (5) & 0.845 (5) & 0.900 (6) & 4.654 (6) & 0.844 (4) \\
QUANT & 0.865 (6) & 0.838 (6) & 0.957 (2) & 0.509 (2) & 0.840 (5) \\
\bottomrule
\end{tabular}
\end{table}

The architecture of TIGER is slightly different for three data sets: DiatomSizeReduction (4 classes), FiftyWords (50 classes), and Phoneme (39 classes).
They are the only datasets for which at least one class has exactly one training sample.
As Section~\ref{sec3sub1} describes, a singleton class cannot be stratified across folds, and the resulting unstratified fallback does not guarantee that the class appears in every fold's training data.
When a fold's training data is missing a class entirely, Ridge outputs no value for that class, and the corresponding out-of-fold scores are discarded as unreliable rather than used.
Extra Trees and Naive Bayes do not raise the same error, since their predicted probabilities can validly be zero for a class that they never saw.
The consequence is that none of the four representation families contributes a Ridge base learner for them: TIGER runs with eight base learners rather than twelve, for every method reported in this section, including its own deployed adaptive rule.
None of the three data sets belongs to the twenty-data-set development subset, so both Table~\ref{tab:multi-metric-142} and Table~\ref{tab:multi-metric-122-heldout} include them under this reduced portfolio.
The comparisons above are consequently not evaluating a strictly identical TIGER configuration across all 142 data sets.
This is a conservative rather than a fundamental limitation: only the specific (row, class) pairs a fold's training data never saw are genuinely uninformative.
Extra Trees and Naive Bayes are not actually exempt either: the same cross-validation call silently zero-fills their own scores for those same rows.
This choice is numerically harmless for a bounded probability but is no more principled than discarding Ridge's.
Recovering this information consistently across all three base learners, rather than singling out Ridge, is left to future work.
Nonetheless, one could argue that a data set with a single training sample for at least one class is a pathological case in classification.

\subsection{Runtimes}\label{sec5subsec3}

Tables~\ref{tab:runtimes-sum}, \ref{tab:runtimes-train}, and \ref{tab:runtimes-inference} report wall-clock total (training-plus-inference) time, training time, and inference time, respectively, for TIGER and the other five algorithms.
Unlike the accuracy comparison of Section~\ref{sec5subsec2}, published runtime figures are not usable here, since hardware differences between papers make wall-clock comparisons meaningless.
QUANT, RDST, WEASEL~2.0, and KG-MTP were therefore run locally, on the same hardware as TIGER itself, purely to obtain comparable timings.
HC2 was run locally for the same reason, using \emph{aeon} 1.6.0 (Section~\ref{sec4sub3}), whose release included speed improvements to several of HC2's own component algorithms (DrCIF, Arsenal, TDE) relative to earlier \emph{aeon} versions.
The runtime figures reported below reflect this faster implementation rather than an older, slower one.
Training time covers feature extraction (or transformation) on the training set and model fitting, including, for TIGER, the twelve base learners' out-of-fold cross-validation fitting and final full-training-set fit alongside the selected meta-classifier's own fit.
Inference time covers the analogous steps on the test set, namely feature extraction (or transformation) and prediction.
Because per-data-set runtime is heavily right-skewed by data set size (a single large data set can cost orders of magnitude more than a typical one), each algorithm's own resamples for a given data set are first collapsed to that data set's median (training, inference, or summed) time, and each table reports the total, median, minimum, and maximum of these 142 per-data-set medians.

Table~\ref{tab:runtimes-sum} reports the combined training-plus-inference runtime.
TIGER's median runtime is 145.4s per data set, against 834.7s for HC2, making TIGER about 5.7 times faster than HC2 on average.
Over the full 142-data-set benchmark, TIGER's total runtime is 20.65h, against HC2's 149.25h, making TIGER 7.2 times faster than HC2.
These differences between TIGER and HC2 are bigger for training than for inference.

Table~\ref{tab:runtimes-train} reports the training runtime.
TIGER's median training runtime is 76.3s per data set, against 418.0s for HC2, making TIGER about 5.5 times faster than HC2 on average.
Over the full 142-data-set benchmark, TIGER's total training runtime is 10.90h, against HC2's 100.86h, making TIGER 9.3 times faster than HC2.
The difference in ratio between the median and total training runtimes is likely explained by the different computational complexities (Section~\ref{sec3sub2}), making HC2 much slower than TIGER for large data sets.

Table~\ref{tab:runtimes-inference} reports the inference runtime.
TIGER's median inference runtime is 39.5s per data set, against 248.2s for HC2, making TIGER about 6.3 times faster than HC2 on average.
Over the full 142-data-set benchmark, TIGER's total inference runtime is 9.76h, against HC2's 48.34h, making TIGER 5.0 times faster than HC2.
For inference, the difference in ratio between the median and total inference runtimes is probably explained by the choice of meta-classifier for TIGER: WHMV is very cheap and much faster than TabICLv2, whereas HC2 always uses CAWPE, which is also very cheap and fast.

\begin{table}[tbp]
\caption{
    Combined training-plus-inference wall-clock runtime for TIGER and five literature algorithms, ordered by median runtime (lowest in bold).
}
\label{tab:runtimes-sum}
\centering
\begin{tabular}{lcccc}
\toprule
Algorithm & Total (h) & Median (s) & Minimum (s) & Maximum (min) \\
\midrule
QUANT & \textbf{0.26} & \textbf{2.0} & \textbf{0.3} & \textbf{1.1} \\
KG-MTP & 2.08 & 14.4 & 1.2 & 10.4 \\
RDST & 3.11 & 19.3 & 1.0 & 13.8 \\
WEASEL~2.0 & 5.94 & 24.8 & 1.6 & 43.2 \\
TIGER & 20.65 & 145.4 & 19.0 & 213.2 \\
HC2 & 149.25 & 834.7 & 55.6 & 1469.9 \\
\bottomrule
\end{tabular}
\end{table}

\begin{table}[tbp]
\caption{
    Training wall-clock runtime for TIGER and five literature algorithms, ordered by median runtime (lowest in bold).
}
\label{tab:runtimes-train}
\centering
\begin{tabular}{lcccc}
\toprule
Algorithm & Total (h) & Median (s) & Minimum (s) & Maximum (min) \\
\midrule
QUANT & \textbf{0.25} & \textbf{1.6} & \textbf{0.2} & \textbf{1.1} \\
KG-MTP & 0.98 & 5.8 & 0.2 & 9.3 \\
RDST & 1.52 & 8.4 & 0.2 & 9.9 \\
WEASEL~2.0 & 2.90 & 11.6 & 0.8 & 17.3 \\
TIGER & 10.90 & 76.3 & 15.2 & 114.4 \\
HC2 & 100.86 & 418.0 & 17.6 & 1331.4 \\
\bottomrule
\end{tabular}
\end{table}

\begin{table}[tbp]
\caption{
    Inference wall-clock runtime for TIGER and five literature algorithms, ordered by median runtime (lowest in bold).
}
\label{tab:runtimes-inference}
\centering
\begin{tabular}{lcccc}
\toprule
Algorithm & Total (h) & Median (s) & Minimum (s) & Maximum (min) \\
\midrule
QUANT & \textbf{0.02} & \textbf{0.1} & \textbf{0.0} & \textbf{0.1} \\
KG-MTP & 1.10 & 7.1 & 0.4 & 6.5 \\
RDST & 1.58 & 8.0 & 0.3 & 11.9 \\
WEASEL~2.0 & 3.04 & 8.7 & 0.4 & 39.3 \\
TIGER & 9.76 & 39.5 & 1.8 & 135.7 \\
HC2 & 48.34 & 248.2 & 14.7 & 361.5 \\
\bottomrule
\end{tabular}
\end{table}

\subsection{Ablation experiments}\label{sec5subsec4}

We perform two ablation experiments, assessing two of TIGER's design decisions in turn.

\subsubsection{Choice of out-of-fold estimator}

For each of the twelve (family, base learner) pairs, we evaluate several candidate out-of-fold train-score estimators: $k$-fold cross-validation, 15-/25-model bagging out-of-bag, and, for Extra Trees, its own native out-of-bag estimate.

Table~\ref{tab:oof-estimator-ablation} reports, for each pair, the mean absolute gap between each candidate estimator's own out-of-fold accuracy and the corresponding held-out test accuracy, computed over all 142 data sets $\times$ 30 resamples.
A smaller gap means a more honest, less optimistic estimator.
$k$-fold CV is the most honest estimator for eleven of the twelve pairs.
The sole exception is KG-MTP's Naive Bayes learner, for which 25-model bagging out-of-bag is slightly more honest (mean gap 0.0494, versus 0.0546 for CV).
Table~\ref{tab:oof-estimator-runtime} reports the runtimes also broken down by (family, learner) pair.
$k$-fold CV is the cheapest of the candidate estimators for every one of the twelve pairs: across the twenty (arm, family, learner) combinations for which a bagging or native-out-of-bag alternative exists, that alternative takes between 1.4 and 15.2 times as long as $k$-fold CV (median 6.0 times).
Overall, $k$-fold CV is thus both the cheapest and most honest out-of-fold train-score estimator.

\begin{table}[tbp]
\caption{
    Mean absolute gap between each out-of-fold estimator's accuracy and the held-out test accuracy, per (family, learner) pair.
    Smaller is more honest, and the best value per row is bold.
}
\label{tab:oof-estimator-ablation}
\centering
\begin{tabular}{llcccc}
\toprule
Algorithm & Learner & $k$-fold & Bag(15) & Bag(25) & Native OOB \\
\midrule
KG-MTP & Ridge & \textbf{0.0327} & 0.0347 & 0.0334 & -- \\
KG-MTP & Extra Trees & \textbf{0.0357} & -- & -- & 0.0390 \\
KG-MTP & Naive Bayes & 0.0546 & 0.0503 & \textbf{0.0494} & -- \\
QUANT & Ridge & \textbf{0.0360} & 0.0403 & 0.0383 & -- \\
QUANT & Extra Trees & \textbf{0.0383} & -- & -- & 0.0407 \\
QUANT & Naive Bayes & \textbf{0.0431} & 0.0535 & 0.0520 & -- \\
RDST & Ridge & \textbf{0.0328} & 0.0361 & 0.0347 & -- \\
RDST & Extra Trees & \textbf{0.0404} & -- & -- & 0.0424 \\
RDST & Naive Bayes & \textbf{0.0494} & 0.0529 & 0.0525 & -- \\
WEASEL~2.0 & Ridge & \textbf{0.0336} & 0.0371 & 0.0359 & -- \\
WEASEL~2.0 & Extra Trees & \textbf{0.0385} & -- & -- & 0.0567 \\
WEASEL~2.0 & Naive Bayes & \textbf{0.0369} & 0.0417 & 0.0411 & -- \\
\bottomrule
\end{tabular}
\end{table}

\begin{table}[tbp]
\caption{
    Mean wall-clock seconds per unit for each out-of-fold estimator, per (family, learner) pair (fastest in bold).
}
\label{tab:oof-estimator-runtime}
\centering
\begin{tabular}{llcccc}
\toprule
Algorithm & Learner & $k$-fold & Bag(15) & Bag(25) & Native OOB \\
\midrule
KG-MTP & Ridge & \textbf{44.1} & 171.1 & 282.5 & -- \\
KG-MTP & Extra Trees & \textbf{11.1} & -- & -- & 54.8 \\
KG-MTP & Naive Bayes & \textbf{25.8} & 217.8 & 360.0 & -- \\
QUANT & Ridge & \textbf{4.1} & 15.9 & 26.6 & -- \\
QUANT & Extra Trees & \textbf{5.8} & -- & -- & 8.0 \\
QUANT & Naive Bayes & \textbf{1.7} & 14.7 & 24.4 & -- \\
RDST & Ridge & \textbf{31.7} & 125.3 & 207.1 & -- \\
RDST & Extra Trees & \textbf{7.2} & -- & -- & 17.8 \\
RDST & Naive Bayes & \textbf{12.1} & 111.0 & 183.8 & -- \\
WEASEL~2.0 & Ridge & \textbf{31.7} & 119.6 & 197.9 & -- \\
WEASEL~2.0 & Extra Trees & \textbf{7.0} & -- & -- & 10.7 \\
WEASEL~2.0 & Naive Bayes & \textbf{15.0} & 52.5 & 86.9 & -- \\
\bottomrule
\end{tabular}
\end{table}

\subsubsection{TabICLv2's many-class handling}

TabICLv2, like TabPFN and TabICL before it, was pretrained primarily on data sets with comparatively few classes.
On data sets with many classes, an additional error-correcting-output-codes (ECOC) wrapper has been reported to improve TabICLv2's own multi-class accuracy, at roughly three times its plain inference cost \citep{quTabICLv2BetterFaster2026}.
Since TIGER's meta-feature matrix (up to twelve base learners times the number of classes, in columns) is structurally different from the generic tabular data sets on which this finding was originally reported, we measure both options on the development data sets with more than ten classes: TabICLv2's native many-class handling and its ECOC-wrapped variant.
Table~\ref{tab:multiclass-handling} reports the outcome, per data set, each averaged over five resamples.
Pooled across the six data sets, ECOC obtains a mean accuracy of 0.7374 against native handling's 0.7345, a difference of roughly 0.3 points.
This result is consistent with the small, ``slightly better but slower'' effect reported in the literature for generic tabular data \citep{quTabICLv2BetterFaster2026}, but at a cost of roughly five times as much wall-clock time on average (1124s versus 225s mean training-plus-inference runtime per data set).

\begin{table}[tbp]
\caption{
    TabICLv2's native many-class handling vs.\ its ECOC-wrapped variant, on the six development data sets with more than ten classes (runtimes in seconds, best accuracy per row in bold).
}
\label{tab:multiclass-handling}
\centering
\begin{tabular}{lccccc}
\toprule
& & \multicolumn{2}{c}{Accuracy} & \multicolumn{2}{c}{Runtimes (s)} \\
\cmidrule(lr){3-4} \cmidrule(lr){5-6}
Data set & Classes & Native & ECOC & Native & ECOC \\
\midrule
EOGHorizontalSignal & 12 & 0.814 & \textbf{0.818} & 132 & 886 \\
EOGVerticalSignal & 12 & \textbf{0.771} & 0.762 & 131 & 880 \\
FacesUCR & 14 & 0.957 & \textbf{0.962} & 361 & 1557 \\
GestureMidAirD3 & 26 & 0.506 & \textbf{0.511} & 150 & 972 \\
PigAirwayPressure & 52 & \textbf{0.548} & 0.526 & 291 & 1248 \\
PigCVP & 52 & 0.812 & \textbf{0.846} & 287 & 1199 \\
\midrule
Pooled mean & -- & 0.7345 & \textbf{0.7374} & 225 & 1124 \\
\bottomrule
\end{tabular}
\end{table}

\subsection{Additional analyses}\label{sec5subsec5}

We complement the ablation experiments with three additional analyses of TIGER's behavior: the accuracy headroom still available to a hypothetical oracle, each representation family's standalone predictive strength, and how the oracle headroom varies with data set characteristics. Appendix~\ref{secE} reports a further analysis, the calibration of the Ridge meta-features used by CAWPE, one of the five methods that TIGER's adaptive rule never selects in practice.

\subsubsection{Headroom of a per-data-set oracle}

We quantify how much accuracy would still be available to a hypothetical, perfectly-informed meta-classification rule that could pick, independently for every (data set, resample) row, whichever of the six candidate methods performs best.

Table~\ref{tab:oracle-headroom} reports this headroom, together with two further reference points.
Restricted to only TIGER's own two constituent methods (weighted hard majority vote and TabICLv2), a perfectly-informed oracle obtains a mean accuracy of 0.8869, only 0.31 points above TIGER's own deployed adaptive rule (0.8838).
This result indicates that the threshold $\tau$ already captures nearly all of the headroom available within these two methods alone.
The full, six-method oracle obtains 0.8872, only 0.0003 above the two-method oracle, that is essentially none of the remaining headroom left.

Unlike what a first look at the two-method oracle alone might suggest, widening the oracle from two candidates to all six barely moves the aggregate number, but not because logistic regression, LightGBM, CAWPE, and plain hard majority vote rarely win individually.
Table~\ref{tab:oracle-headroom-breakdown} shows that each of them is the outright best of all six candidates on a meaningful number of data sets (6, 10, 19, and 17 of 142, respectively, ties counted for every tied method).
Each win's margin over the two-method oracle is tiny, however (mean margin between 0.0002 and 0.0018 points, maximum at most 0.0103 points), which is why their combined contribution to the six-method oracle's aggregate accuracy is only the 0.0003 gap reported above.
A richer, still data-driven selection rule that also considered these four methods would accordingly be expected to close only a negligible fraction of TIGER's remaining 0.34-point gap to the six-method oracle, not the larger gap a comparison against the wrong reference point once suggested (see below).
For reference, always picking the single best base learner by cross-validated out-of-fold accuracy alone (no meta-classification stage at all, and no hindsight: this selection never uses test-set information) obtains only 0.8778, underscoring that TIGER's stacking approach itself, not just its adaptive rule between two meta-classifiers, is responsible for most of the accuracy gained over any individual base learner.

Table~\ref{tab:oracle-headroom}'s final row reports one further, more permissive upper bound: the single base learner (of TIGER's own twelve) that scores the highest on the held-out test set itself, rather than by cross-validated out-of-fold accuracy.
This is not an achievable selection rule, since it requires already knowing the test labels, and it is not one of the six candidate meta-classification methods, so it is excluded from the ``gap to six-method oracle'' column entirely.
We report it only because it is, perhaps surprisingly, higher than the six-method oracle itself (0.8922 versus 0.8872).
This indicates that none of the six candidate meta-classification methods is an oracle-perfect combiner of TIGER's twelve base learners: for a meaningful share of data sets, one specific base learner, credited with unrealistic hindsight knowledge of the test labels, outperforms every one of the six methods that combine information across all twelve.

\begin{table}[tbp]
\caption{
    Mean accuracy of TIGER's deployed rule against four reference points (red rows use test-set hindsight and cannot be deployed).
}
\label{tab:oracle-headroom}
\centering
\begin{tabular}{lcc}
\toprule
Method & Mean accuracy & Gap to six-method oracle \\
\midrule
Best single base learner (CV-picked) & 0.8778 & 0.0094 \\
TIGER's adaptive rule (deployed) & 0.8838 & 0.0034 \\
\rowcolor{hindsightrow} WHMV / TabICLv2 oracle (hindsight) & 0.8869 & 0.0003 \\
\rowcolor{hindsightrow} Six-method oracle (hindsight) & \textbf{0.8872} & 0.0000 \\
\rowcolor{testhindsightrow} Best single base learner (test hindsight) & 0.8922 & -- \\
\bottomrule
\end{tabular}
\end{table}

\begin{table}[tbp]
\caption{
    Per-method breakdown of the six-method oracle's gain over Table~\ref{tab:oracle-headroom}'s two-method oracle, for the four methods TIGER does not deploy.
    ``Data sets won'' counts the data sets where that method ties or leads all six candidates.
    ``Margin'' is its own accuracy minus the two-method oracle's, on the data sets it wins.
}
\label{tab:oracle-headroom-breakdown}
\centering
\begin{tabular}{lrrr}
\toprule
Method & Data sets won & Mean margin & Max margin \\
\midrule
CAWPE & 19 & 0.0013 & 0.0103 \\
Hard majority vote & 17 & 0.0005 & 0.0026 \\
LightGBM & 10 & 0.0002 & 0.0013 \\
Logistic regression & 6 & 0.0018 & 0.0082 \\
\bottomrule
\end{tabular}
\end{table}

\subsubsection{Each family's standalone predictive strength}

We also assess how well each of the four representation families performs entirely on its own, before any cross-family stacking.
For each family, we combine its own three base learners (Ridge, Extra Trees, and Naive Bayes, each on the cross-validated out-of-fold arm actually deployed throughout this paper) by a within-family hard majority vote, the same combination method TIGER's own meta-classification stage offers as one of its six candidate methods, here restricted to one family's own three base learners rather than all twelve.
We evaluate the result on the full 142-data-set benchmark, following the same per-data-set-then-across-data-sets aggregation as every other comparison in this paper.
Three data sets (DiatomSizeReduction, FiftyWords, and Phoneme) lack a cross-validated Ridge arm for every resample and every family (Section~\ref{sec5subsec2} discusses why) and are excluded from this comparison only, leaving 139.

Table~\ref{tab:family-standalone-strength} reports the outcome.
KG-MTP is clearly the strongest of the four on its own, with a mean accuracy of 0.866 and the highest accuracy of any family on 75 of the 139 data sets.
RDST follows at 0.858 (30 data sets won), then WEASEL~2.0 at 0.843 (27 data sets won), with QUANT the weakest of the four at 0.833 (21 data sets won).
This ordering is consistent with convolution-based representations' strong standalone performance already established in the wider time series classification literature, and with KG-MTP's otherwise prominent role in this paper.

\begin{table}[tbp]
\caption{
    Each representation family's standalone predictive strength before stacking, on 139 of 142 data sets (best per column in bold).
}
\label{tab:family-standalone-strength}
\centering
\begin{tabular}{lccc}
\toprule
Algorithm & Mean accuracy & Mean balanced accuracy & Data sets won \\
\midrule
KG-MTP & \textbf{0.8661} & \textbf{0.8437} & \textbf{75} \\
RDST & 0.8577 & 0.8301 & 30 \\
WEASEL~2.0 & 0.8432 & 0.8148 & 27 \\
QUANT & 0.8328 & 0.8077 & 21 \\
\bottomrule
\end{tabular}
\end{table}

\subsubsection{Meta-oracle headroom by data set characteristics}
Finally, we break down the six-method oracle headroom of Table~\ref{tab:oracle-headroom} by two data set characteristics: number of classes and training set size, again using the correctly-computed six-method oracle from that table rather than the best-single-base-learner diagnostic in its final row.
Tables~\ref{tab:oracle-headroom-by-nclasses} and~\ref{tab:oracle-headroom-by-ntrain} report the results, binned by quartile for training set size and by natural class-count ranges otherwise.

The gap to the six-method oracle grows with the number of classes, from 0.002 on the 50 binary data sets to 0.006 on the 24 data sets with more than ten classes.
This result is consistent with $\tau$ being tuned on a single scalar statistic (training samples per class) that does not itself encode class count, so the adaptive rule has less information to work with as the number of classes grows, even though its overall accuracy is naturally lower there regardless of method.
The breakdown by training set size shows no comparably clean trend: the gap is smallest for the largest data sets (0.001 on the 35 data sets with at least 467 training samples) and largest for the 218-to-466 range (0.007), with the two smaller bins in between (0.002 and 0.003).
This is broadly consistent with TIGER's adaptive rule already targeting training-set size directly via $\tau$, leaving number of classes as the more informative direction for any future, richer selection rule to add.

\begin{table}[tbp]
\caption{Meta-oracle headroom broken down by number of classes.}
\label{tab:oracle-headroom-by-nclasses}
\centering
\begin{tabular}{lcccc}
\toprule
Classes & Data sets & TIGER's adaptive rule & Six-method oracle & Gap \\
\midrule
2 (binary) & 50 & 0.9291 & 0.9307 & 0.0016 \\
3--5 & 38 & 0.8905 & 0.8930 & 0.0025 \\
6--10 & 30 & 0.8478 & 0.8532 & 0.0054 \\
11+ & 24 & 0.8241 & 0.8297 & 0.0057 \\
\bottomrule
\end{tabular}
\end{table}

\begin{table}[tbp]
\caption{Meta-oracle headroom broken down by training set size, binned at the quartiles.}
\label{tab:oracle-headroom-by-ntrain}
\centering
\begin{tabular}{lcccc}
\toprule
Training samples & Data sets & TIGER's adaptive rule & Six-method oracle & Gap \\
\midrule
$\leq$60 & 36 & 0.9447 & 0.9469 & 0.0022 \\
61--217 & 35 & 0.8347 & 0.8378 & 0.0031 \\
218--466 & 36 & 0.8273 & 0.8339 & 0.0066 \\
$\geq$467 & 35 & 0.9285 & 0.9299 & 0.0014 \\
\bottomrule
\end{tabular}
\end{table}

\subsection{Where TabICLv2 provides the largest individual gains}\label{sec5subsec6}

In this subsection, we complement the oracle-headroom analysis of Section~\ref{sec5subsec5} with a closer look at where TabICLv2 helps the most as a meta-classifier, to better understand the regime in which the adaptive rule's threshold $\tau$ operates.

Table~\ref{tab:tabicl-wins} lists the ten data sets on which TabICLv2 wins the most, in accuracy, relative to the WHMV/TabICLv2 oracle, mirroring the ten biggest losses already discussed in Section~\ref{sec5subsec1}.
Eight of these ten data sets have at least $\tau$ training samples per class, consistent with the adaptive rule's threshold: on the large majority of the data sets where TabICLv2 provides the largest gains, the rule would indeed select TabICLv2.
The two exceptions, BeetleFly (10 training samples per class) and AllGestureWiimoteZ (30 training samples per class), are both below $\tau$, meaning that the adaptive rule would select weighted hard majority vote instead and forgo these two specific gains.
Given that every one of TabICLv2's ten largest losses also falls below $\tau$, and by a considerably larger margin, this remains a favorable trade-off in aggregate, as the critical difference diagram of Figure~\ref{fig:cd-methods} confirms.

\begin{table}[tbp]
\caption{
    The ten data sets on which TabICLv2, used alone as the meta-classifier, gains the most accuracy relative to the WHMV/TabICLv2 oracle (checkmark: at or above $\tau = 60.83$).
}
\label{tab:tabicl-wins}
\centering
\begin{tabular}{lccc}
\toprule
Data set & Mean samples/class & TabICLv2 vs.\ oracle & $\geq \tau$ \\
\midrule
PhoneHeartbeatSound & 84.8 & +0.072 & \checkmark \\
EthanolLevel & 126.0 & +0.072 & \checkmark \\
SemgHandMovementCh2 & 75.0 & +0.061 & \checkmark \\
ScreenType & 125.0 & +0.044 & \checkmark \\
ChlorineConcentration & 155.7 & +0.042 & \checkmark \\
SemgHandSubjectCh2 & 90.0 & +0.041 & \checkmark \\
Tools & 62.0 & +0.035 & \checkmark \\
BeetleFly & 10.0 & +0.035 & \\
AllGestureWiimoteZ & 30.0 & +0.031 & \\
DistalPhalanxTW & 66.7 & +0.026 & \checkmark \\
\bottomrule
\end{tabular}
\end{table}

\section{Discarded design alternatives}\label{sec6}

Before converging on TIGER's stacking design, we explored an earlier architecture in which TabICLv2 was fit directly on each representation family's own extracted features, one fit per family, and the four resulting predictions were then combined by a CAWPE-style fused vote, rather than stacked into a single small meta-feature matrix and fed to one meta-classifier as TIGER does.
We investigated two variants of this design, primarily on the twenty-data-set development subset, and abandoned both.
We wanted to develop an algorithm that could be run on CPU only in reasonable time, while being start-of-the-art in terms of predictive performance.
Our observations made us conclude that this condition was unlikely to be met with both approaches.

\subsection{TabICLv2 directly on raw, per-family features}

The most direct version of this design fits TabICLv2 on each family's raw extracted features, with no reduction step at all.
This is very computationally expensive, especially without hardware acceleration.
TabICLv2 is pretrained on tables of only 2 to 100 columns, and its own documentation reports degraded performance well beyond that width, so any wider feature matrix must first be split into column groups of that size, fit independently, and combined, the same grouping mechanism our reduced-feature variant below also relies on.
The number of groups, and so the total compute, then scales with the family's raw feature count, and that count is large.
The convolution family's raw features alone can reach 50,000 to 60,000 columns on some development-subset data sets, and the other three families' raw outputs are comparably wide before any reduction.

This same tension, between a tabular foundation model's narrow column budget and a time-series feature extractor's naturally wide output, recurs throughout the concurrent literature that applies time-series-derived features directly to a pretrained tabular in-context-learning model, and every one of these papers reports capping or grouping its features for essentially the same reason.
RocketPFN \citep{orourkeRocketPFNAccurateTime2026} feeds ROCKET's own 20,000 raw features (10,000 kernels) to TabPFN~2.5 by distributing its kernels across several groups of at most 2,000 features each, TabPFN~2.5's own stated per-forward-pass limit, a grouping mechanism essentially identical to the one we use here.
MASHT \citep{cuppersInContextTimeSeries2026} caps its combined Hydra and MultiROCKET feature budget adaptively by data-set size, from 10,000 features for data sets under 1,000 instances down to only 200 features for the largest ones, rather than using one fixed budget throughout.
TS2TabPFN \citep{dacostamerlinTS2TabPFNTimeSeries2026} reports a tighter limit still, restricting individual TabPFN forward passes to 500 features and falling back to an ensemble of feature-subsampled passes above that, and states explicitly that such transformer-based tabular foundation models must load their entire context into GPU memory for a single forward pass, so memory scales quadratically with context size, a constraint that led its authors to exclude data sets with especially long series or large training sets for exceeding available GPU memory outright.
We reached both of these conclusions independently and empirically.

\subsection{TabICLv2 on reduced or selected per-family features}

\begin{figure}[tbp]
    \centering
    \includegraphics[width=0.85\textwidth]{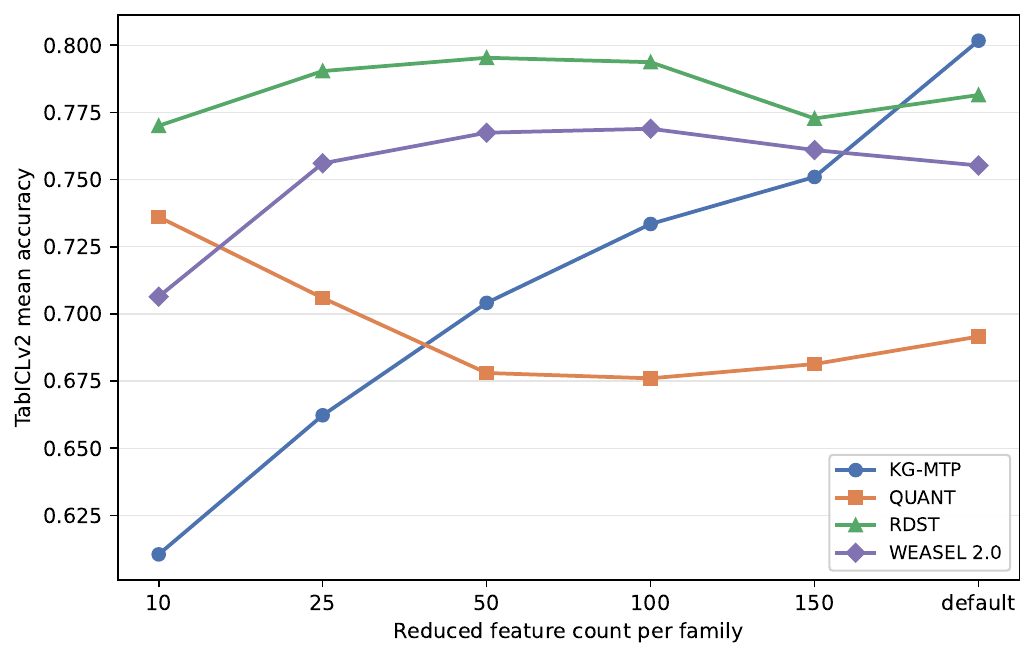}
    \caption{
        TabICLv2's mean accuracy per family as a function of reduced feature count.
        ``default'' is each family's own unconstrained reduction.
    }
    \label{fig:discarded-design-feature-counts}
\end{figure}

Since raw features were prohibitively wide, we also tried reducing each family's raw feature matrix before fitting TabICLv2 on it, again one fit per family: a partial least squares discriminant projection \citep{barkerPartialLeastSquares2003} for the interval and shapelet families, a detached-ROCKET-style feature selector \citep{uribarriDetachROCKETSequential2024} for the convolution family, and, for the dictionary family, a chi-squared association filter reusing WEASEL 2.0's own native dictionary-term-selection step \citep{schaferWEASEL20Random2023} followed by truncated (randomized) SVD compression \citep{halkoFindingStructureRandomness2011}.
We evaluated this design on the twenty-data-set development subset over 5 resamples, at feature counts of 10, 25, 50, 100, and 150 components per family, as well as each family's own unconstrained default reduction.
Figure~\ref{fig:discarded-design-feature-counts} shows TabICLv2's own mean accuracy across this grid, per family.

Two problems emerged, and neither pointed toward simply picking a single feature count and moving on.
First, as Figure~\ref{fig:discarded-design-feature-counts} shows, more features did not reliably mean better accuracy.
The convolution family kept improving across the entire tested range with no sign of plateauing, from a mean accuracy around 0.61 at 10 features to around 0.80 at its own unconstrained default (which can reach several hundred features).
The interval family showed the opposite pattern, peaking at just 10 components (around 0.74) and getting markedly worse by 100 components (around 0.68) before partially recovering at its own, much narrower default.
The dictionary and shapelet families fell in between, rising quickly and then flattening once past roughly 50 to 100 components.
A closely related, earlier investigation that pushed the number of PLS components as high as 200 found that held-out log-loss got monotonically worse throughout for both the interval and shapelet families, and that interval accuracy dropped substantially on two of five real data sets tested, confirming that TabICLv2 was overfitting on the added components rather than merely failing to benefit from them.
There was consequently no single feature count safe to recommend across all four families.

Second, even the best-performing configuration that we tested remained far too slow to run at the scale of the full evaluation benchmark.
A feature-selection variant that kept up to 1,000 of each family's most informative raw columns, rather than a PLS or SVD projection, did give TabICLv2 a consistent accuracy edge over that same family's own original classifier (Ridge for KG-MTP, RDST, and WEASEL~2.0, and Extra Trees for QUANT) fit on the same reduced features, across all four families, of roughly 3.5 to 4.4 accuracy points (Table~\ref{tab:discarded-design-wide-selection}).
Selection itself was not free for that cheap classifier either.
Fit instead on the family's full, unreduced raw features, it reached higher accuracy still for convolution and dictionary (0.824 and 0.775, respectively, vs.\ 0.779 and 0.733 after selection) and lost about a point or less for interval and shapelet (0.792 vs.\ 0.779, and 0.803 vs.\ 0.797, respectively).
So feature selection, not only TabICLv2's downstream fit, cost real accuracy for two of the four families.
Table~\ref{tab:discarded-design-wide-selection} also reports each design's wall-clock runtime: for every data set we average its own 5 resamples' elapsed time, end to end including feature selection's own cost wherever selection is used, and sum those 20 per-data-set averages, giving the total cost of running each design once across the whole development subset.
TabICLv2's own edge came at a cost of roughly 20 hours per family this way, against well under two minutes, summed the same way, for the cheap classifier at either feature budget.
Scaled to the full 142-data-set, 30-resample benchmark across four families, that cost would run into thousands of CPU-hours, dwarfing every other component of TIGER's own pipeline by orders of magnitude, for a design that, per the first problem above, still had no reliable answer for how many features to keep in the first place.

\begin{table}[tbp]
\centering
\sbox{\resizedtablebox}{\resizebox{\linewidth}{!}{
\begin{tabular}{lcccccc}
\toprule
 & \multicolumn{2}{c}{Full features} & \multicolumn{2}{c}{Wide-selected, own classifier} & \multicolumn{2}{c}{Wide-selected, TabICLv2} \\
\cmidrule(lr){2-3} \cmidrule(lr){4-5} \cmidrule(lr){6-7}
Algorithm & Accuracy & Runtime (min) & Accuracy & Runtime (min) & Accuracy & Runtime (min) \\
\midrule
KG-MTP & 0.824 & 0.53 & 0.779 & 0.27 & 0.815 & 1203.9 \\
QUANT & 0.792 & 0.27 & 0.779 & 0.22 & 0.815 & 1188.9 \\
RDST & 0.803 & 0.29 & 0.797 & 0.15 & 0.832 & 1198.6 \\
WEASEL~2.0 & 0.775 & 0.27 & 0.733 & 1.14 & 0.776 & 1203.7 \\
\bottomrule
\end{tabular}
}}
\caption{
    The discarded ``wide selection'' design: each family's original classifier on full vs.\ wide-selected raw features, against TabICLv2 on the selected features.
    Accuracy is the mean over the 20 data sets' 5 resamples each.
    Runtime (minutes) is the sum over the 20 datasets, averaged over the 5 resamples.
}
\label{tab:discarded-design-wide-selection}
\usebox{\resizedtablebox}
\end{table}

\begin{figure}[tbp]
    \centering
    \includegraphics[width=0.85\textwidth]{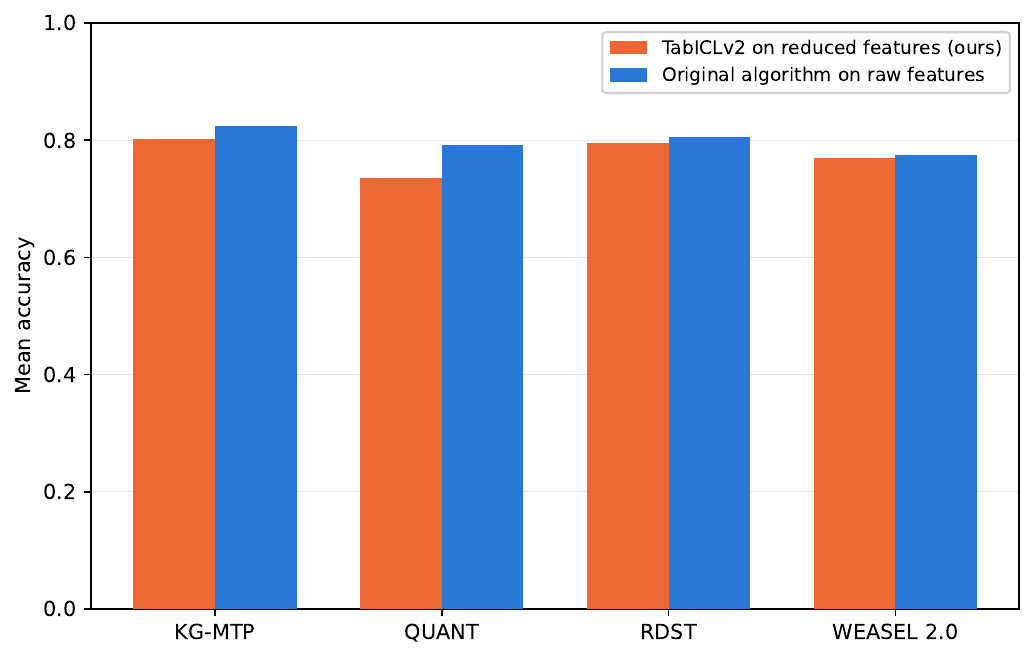}
    \caption{
        TabICLv2 at its best reduced-feature tag vs.\ each family's original algorithm on raw features.
    }
    \label{fig:discarded-design-comparison-with-originals}
\end{figure}

Even restricted to each family's single best-performing reduced-feature tag from Figure~\ref{fig:discarded-design-feature-counts}, TabICLv2 did not close the gap to that family's own original algorithm fit on raw features.
Figure~\ref{fig:discarded-design-comparison-with-originals} compares, per family, TabICLv2's mean accuracy at its best tag against the corresponding original algorithm's own mean accuracy, on the same twenty-data-set development subset.
TabICLv2 on reduced features trailed the original algorithm in every one of the four families, significantly so in three of them (convolution $-0.023$, $p = 2.7 \times 10^{-2}$, interval $-0.055$, $p = 3.2 \times 10^{-4}$, and shapelet $-0.010$, $p = 3.3 \times 10^{-2}$), with only the dictionary family a statistical toss-up ($-0.005$, $p = 0.26$).
This reinforced that feature reduction itself, rather than the downstream classifier, was the main source of the accuracy lost relative to each family's own specialized algorithm.

We ultimately abandoned this entire family of designs, TabICLv2 fit independently per representation family, on either raw or reduced features, and combined afterward by a CAWPE-style rule, in favor of TIGER's stacking design.
Stacking sidesteps both problems at once.
The meta-feature matrix TabICLv2 ultimately sees has a small, fixed number of columns, one score per (base learner, class) pair, regardless of how many raw features any individual family originally extracted, so neither the feature-count instability observed above nor the runtime blowup with feature width or data-set size can arise, and TabICLv2 is fit only once per data set rather than once per family.

\section{Conclusion}\label{sec7}

We have presented TIGER, a novel time series classification algorithm that stacks twelve base learners, obtained by applying a shared portfolio of three general-purpose classifiers (Ridge, Extra Trees, and Naive Bayes) to four representation families (convolution, interval, shapelet, and dictionary), and combines their predictions using an adaptive meta-classification rule that chooses, for each data set, between a weighted hard majority vote and a tabular foundation model used in-context (TabICLv2), based on the number of training samples available per class.
The evaluation of TIGER on a 142-data-set benchmark from the UCR archive has shown that it significantly outperforms current state-of-the-art time series classification algorithms, including HIVE-COTE~2.0, on classification accuracy, balanced accuracy, and F1-score.
The same conclusion remains valid on the development-subset-free 122-data-set view.
TIGER is also much faster than HIVE-COTE~2.0: over the 142-data-set benchmark, TIGER is about 7.2 times faster than HIVE-COTE~2.0 in terms of training-plus-inference runtime.
The adaptive meta-classification rule that chooses between a weighted hard majority vote and TabICLv2 was shown to be better than each method alone.
We have also shown that this adaptive rule, whose single hyperparameter was derived using only a twenty-data-set development subset, generalizes well to the full evaluation benchmark, and that its behavior is consistent with an intuitive explanation: TabICLv2, an in-context-learning model, needs a non-trivial number of labeled examples per class in its context to outperform a simple, fixed weighted-voting rule.

Our work still has several limitations.
First, TIGER depends on TabICLv2, an external, actively developed tabular foundation model.
Unlike TIGER's other components, its predictions are not obtained from a model trained end-to-end on time series data, and its own pretraining data and any of its own limitations are inherited by TIGER whenever the adaptive rule selects it.
Second, the adaptive rule itself is a single scalar threshold on one summary statistic (training samples per class).
Section~\ref{sec5subsec5} shows that a hypothetical oracle allowed to choose among all six candidate meta-classification methods, rather than only the two selected by TIGER's rule, still outperforms TIGER by roughly 0.3 accuracy points on average, but that this gap barely narrows at all (well under 0.1 points) when the oracle is restricted from six candidates down to only TIGER's own two, indicating that a richer selection rule considering the other four candidates would close only a small fraction of what headroom remains.
The more promising direction, per the same section's breakdown by data set characteristics, is a rule that also accounts for the number of classes, where the gap is largest.
Third, Section~\ref{sec5subsec2} identifies three data sets, each with a class that has only a single training instance, for which the adaptive-fold-count cross-validation underlying TIGER's base learners cannot guarantee that every class appears in every fold's training data.
The current pipeline conservatively discards Ridge's out-of-fold scores for the affected cells, while Extra Trees and Naive Bayes silently substitute zero for the same cells instead, so TIGER runs with eight of its twelve base learners, rather than the full twelve, on exactly these three data sets.
Treating this missing information consistently across all three base learners, rather than discarding it for one and silently filling it for the other two, is left to future work.
Fourth, we have not investigated whether all three base classifiers in TIGER's shared portfolio (Ridge, Extra Trees, and Naive Bayes) are necessary.
A reduced portfolio, for instance Ridge and Extra Trees alone, might match TIGER's accuracy at a lower computational cost.
We leave this investigation to future work.
Finally, TIGER has only been evaluated on univariate time series classification.
Extending it to multivariate time series, and to other tasks such as extrinsic regression, is also left to future work.

In conclusion, TIGER is a highly accurate time series classification algorithm that departs from the bespoke, per-family classifier design of prior ensemble methods such as HIVE-COTE~2.0, in favor of a shared portfolio of general-purpose base classifiers and a meta-classification stage that adapts, using a tabular foundation model where appropriate, to how much labeled data is available.
We believe that TIGER is a valuable addition to the list of time series classification algorithms, and that the use of tabular foundation models as an adaptive meta-classification layer is a promising direction for future ensemble designs, both in time series classification and more broadly.

\section{Declarations}

All the data sets used in this study are publicly available from the UCR Time Series Archive \citep{dauUCRTimeSeries2019e}.
We would like to thank Professor Eamonn Keogh and all the people who have contributed to this archive.
The data sets can be downloaded in several ways, either manually from the website\footnote{\url{https://timeseriesclassification.com/dataset.php}} or automatically using libraries such as the \emph{aeon} \citep{middlehurstAeonPythonToolkit2024}, \emph{tslearn} \citep{tavenardTslearnMachineLearning2020}, or \emph{pyts} \citep{faouziPytsPythonPackage2020a} Python packages.
The whole source code supporting this study is publicly available on a GitHub repository,\footnote{\url{https://github.com/johannfaouzi/TIGER}} with detailed instructions to reproduce all the experiments.
The results are also directly available, so that other researchers can easily compare their algorithms with ours.
The repository is under the BSD 3-Clause License, making it reusable by other researchers.

Generative artificial intelligence has been used in this study, both for generating code and analyzing the results.
Nonetheless, the authors agree to be accountable for all the materials associated with this study (this manuscript and the provided public GitHub repository).

No funding was received for conducting this study.
The authors have no relevant financial or non-financial interests to disclose.
The authors have no conflicts of interest to declare that are relevant to the content of this article.
The authors certify that they have no affiliations with or involvement in any organization or entity with any financial interest or non-financial interest in the matter or materials discussed in this manuscript.
The authors have no financial or proprietary interests in any material discussed in this article.

\bibliography{sn-bibliography}

\begin{thebibliography}{}
\renewcommand{\doi}[1]{\url{https://doi.org/#1}}
\bibcommenthead

\bibitem [\protect \citeauthoryear {%
Agrawal%
, Kumar%
, Pandey%
\BCBL {}\ \BBA {} Khan%
}{%
Agrawal%
\ \protect \BOthers {.}}{%
{\protect \APACyear {2012}}%
}]{%
agrawal2012application}
\APACinsertmetastar {%
agrawal2012application}%
\begin{APACrefauthors}%
Agrawal, A.%
, Kumar, V.%
, Pandey, A.%
\BCBL {} Khan, I.%
\end{APACrefauthors}%
\unskip\
\newblock
\APACrefYearMonthDay{2012}{}{}.
\newblock
{\BBOQ}\APACrefatitle {An application of time series analysis for weather forecasting} {An application of time series analysis for weather forecasting}.{\BBCQ}
\newblock
\APACjournalVolNumPages{International Journal of Engineering Research and Applications}{2}{2}{974--980,}
\newblock

\newblock

\PrintBackRefs{\CurrentBib}

\bibitem [\protect \citeauthoryear {%
Amei%
, Fu%
\BCBL {}\ \BBA {} Ho%
}{%
Amei%
\ \protect \BOthers {.}}{%
{\protect \APACyear {2012}}%
}]{%
amei2012time}
\APACinsertmetastar {%
amei2012time}%
\begin{APACrefauthors}%
Amei, A.%
, Fu, W.%
\BCBL {} Ho, C\BHBI H.%
\end{APACrefauthors}%
\unskip\
\newblock
\APACrefYearMonthDay{2012}{}{}.
\newblock
{\BBOQ}\APACrefatitle {Time series analysis for predicting the occurrences of large scale earthquakes} {Time series analysis for predicting the occurrences of large scale earthquakes}.{\BBCQ}
\newblock
\APACjournalVolNumPages{International Journal of Applied Science and Technology}{2}{7}{,}
\newblock

\newblock

\PrintBackRefs{\CurrentBib}

\bibitem [\protect \citeauthoryear {%
Bagnall%
, Flynn%
, Large%
, Lines%
\BCBL {}\ \BBA {} Middlehurst%
}{%
Bagnall%
\ \protect \BOthers {.}}{%
{\protect \APACyear {2020}}%
}]{%
bagnallUsagePerformance2020}
\APACinsertmetastar {%
bagnallUsagePerformance2020}%
\begin{APACrefauthors}%
Bagnall, A.%
, Flynn, M.%
, Large, J.%
, Lines, J.%
\BCBL {} Middlehurst, M.%
\end{APACrefauthors}%
\unskip\
\newblock
\APACrefYearMonthDay{2020}{}{}.
\newblock
{\BBOQ}\APACrefatitle {On the {{Usage}} and {{Performance}} of the {{Hierarchical Vote Collective}} of {{Transformation-Based Ensembles Version}} 1.0 ({{HIVE-COTE}} 1.0)} {On the {{Usage}} and {{Performance}} of the {{Hierarchical Vote Collective}} of {{Transformation-Based Ensembles Version}} 1.0 ({{HIVE-COTE}} 1.0)}.{\BBCQ}
\newblock
 \APACrefbtitle {Advanced {{Analytics}} and {{Learning}} on {{Temporal Data}}: {{5th ECML PKDD Workshop}}, {{AALTD}} 2020} {Advanced {{Analytics}} and {{Learning}} on {{Temporal Data}}: {{5th ECML PKDD Workshop}}, {{AALTD}} 2020}\ (\BPGS\ 3--18).
\newblock
\APACaddressPublisher{Cham}{Springer}.
\PrintBackRefs{\CurrentBib}

\bibitem [\protect \citeauthoryear {%
Bagnall%
, Lines%
, Hills%
\BCBL {}\ \BBA {} Bostrom%
}{%
Bagnall%
\ \protect \BOthers {.}}{%
{\protect \APACyear {2015}}%
}]{%
bagnallTimeSeriesClassification2015}
\APACinsertmetastar {%
bagnallTimeSeriesClassification2015}%
\begin{APACrefauthors}%
Bagnall, A.%
, Lines, J.%
, Hills, J.%
\BCBL {} Bostrom, A.%
\end{APACrefauthors}%
\unskip\
\newblock
\APACrefYearMonthDay{2015}{}{}.
\newblock
{\BBOQ}\APACrefatitle {Time-{{Series Classification}} with {{COTE}}: {{The Collective}} of {{Transformation-Based Ensembles}}} {Time-{{Series Classification}} with {{COTE}}: {{The Collective}} of {{Transformation-Based Ensembles}}}.{\BBCQ}
\newblock
\APACjournalVolNumPages{IEEE Transactions on Knowledge and Data Engineering}{27}{9}{2522--2535,}
\newblock
\begin{APACrefDOI} \doi{10.1109/TKDE.2015.2416723} \end{APACrefDOI}
\newblock

\newblock

\PrintBackRefs{\CurrentBib}

\bibitem [\protect \citeauthoryear {%
Barker%
\ \BBA {} Rayens%
}{%
Barker%
\ \BBA {} Rayens%
}{%
{\protect \APACyear {2003}}%
}]{%
barkerPartialLeastSquares2003}
\APACinsertmetastar {%
barkerPartialLeastSquares2003}%
\begin{APACrefauthors}%
Barker, M.%
\BCBT {}\ \BBA {} Rayens, W.%
\end{APACrefauthors}%
\unskip\
\newblock
\APACrefYearMonthDay{2003}{}{}.
\newblock
{\BBOQ}\APACrefatitle {Partial Least Squares for Discrimination} {Partial least squares for discrimination}.{\BBCQ}
\newblock
\APACjournalVolNumPages{Journal of Chemometrics}{17}{3}{166--173,}
\newblock
\begin{APACrefDOI} \doi{10.1002/cem.785} \end{APACrefDOI}
\newblock

\newblock

\PrintBackRefs{\CurrentBib}

\bibitem [\protect \citeauthoryear {%
Benavoli%
, Corani%
\BCBL {}\ \BBA {} Mangili%
}{%
Benavoli%
\ \protect \BOthers {.}}{%
{\protect \APACyear {2016}}%
}]{%
benavoliShouldWeReally2016}
\APACinsertmetastar {%
benavoliShouldWeReally2016}%
\begin{APACrefauthors}%
Benavoli, A.%
, Corani, G.%
\BCBL {} Mangili, F.%
\end{APACrefauthors}%
\unskip\
\newblock
\APACrefYearMonthDay{2016}{}{}.
\newblock
{\BBOQ}\APACrefatitle {Should {{We Really Use Post-Hoc Tests Based}} on {{Mean-Ranks}}?} {Should {{We Really Use Post-Hoc Tests Based}} on {{Mean-Ranks}}?}{\BBCQ}
\newblock
\APACjournalVolNumPages{Journal of Machine Learning Research}{17}{5}{1--10,}
\newblock

\newblock

\PrintBackRefs{\CurrentBib}

\bibitem [\protect \citeauthoryear {%
Breiman%
}{%
Breiman%
}{%
{\protect \APACyear {2001}}%
}]{%
breimanRandomForests2001a}
\APACinsertmetastar {%
breimanRandomForests2001a}%
\begin{APACrefauthors}%
Breiman, L.%
\end{APACrefauthors}%
\unskip\
\newblock
\APACrefYearMonthDay{2001}{{\APACmonth{10}}}{}.
\newblock
{\BBOQ}\APACrefatitle {Random {{Forests}}} {Random {{Forests}}}.{\BBCQ}
\newblock
\APACjournalVolNumPages{Machine Learning}{45}{1}{5--32,}
\newblock
\begin{APACrefDOI} \doi{10.1023/A:1010933404324} \end{APACrefDOI}
\newblock

\newblock

\PrintBackRefs{\CurrentBib}

\bibitem [\protect \citeauthoryear {%
Cabello%
\ \BBA {} Kulik%
}{%
Cabello%
\ \BBA {} Kulik%
}{%
{\protect \APACyear {2025}}%
}]{%
cabelloPULSARAdvancingIntervalBased2025}
\APACinsertmetastar {%
cabelloPULSARAdvancingIntervalBased2025}%
\begin{APACrefauthors}%
Cabello, N.%
\BCBT {}\ \BBA {} Kulik, L.%
\end{APACrefauthors}%
\unskip\
\newblock
\APACrefYearMonthDay{2025}{{\APACmonth{11}}}{}.
\newblock
{\BBOQ}\APACrefatitle {{{PULSAR}}: {{Advancing Interval-Based Time Series Classification}} to {{State-of-the-Art Performance}}} {{{PULSAR}}: {{Advancing Interval-Based Time Series Classification}} to {{State-of-the-Art Performance}}}.{\BBCQ}
\newblock
 \APACrefbtitle {2025 {{IEEE International Conference}} on {{Data Mining}} ({{ICDM}}).} {2025 {{IEEE International Conference}} on {{Data Mining}} ({{ICDM}}).}
\PrintBackRefs{\CurrentBib}

\bibitem [\protect \citeauthoryear {%
C{\"u}ppers%
\ \BBA {} Vreeken%
}{%
C{\"u}ppers%
\ \BBA {} Vreeken%
}{%
{\protect \APACyear {2026}}%
}]{%
cuppersInContextTimeSeries2026}
\APACinsertmetastar {%
cuppersInContextTimeSeries2026}%
\begin{APACrefauthors}%
C{\"u}ppers, J.%
\BCBT {}\ \BBA {} Vreeken, J.%
\end{APACrefauthors}%
\unskip\
\newblock
\APACrefYearMonthDay{2026}{{\APACmonth{07}}}{}.
\newblock
\APACrefbtitle {In-{{Context Time Series Classification}} with {{Random Convolutional Features}}.} {In-{{Context Time Series Classification}} with {{Random Convolutional Features}}.}
\newblock
\APACrefnote{arXiv preprint}
\PrintBackRefs{\CurrentBib}

\bibitem [\protect \citeauthoryear {%
{da Costa Merlin}%
\ \BBA {} Silva%
}{%
{da Costa Merlin}%
\ \BBA {} Silva%
}{%
{\protect \APACyear {2026}}%
}]{%
dacostamerlinTS2TabPFNTimeSeries2026}
\APACinsertmetastar {%
dacostamerlinTS2TabPFNTimeSeries2026}%
\begin{APACrefauthors}%
{da Costa Merlin}, G.%
\BCBT {}\ \BBA {} Silva, D.F.%
\end{APACrefauthors}%
\unskip\
\newblock
\APACrefYearMonthDay{2026}{{\APACmonth{08}}}{}.
\newblock
\APACrefbtitle {{{TS2TabPFN}}: Time Series Classification and Extrinsic Regression through Feature Extraction and a Tabular Foundation Model.} {{{TS2TabPFN}}: Time series classification and extrinsic regression through feature extraction and a tabular foundation model.}
\newblock
\APACrefnote{arXiv preprint}
\PrintBackRefs{\CurrentBib}

\bibitem [\protect \citeauthoryear {%
Dau%
\ \protect \BOthers {.}}{%
Dau%
\ \protect \BOthers {.}}{%
{\protect \APACyear {2019}}%
}]{%
dauUCRTimeSeries2019e}
\APACinsertmetastar {%
dauUCRTimeSeries2019e}%
\begin{APACrefauthors}%
Dau, H.A.%
, Bagnall, A.%
, Kamgar, K.%
, Yeh, C\BHBI C.M.%
, Zhu, Y.%
, Gharghabi, S.%
\BDBL {}Keogh, E.%
\end{APACrefauthors}%
\unskip\
\newblock
\APACrefYearMonthDay{2019}{{\APACmonth{11}}}{}.
\newblock
{\BBOQ}\APACrefatitle {The {{UCR}} Time Series Archive} {The {{UCR}} time series archive}.{\BBCQ}
\newblock
\APACjournalVolNumPages{IEEE/CAA Journal of Automatica Sinica}{6}{6}{1293--1305,}
\newblock
\begin{APACrefDOI} \doi{10.1109/JAS.2019.1911747} \end{APACrefDOI}
\newblock

\newblock

\PrintBackRefs{\CurrentBib}

\bibitem [\protect \citeauthoryear {%
Dempster%
, Petitjean%
\BCBL {}\ \BBA {} Webb%
}{%
Dempster%
\ \protect \BOthers {.}}{%
{\protect \APACyear {2020}}%
}]{%
dempsterROCKETExceptionallyFast2020}
\APACinsertmetastar {%
dempsterROCKETExceptionallyFast2020}%
\begin{APACrefauthors}%
Dempster, A.%
, Petitjean, F.%
\BCBL {} Webb, G.I.%
\end{APACrefauthors}%
\unskip\
\newblock
\APACrefYearMonthDay{2020}{{\APACmonth{09}}}{}.
\newblock
{\BBOQ}\APACrefatitle {{{ROCKET}}: Exceptionally Fast and Accurate Time Series Classification Using Random Convolutional Kernels} {{{ROCKET}}: Exceptionally fast and accurate time series classification using random convolutional kernels}.{\BBCQ}
\newblock
\APACjournalVolNumPages{Data Mining and Knowledge Discovery}{34}{5}{1454--1495,}
\newblock
\begin{APACrefDOI} \doi{10.1007/s10618-020-00701-z} \end{APACrefDOI}
\newblock

\newblock

\PrintBackRefs{\CurrentBib}

\bibitem [\protect \citeauthoryear {%
Dempster%
, Schmidt%
\BCBL {}\ \BBA {} Webb%
}{%
Dempster%
\ \protect \BOthers {.}}{%
{\protect \APACyear {2021}}%
}]{%
dempsterMiniRocketVeryFast2021a}
\APACinsertmetastar {%
dempsterMiniRocketVeryFast2021a}%
\begin{APACrefauthors}%
Dempster, A.%
, Schmidt, D.F.%
\BCBL {} Webb, G.I.%
\end{APACrefauthors}%
\unskip\
\newblock
\APACrefYearMonthDay{2021}{{\APACmonth{08}}}{}.
\newblock
{\BBOQ}\APACrefatitle {{{MiniRocket}}: {{A Very Fast}} ({{Almost}}) {{Deterministic Transform}} for {{Time Series Classification}}} {{{MiniRocket}}: {{A Very Fast}} ({{Almost}}) {{Deterministic Transform}} for {{Time Series Classification}}}.{\BBCQ}
\newblock
 \APACrefbtitle {Proceedings of the 27th {{ACM SIGKDD Conference}} on {{Knowledge Discovery}} \& {{Data Mining}}} {Proceedings of the 27th {{ACM SIGKDD Conference}} on {{Knowledge Discovery}} \& {{Data Mining}}}\ (\BPGS\ 248--257).
\newblock
\APACaddressPublisher{New York, NY, USA}{Association for Computing Machinery}.
\PrintBackRefs{\CurrentBib}

\bibitem [\protect \citeauthoryear {%
Dempster%
, Schmidt%
\BCBL {}\ \BBA {} Webb%
}{%
Dempster%
\ \protect \BOthers {.}}{%
{\protect \APACyear {2023}}%
}]{%
dempsterHydraCompetingConvolutional2023}
\APACinsertmetastar {%
dempsterHydraCompetingConvolutional2023}%
\begin{APACrefauthors}%
Dempster, A.%
, Schmidt, D.F.%
\BCBL {} Webb, G.I.%
\end{APACrefauthors}%
\unskip\
\newblock
\APACrefYearMonthDay{2023}{{\APACmonth{09}}}{}.
\newblock
{\BBOQ}\APACrefatitle {Hydra: Competing Convolutional Kernels for Fast and Accurate Time Series Classification} {Hydra: Competing convolutional kernels for fast and accurate time series classification}.{\BBCQ}
\newblock
\APACjournalVolNumPages{Data Mining and Knowledge Discovery}{37}{5}{1779--1805,}
\newblock
\begin{APACrefDOI} \doi{10.1007/s10618-023-00939-3} \end{APACrefDOI}
\newblock

\newblock

\PrintBackRefs{\CurrentBib}

\bibitem [\protect \citeauthoryear {%
Dempster%
, Schmidt%
\BCBL {}\ \BBA {} Webb%
}{%
Dempster%
\ \protect \BOthers {.}}{%
{\protect \APACyear {2024}}%
}]{%
dempsterQuantMinimalistInterval2024}
\APACinsertmetastar {%
dempsterQuantMinimalistInterval2024}%
\begin{APACrefauthors}%
Dempster, A.%
, Schmidt, D.F.%
\BCBL {} Webb, G.I.%
\end{APACrefauthors}%
\unskip\
\newblock
\APACrefYearMonthDay{2024}{{\APACmonth{07}}}{}.
\newblock
{\BBOQ}\APACrefatitle {Quant: A Minimalist Interval Method for Time Series Classification} {Quant: A minimalist interval method for time series classification}.{\BBCQ}
\newblock
\APACjournalVolNumPages{Data Mining and Knowledge Discovery}{38}{4}{2377--2402,}
\newblock
\begin{APACrefDOI} \doi{10.1007/s10618-024-01036-9} \end{APACrefDOI}
\newblock

\newblock

\PrintBackRefs{\CurrentBib}

\bibitem [\protect \citeauthoryear {%
Dem{\v s}ar%
}{%
Dem{\v s}ar%
}{%
{\protect \APACyear {2006}}%
}]{%
demsarStatisticalComparisonsClassifiers2006a}
\APACinsertmetastar {%
demsarStatisticalComparisonsClassifiers2006a}%
\begin{APACrefauthors}%
Dem{\v s}ar, J.%
\end{APACrefauthors}%
\unskip\
\newblock
\APACrefYearMonthDay{2006}{}{}.
\newblock
{\BBOQ}\APACrefatitle {Statistical {{Comparisons}} of {{Classifiers}} over {{Multiple Data Sets}}} {Statistical {{Comparisons}} of {{Classifiers}} over {{Multiple Data Sets}}}.{\BBCQ}
\newblock
\APACjournalVolNumPages{Journal of Machine Learning Research}{7}{1}{1--30,}
\newblock

\newblock

\PrintBackRefs{\CurrentBib}

\bibitem [\protect \citeauthoryear {%
Deng%
, Runger%
, Tuv%
\BCBL {}\ \BBA {} Vladimir%
}{%
Deng%
\ \protect \BOthers {.}}{%
{\protect \APACyear {2013}}%
}]{%
dengTimeSeriesForest2013}
\APACinsertmetastar {%
dengTimeSeriesForest2013}%
\begin{APACrefauthors}%
Deng, H.%
, Runger, G.%
, Tuv, E.%
\BCBL {} Vladimir, M.%
\end{APACrefauthors}%
\unskip\
\newblock
\APACrefYearMonthDay{2013}{{\APACmonth{08}}}{}.
\newblock
{\BBOQ}\APACrefatitle {A Time Series Forest for Classification and Feature Extraction} {A time series forest for classification and feature extraction}.{\BBCQ}
\newblock
\APACjournalVolNumPages{Information Sciences}{239}{}{142--153,}
\newblock
\begin{APACrefDOI} \doi{10.1016/j.ins.2013.02.030} \end{APACrefDOI}
\newblock

\newblock

\PrintBackRefs{\CurrentBib}

\bibitem [\protect \citeauthoryear {%
Faouzi%
}{%
Faouzi%
}{%
{\protect \APACyear {2026}}%
}]{%
faouzi2026momentquantminimalistintervalmethod}
\APACinsertmetastar {%
faouzi2026momentquantminimalistintervalmethod}%
\begin{APACrefauthors}%
Faouzi, J.%
\end{APACrefauthors}%
\unskip\
\newblock
\APACrefYearMonthDay{2026}{}{}.
\newblock
\APACrefbtitle {MomentQuant: an even more minimalist interval method with linear time complexity for time series classification.} {Momentquant: an even more minimalist interval method with linear time complexity for time series classification.}
\newblock
\begin{APACrefURL} {https://arxiv.org/abs/2609.05136} \end{APACrefURL}
\PrintBackRefs{\CurrentBib}

\bibitem [\protect \citeauthoryear {%
Faouzi%
\ \BBA {} Janati%
}{%
Faouzi%
\ \BBA {} Janati%
}{%
{\protect \APACyear {2020}}%
}]{%
faouziPytsPythonPackage2020a}
\APACinsertmetastar {%
faouziPytsPythonPackage2020a}%
\begin{APACrefauthors}%
Faouzi, J.%
\BCBT {}\ \BBA {} Janati, H.%
\end{APACrefauthors}%
\unskip\
\newblock
\APACrefYearMonthDay{2020}{}{}.
\newblock
{\BBOQ}\APACrefatitle {Pyts: {{A Python Package}} for {{Time Series Classification}}} {Pyts: {{A Python Package}} for {{Time Series Classification}}}.{\BBCQ}
\newblock
\APACjournalVolNumPages{Journal of Machine Learning Research}{21}{46}{1--6,}
\newblock

\newblock

\PrintBackRefs{\CurrentBib}

\bibitem [\protect \citeauthoryear {%
Feofanov%
\ \protect \BOthers {.}}{%
Feofanov%
\ \protect \BOthers {.}}{%
{\protect \APACyear {2025}}%
}]{%
feofanovMantisLightweightFoundation2025}
\APACinsertmetastar {%
feofanovMantisLightweightFoundation2025}%
\begin{APACrefauthors}%
Feofanov, V.%
, Wen, S.%
, Xie, S.%
, Roschmann, S.%
, Alonso, M.%
, Guo, H.%
\BDBL {}Redko, I.%
\end{APACrefauthors}%
\unskip\
\newblock
\APACrefYearMonthDay{2025}{{\APACmonth{02}}}{}.
\newblock
\APACrefbtitle {Mantis: Lightweight Foundation Model for Time Series Classification.} {Mantis: Lightweight foundation model for time series classification.}
\newblock
\APACrefnote{arXiv preprint}
\PrintBackRefs{\CurrentBib}

\bibitem [\protect \citeauthoryear {%
Garc{\'i}a%
\ \BBA {} Herrera%
}{%
Garc{\'i}a%
\ \BBA {} Herrera%
}{%
{\protect \APACyear {2008}}%
}]{%
garciaExtensionStatisticalComparisons2008}
\APACinsertmetastar {%
garciaExtensionStatisticalComparisons2008}%
\begin{APACrefauthors}%
Garc{\'i}a, S.%
\BCBT {}\ \BBA {} Herrera, F.%
\end{APACrefauthors}%
\unskip\
\newblock
\APACrefYearMonthDay{2008}{}{}.
\newblock
{\BBOQ}\APACrefatitle {An {{Extension}} on ``{{Statistical Comparisons}} of {{Classifiers}} over {{Multiple Data Sets}}'' for All {{Pairwise Comparisons}}} {An {{Extension}} on ``{{Statistical Comparisons}} of {{Classifiers}} over {{Multiple Data Sets}}'' for all {{Pairwise Comparisons}}}.{\BBCQ}
\newblock
\APACjournalVolNumPages{Journal of Machine Learning Research}{9}{89}{2677--2694,}
\newblock

\newblock

\PrintBackRefs{\CurrentBib}

\bibitem [\protect \citeauthoryear {%
Geurts%
, Ernst%
\BCBL {}\ \BBA {} Wehenkel%
}{%
Geurts%
\ \protect \BOthers {.}}{%
{\protect \APACyear {2006}}%
}]{%
geurtsExtremelyRandomizedTrees2006a}
\APACinsertmetastar {%
geurtsExtremelyRandomizedTrees2006a}%
\begin{APACrefauthors}%
Geurts, P.%
, Ernst, D.%
\BCBL {} Wehenkel, L.%
\end{APACrefauthors}%
\unskip\
\newblock
\APACrefYearMonthDay{2006}{{\APACmonth{04}}}{}.
\newblock
{\BBOQ}\APACrefatitle {Extremely Randomized Trees} {Extremely randomized trees}.{\BBCQ}
\newblock
\APACjournalVolNumPages{Machine Learning}{63}{1}{3--42,}
\newblock
\begin{APACrefDOI} \doi{10.1007/s10994-006-6226-1} \end{APACrefDOI}
\newblock

\newblock

\PrintBackRefs{\CurrentBib}

\bibitem [\protect \citeauthoryear {%
Goswami%
\ \protect \BOthers {.}}{%
Goswami%
\ \protect \BOthers {.}}{%
{\protect \APACyear {2024}}%
}]{%
goswamiMOMENTFamilyOpen2024}
\APACinsertmetastar {%
goswamiMOMENTFamilyOpen2024}%
\begin{APACrefauthors}%
Goswami, M.%
, Szafer, K.%
, Choudhry, A.%
, Cai, Y.%
, Li, S.%
\BCBL {} Dubrawski, A.%
\end{APACrefauthors}%
\unskip\
\newblock
\APACrefYearMonthDay{2024}{}{}.
\newblock
{\BBOQ}\APACrefatitle {{{MOMENT}}: A Family of Open Time-Series Foundation Models} {{{MOMENT}}: A family of open time-series foundation models}.{\BBCQ}
\newblock
 \APACrefbtitle {Proceedings of the 41st International Conference on Machine Learning} {Proceedings of the 41st international conference on machine learning}\ (\BVOL~235).
\PrintBackRefs{\CurrentBib}

\bibitem [\protect \citeauthoryear {%
Grinsztajn%
\ \protect \BOthers {.}}{%
Grinsztajn%
\ \protect \BOthers {.}}{%
{\protect \APACyear {2025}}%
}]{%
grinsztajnTabPFN25AdvancingState2025}
\APACinsertmetastar {%
grinsztajnTabPFN25AdvancingState2025}%
\begin{APACrefauthors}%
Grinsztajn, L.%
, Fl{\"o}ge, K.%
, Key, O.%
, Birkel, F.%
, Jund, P.%
, Roof, B.%
\BDBL {}Hutter, F.%
\end{APACrefauthors}%
\unskip\
\newblock
\APACrefYearMonthDay{2025}{{\APACmonth{11}}}{}.
\newblock
\APACrefbtitle {{{TabPFN-2.5}}: Advancing the State of the Art in Tabular Foundation Models.} {{{TabPFN-2.5}}: Advancing the state of the art in tabular foundation models.}
\newblock
\APACrefnote{arXiv preprint}
\PrintBackRefs{\CurrentBib}

\bibitem [\protect \citeauthoryear {%
Guillaume%
, Vrain%
\BCBL {}\ \BBA {} Elloumi%
}{%
Guillaume%
\ \protect \BOthers {.}}{%
{\protect \APACyear {2022}}%
}]{%
guillaumeRandomDilatedShapelet2022}
\APACinsertmetastar {%
guillaumeRandomDilatedShapelet2022}%
\begin{APACrefauthors}%
Guillaume, A.%
, Vrain, C.%
\BCBL {} Elloumi, W.%
\end{APACrefauthors}%
\unskip\
\newblock
\APACrefYearMonthDay{2022}{{\APACmonth{06}}}{}.
\newblock
{\BBOQ}\APACrefatitle {Random {{Dilated Shapelet Transform}}: {{A New Approach}} for~{{Time Series Shapelets}}} {Random {{Dilated Shapelet Transform}}: {{A New Approach}} for~{{Time Series Shapelets}}}.{\BBCQ}
\newblock
 \APACrefbtitle {Pattern {{Recognition}} and {{Artificial Intelligence}}: {{Third International Conference}}, {{ICPRAI}} 2022, {{Paris}}, {{France}}, {{June}} 1--3, 2022, {{Proceedings}}, {{Part I}}} {Pattern {{Recognition}} and {{Artificial Intelligence}}: {{Third International Conference}}, {{ICPRAI}} 2022, {{Paris}}, {{France}}, {{June}} 1--3, 2022, {{Proceedings}}, {{Part I}}}\ (\BPGS\ 653--664).
\newblock
\APACaddressPublisher{Berlin, Heidelberg}{Springer-Verlag}.
\PrintBackRefs{\CurrentBib}

\bibitem [\protect \citeauthoryear {%
Halko%
, Martinsson%
\BCBL {}\ \BBA {} Tropp%
}{%
Halko%
\ \protect \BOthers {.}}{%
{\protect \APACyear {2011}}%
}]{%
halkoFindingStructureRandomness2011}
\APACinsertmetastar {%
halkoFindingStructureRandomness2011}%
\begin{APACrefauthors}%
Halko, N.%
, Martinsson, P\BHBI G.%
\BCBL {} Tropp, J.A.%
\end{APACrefauthors}%
\unskip\
\newblock
\APACrefYearMonthDay{2011}{}{}.
\newblock
{\BBOQ}\APACrefatitle {Finding {{Structure}} with {{Randomness}}: Probabilistic Algorithms for Constructing Approximate Matrix Decompositions} {Finding {{Structure}} with {{Randomness}}: Probabilistic algorithms for constructing approximate matrix decompositions}.{\BBCQ}
\newblock
\APACjournalVolNumPages{SIAM Review}{53}{2}{217--288,}
\newblock
\begin{APACrefDOI} \doi{10.1137/090771806} \end{APACrefDOI}
\newblock

\newblock

\PrintBackRefs{\CurrentBib}

\bibitem [\protect \citeauthoryear {%
Harris%
\ \protect \BOthers {.}}{%
Harris%
\ \protect \BOthers {.}}{%
{\protect \APACyear {2020}}%
}]{%
harrisArrayProgrammingNumPy2020}
\APACinsertmetastar {%
harrisArrayProgrammingNumPy2020}%
\begin{APACrefauthors}%
Harris, C.R.%
, Millman, K.J.%
, {van der Walt}, S.J.%
, Gommers, R.%
, Virtanen, P.%
, Cournapeau, D.%
\BDBL {}Oliphant, T.E.%
\end{APACrefauthors}%
\unskip\
\newblock
\APACrefYearMonthDay{2020}{{\APACmonth{09}}}{}.
\newblock
{\BBOQ}\APACrefatitle {Array Programming with {{NumPy}}} {Array programming with {{NumPy}}}.{\BBCQ}
\newblock
\APACjournalVolNumPages{Nature}{585}{7825}{357--362,}
\newblock
\begin{APACrefDOI} \doi{10.1038/s41586-020-2649-2} \end{APACrefDOI}
\newblock

\newblock

\PrintBackRefs{\CurrentBib}

\bibitem [\protect \citeauthoryear {%
Hills%
, Lines%
, Baranauskas%
, Mapp%
\BCBL {}\ \BBA {} Bagnall%
}{%
Hills%
\ \protect \BOthers {.}}{%
{\protect \APACyear {2014}}%
}]{%
hillsClassificationTimeSeries2014a}
\APACinsertmetastar {%
hillsClassificationTimeSeries2014a}%
\begin{APACrefauthors}%
Hills, J.%
, Lines, J.%
, Baranauskas, E.%
, Mapp, J.%
\BCBL {} Bagnall, A.%
\end{APACrefauthors}%
\unskip\
\newblock
\APACrefYearMonthDay{2014}{{\APACmonth{07}}}{}.
\newblock
{\BBOQ}\APACrefatitle {Classification of Time Series by Shapelet Transformation} {Classification of time series by shapelet transformation}.{\BBCQ}
\newblock
\APACjournalVolNumPages{Data Min. Knowl. Discov.}{28}{4}{851--881,}
\newblock
\begin{APACrefDOI} \doi{10.1007/s10618-013-0322-1} \end{APACrefDOI}
\newblock

\newblock

\PrintBackRefs{\CurrentBib}

\bibitem [\protect \citeauthoryear {%
Hoerl%
\ \BBA {} Kennard%
}{%
Hoerl%
\ \BBA {} Kennard%
}{%
{\protect \APACyear {1970}}%
{\protect \APACexlab {{\protect \BCnt {1}}}}}]{%
hoerlRidgeRegressionApplications1970}
\APACinsertmetastar {%
hoerlRidgeRegressionApplications1970}%
\begin{APACrefauthors}%
Hoerl, A.E.%
\BCBT {}\ \BBA {} Kennard, R.W.%
\end{APACrefauthors}%
\unskip\
\newblock
\APACrefYearMonthDay{1970{\protect \BCnt {1}}}{}{}.
\newblock
{\BBOQ}\APACrefatitle {Ridge {{Regression}}: {{Applications}} to {{Nonorthogonal Problems}}} {Ridge {{Regression}}: {{Applications}} to {{Nonorthogonal Problems}}}.{\BBCQ}
\newblock
\APACjournalVolNumPages{Technometrics}{12}{1}{69--82,}
\newblock
\begin{APACrefDOI} \doi{10.2307/1267352} \end{APACrefDOI}
\newblock
{\href{https://arxiv.org/abs/1267352}{{1267352}}}
\newblock

\PrintBackRefs{\CurrentBib}

\bibitem [\protect \citeauthoryear {%
Hoerl%
\ \BBA {} Kennard%
}{%
Hoerl%
\ \BBA {} Kennard%
}{%
{\protect \APACyear {1970}}%
{\protect \APACexlab {{\protect \BCnt {2}}}}}]{%
hoerlRidgeRegressionBiased1970}
\APACinsertmetastar {%
hoerlRidgeRegressionBiased1970}%
\begin{APACrefauthors}%
Hoerl, A.E.%
\BCBT {}\ \BBA {} Kennard, R.W.%
\end{APACrefauthors}%
\unskip\
\newblock
\APACrefYearMonthDay{1970{\protect \BCnt {2}}}{}{}.
\newblock
{\BBOQ}\APACrefatitle {Ridge {{Regression}}: {{Biased Estimation}} for {{Nonorthogonal Problems}}} {Ridge {{Regression}}: {{Biased Estimation}} for {{Nonorthogonal Problems}}}.{\BBCQ}
\newblock
\APACjournalVolNumPages{Technometrics}{12}{1}{55--67,}
\newblock
\begin{APACrefDOI} \doi{10.2307/1267351} \end{APACrefDOI}
\newblock
{\href{https://arxiv.org/abs/1267351}{{1267351}}}
\newblock

\PrintBackRefs{\CurrentBib}

\bibitem [\protect \citeauthoryear {%
Hollmann%
, M{\"u}ller%
, Eggensperger%
\BCBL {}\ \BBA {} Hutter%
}{%
Hollmann%
\ \protect \BOthers {.}}{%
{\protect \APACyear {2023}}%
}]{%
hollmannTabPFNTransformerSolves2023}
\APACinsertmetastar {%
hollmannTabPFNTransformerSolves2023}%
\begin{APACrefauthors}%
Hollmann, N.%
, M{\"u}ller, S.%
, Eggensperger, K.%
\BCBL {} Hutter, F.%
\end{APACrefauthors}%
\unskip\
\newblock
\APACrefYearMonthDay{2023}{}{}.
\newblock
{\BBOQ}\APACrefatitle {{{TabPFN}}: A Transformer That Solves Small Tabular Classification Problems in a Second} {{{TabPFN}}: A transformer that solves small tabular classification problems in a second}.{\BBCQ}
\newblock
 \APACrefbtitle {International Conference on Learning Representations.} {International conference on learning representations.}
\PrintBackRefs{\CurrentBib}

\bibitem [\protect \citeauthoryear {%
Hollmann%
\ \protect \BOthers {.}}{%
Hollmann%
\ \protect \BOthers {.}}{%
{\protect \APACyear {2025}}%
}]{%
hollmannAccuratePredictionsSmall2025}
\APACinsertmetastar {%
hollmannAccuratePredictionsSmall2025}%
\begin{APACrefauthors}%
Hollmann, N.%
, M{\"u}ller, S.%
, Purucker, L.%
, Krishnakumar, A.%
, K{\"o}rfer, M.%
, Hoo, S.B.%
\BDBL {}Hutter, F.%
\end{APACrefauthors}%
\unskip\
\newblock
\APACrefYearMonthDay{2025}{{\APACmonth{01}}}{}.
\newblock
{\BBOQ}\APACrefatitle {Accurate Predictions on Small Data with a Tabular Foundation Model} {Accurate predictions on small data with a tabular foundation model}.{\BBCQ}
\newblock
\APACjournalVolNumPages{Nature}{637}{8045}{319--326,}
\newblock
\begin{APACrefDOI} \doi{10.1038/s41586-024-08328-6} \end{APACrefDOI}
\newblock

\newblock

\PrintBackRefs{\CurrentBib}

\bibitem [\protect \citeauthoryear {%
Hunter%
}{%
Hunter%
}{%
{\protect \APACyear {2007}}%
}]{%
hunterMatplotlib2DGraphics2007}
\APACinsertmetastar {%
hunterMatplotlib2DGraphics2007}%
\begin{APACrefauthors}%
Hunter, J.D.%
\end{APACrefauthors}%
\unskip\
\newblock
\APACrefYearMonthDay{2007}{{\APACmonth{05}}}{}.
\newblock
{\BBOQ}\APACrefatitle {Matplotlib: {{A 2D Graphics Environment}}} {Matplotlib: {{A 2D Graphics Environment}}}.{\BBCQ}
\newblock
\APACjournalVolNumPages{Computing in Science Engineering}{9}{3}{90--95,}
\newblock
\begin{APACrefDOI} \doi{10.1109/MCSE.2007.55} \end{APACrefDOI}
\newblock

\newblock

\PrintBackRefs{\CurrentBib}

\bibitem [\protect \citeauthoryear {%
Ke%
\ \protect \BOthers {.}}{%
Ke%
\ \protect \BOthers {.}}{%
{\protect \APACyear {2017}}%
}]{%
keLightGBMHighlyEfficient2017}
\APACinsertmetastar {%
keLightGBMHighlyEfficient2017}%
\begin{APACrefauthors}%
Ke, G.%
, Meng, Q.%
, Finley, T.%
, Wang, T.%
, Chen, W.%
, Ma, W.%
\BDBL {}Liu, T\BHBI Y.%
\end{APACrefauthors}%
\unskip\
\newblock
\APACrefYearMonthDay{2017}{}{}.
\newblock
{\BBOQ}\APACrefatitle {{{LightGBM}}: A Highly Efficient Gradient Boosting Decision Tree} {{{LightGBM}}: A highly efficient gradient boosting decision tree}.{\BBCQ}
\newblock
 \APACrefbtitle {Advances in Neural Information Processing Systems} {Advances in neural information processing systems}\ (\BVOL~30, \BPGS\ 3146--3154).
\PrintBackRefs{\CurrentBib}

\bibitem [\protect \citeauthoryear {%
Kottapalli%
\ \protect \BOthers {.}}{%
Kottapalli%
\ \protect \BOthers {.}}{%
{\protect \APACyear {2025}}%
}]{%
kottapalliFoundationModelsTime2025}
\APACinsertmetastar {%
kottapalliFoundationModelsTime2025}%
\begin{APACrefauthors}%
Kottapalli, S.R.K.%
, Hubli, K.%
, Chandrashekhara, S.%
, Jain, G.%
, Hubli, S.%
, Botla, G.%
\BCBL {} Doddaiah, R.%
\end{APACrefauthors}%
\unskip\
\newblock
\APACrefYearMonthDay{2025}{{\APACmonth{04}}}{}.
\newblock
\APACrefbtitle {Foundation Models for Time Series: A Survey.} {Foundation models for time series: A survey.}
\newblock
\APACrefnote{arXiv preprint}
\PrintBackRefs{\CurrentBib}

\bibitem [\protect \citeauthoryear {%
K{\"u}ken%
, Hoo%
, Mr{\'a}z%
, Hutter%
\BCBL {}\ \BBA {} Purucker%
}{%
K{\"u}ken%
\ \protect \BOthers {.}}{%
{\protect \APACyear {2026}}%
}]{%
kukenTimEEEndtoendTime2026}
\APACinsertmetastar {%
kukenTimEEEndtoendTime2026}%
\begin{APACrefauthors}%
K{\"u}ken, J.%
, Hoo, S.B.%
, Mr{\'a}z, M.%
, Hutter, F.%
\BCBL {} Purucker, L.%
\end{APACrefauthors}%
\unskip\
\newblock
\APACrefYearMonthDay{2026}{{\APACmonth{07}}}{}.
\newblock
\APACrefbtitle {{{TimEE}}: End-to-End Time Series Classification via In-Context Learning.} {{{TimEE}}: End-to-end time series classification via in-context learning.}
\newblock
\APACrefnote{arXiv preprint}
\PrintBackRefs{\CurrentBib}

\bibitem [\protect \citeauthoryear {%
Lam%
, Pitrou%
\BCBL {}\ \BBA {} Seibert%
}{%
Lam%
\ \protect \BOthers {.}}{%
{\protect \APACyear {2015}}%
}]{%
lamNumbaLLVMbasedPython2015}
\APACinsertmetastar {%
lamNumbaLLVMbasedPython2015}%
\begin{APACrefauthors}%
Lam, S.K.%
, Pitrou, A.%
\BCBL {} Seibert, S.%
\end{APACrefauthors}%
\unskip\
\newblock
\APACrefYearMonthDay{2015}{}{}.
\newblock
{\BBOQ}\APACrefatitle {Numba: {{A LLVM-based Python JIT compiler}}} {Numba: {{A LLVM-based Python JIT compiler}}}.{\BBCQ}
\newblock
 \APACrefbtitle {Proceedings of the {{Second Workshop}} on the {{LLVM Compiler Infrastructure}} in {{HPC}}} {Proceedings of the {{Second Workshop}} on the {{LLVM Compiler Infrastructure}} in {{HPC}}}\ (\BPGS\ 1--6).
\newblock
\APACaddressPublisher{New York, NY, USA}{Association for Computing Machinery}.
\PrintBackRefs{\CurrentBib}

\bibitem [\protect \citeauthoryear {%
Lin%
, Keogh%
, Wei%
\BCBL {}\ \BBA {} Lonardi%
}{%
Lin%
\ \protect \BOthers {.}}{%
{\protect \APACyear {2007}}%
}]{%
linExperiencingSAXNovel2007a}
\APACinsertmetastar {%
linExperiencingSAXNovel2007a}%
\begin{APACrefauthors}%
Lin, J.%
, Keogh, E.%
, Wei, L.%
\BCBL {} Lonardi, S.%
\end{APACrefauthors}%
\unskip\
\newblock
\APACrefYearMonthDay{2007}{{\APACmonth{10}}}{}.
\newblock
{\BBOQ}\APACrefatitle {Experiencing {{SAX}}: A Novel Symbolic Representation of Time Series} {Experiencing {{SAX}}: A novel symbolic representation of time series}.{\BBCQ}
\newblock
\APACjournalVolNumPages{Data Mining and Knowledge Discovery}{15}{2}{107--144,}
\newblock
\begin{APACrefDOI} \doi{10.1007/s10618-007-0064-z} \end{APACrefDOI}
\newblock

\newblock

\PrintBackRefs{\CurrentBib}

\bibitem [\protect \citeauthoryear {%
Lines%
, Taylor%
\BCBL {}\ \BBA {} Bagnall%
}{%
Lines%
\ \protect \BOthers {.}}{%
{\protect \APACyear {2018}}%
}]{%
linesTimeSeriesClassification2018a}
\APACinsertmetastar {%
linesTimeSeriesClassification2018a}%
\begin{APACrefauthors}%
Lines, J.%
, Taylor, S.%
\BCBL {} Bagnall, A.%
\end{APACrefauthors}%
\unskip\
\newblock
\APACrefYearMonthDay{2018}{{\APACmonth{07}}}{}.
\newblock
{\BBOQ}\APACrefatitle {Time {{Series Classification}} with {{HIVE-COTE}}: {{The Hierarchical Vote Collective}} of {{Transformation-Based Ensembles}}} {Time {{Series Classification}} with {{HIVE-COTE}}: {{The Hierarchical Vote Collective}} of {{Transformation-Based Ensembles}}}.{\BBCQ}
\newblock
\APACjournalVolNumPages{ACM Trans. Knowl. Discov. Data}{12}{5}{52:1--52:35,}
\newblock
\begin{APACrefDOI} \doi{10.1145/3182382} \end{APACrefDOI}
\newblock

\newblock

\PrintBackRefs{\CurrentBib}

\bibitem [\protect \citeauthoryear {%
Liu%
\ \protect \BOthers {.}}{%
Liu%
\ \protect \BOthers {.}}{%
{\protect \APACyear {2026}}%
}]{%
liuUniShapeUnifiedShapeAware2026}
\APACinsertmetastar {%
liuUniShapeUnifiedShapeAware2026}%
\begin{APACrefauthors}%
Liu, Z.%
, Wang, Y.%
, Li, B.%
, Zheng, J.%
, Eldele, E.%
, Wu, M.%
\BCBL {} Ma, Q.%
\end{APACrefauthors}%
\unskip\
\newblock
\APACrefYearMonthDay{2026}{{\APACmonth{01}}}{}.
\newblock
\APACrefbtitle {A Unified Shape-Aware Foundation Model for Time Series Classification.} {A unified shape-aware foundation model for time series classification.}
\newblock
\APACrefnote{arXiv preprint}
\PrintBackRefs{\CurrentBib}

\bibitem [\protect \citeauthoryear {%
Lucas%
\ \protect \BOthers {.}}{%
Lucas%
\ \protect \BOthers {.}}{%
{\protect \APACyear {2019}}%
}]{%
lucasProximityForest2019}
\APACinsertmetastar {%
lucasProximityForest2019}%
\begin{APACrefauthors}%
Lucas, B.%
, Shifaz, A.%
, Pelletier, C.%
, O'Neill, L.%
, Zaidi, N.%
, Goethals, B.%
\BDBL {}Webb, G.I.%
\end{APACrefauthors}%
\unskip\
\newblock
\APACrefYearMonthDay{2019}{}{}.
\newblock
{\BBOQ}\APACrefatitle {Proximity {{Forest}}: An Effective and Scalable Distance-Based Classifier for Time Series} {Proximity {{Forest}}: An effective and scalable distance-based classifier for time series}.{\BBCQ}
\newblock
\APACjournalVolNumPages{Data Mining and Knowledge Discovery}{33}{3}{607--635,}
\newblock
\begin{APACrefDOI} \doi{10.1007/s10618-019-00617-3} \end{APACrefDOI}
\newblock

\newblock

\PrintBackRefs{\CurrentBib}

\bibitem [\protect \citeauthoryear {%
Middlehurst%
\ \BBA {} Bagnall%
}{%
Middlehurst%
\ \BBA {} Bagnall%
}{%
{\protect \APACyear {2023}}%
}]{%
middlehurstExtractingFeatures2023}
\APACinsertmetastar {%
middlehurstExtractingFeatures2023}%
\begin{APACrefauthors}%
Middlehurst, M.%
\BCBT {}\ \BBA {} Bagnall, A.%
\end{APACrefauthors}%
\unskip\
\newblock
\APACrefYearMonthDay{2023}{}{}.
\newblock
{\BBOQ}\APACrefatitle {Extracting {{Features}} from {{Random Subseries}}: {{A Hybrid Pipeline}} for {{Time Series Classification}} and {{Extrinsic Regression}}} {Extracting {{Features}} from {{Random Subseries}}: {{A Hybrid Pipeline}} for {{Time Series Classification}} and {{Extrinsic Regression}}}.{\BBCQ}
\newblock
 \APACrefbtitle {Advanced {{Analytics}} and {{Learning}} on {{Temporal Data}}: {{8th ECML PKDD Workshop}}, {{AALTD}} 2023.} {Advanced {{Analytics}} and {{Learning}} on {{Temporal Data}}: {{8th ECML PKDD Workshop}}, {{AALTD}} 2023.}
\newblock
\APACaddressPublisher{Cham}{Springer}.
\PrintBackRefs{\CurrentBib}

\bibitem [\protect \citeauthoryear {%
Middlehurst%
, {Ismail-Fawaz}%
\BCBL {}\ \protect \BOthers {.}}{%
Middlehurst%
, {Ismail-Fawaz}%
\BCBL {}\ \protect \BOthers {.}}{%
{\protect \APACyear {2024}}%
}]{%
middlehurstAeonPythonToolkit2024}
\APACinsertmetastar {%
middlehurstAeonPythonToolkit2024}%
\begin{APACrefauthors}%
Middlehurst, M.%
, {Ismail-Fawaz}, A.%
, Guillaume, A.%
, Holder, C.%
, {Guijo-Rubio}, D.%
, Bulatova, G.%
\BDBL {}Bagnall, A.%
\end{APACrefauthors}%
\unskip\
\newblock
\APACrefYearMonthDay{2024}{}{}.
\newblock
{\BBOQ}\APACrefatitle {Aeon: A {{Python Toolkit}} for {{Learning}} from {{Time Series}}} {Aeon: A {{Python Toolkit}} for {{Learning}} from {{Time Series}}}.{\BBCQ}
\newblock
\APACjournalVolNumPages{Journal of Machine Learning Research}{25}{289}{1--10,}
\newblock

\newblock

\PrintBackRefs{\CurrentBib}

\bibitem [\protect \citeauthoryear {%
Middlehurst%
, Large%
\BCBL {}\ \BBA {} Bagnall%
}{%
Middlehurst%
\ \protect \BOthers {.}}{%
{\protect \APACyear {2020}}%
}]{%
middlehurstCanonicalIntervalForest2020}
\APACinsertmetastar {%
middlehurstCanonicalIntervalForest2020}%
\begin{APACrefauthors}%
Middlehurst, M.%
, Large, J.%
\BCBL {} Bagnall, A.%
\end{APACrefauthors}%
\unskip\
\newblock
\APACrefYearMonthDay{2020}{{\APACmonth{12}}}{}.
\newblock
{\BBOQ}\APACrefatitle {The {{Canonical Interval Forest}} ({{CIF}}) {{Classifier}} for {{Time Series Classification}}} {The {{Canonical Interval Forest}} ({{CIF}}) {{Classifier}} for {{Time Series Classification}}}.{\BBCQ}
\newblock
 \APACrefbtitle {2020 {{IEEE International Conference}} on {{Big Data}} ({{Big Data}})} {2020 {{IEEE International Conference}} on {{Big Data}} ({{Big Data}})}\ (\BPGS\ 188--195).
\PrintBackRefs{\CurrentBib}

\bibitem [\protect \citeauthoryear {%
Middlehurst%
\ \protect \BOthers {.}}{%
Middlehurst%
\ \protect \BOthers {.}}{%
{\protect \APACyear {2021}}%
}]{%
middlehurstHIVECOTE20New2021a}
\APACinsertmetastar {%
middlehurstHIVECOTE20New2021a}%
\begin{APACrefauthors}%
Middlehurst, M.%
, Large, J.%
, Flynn, M.%
, Lines, J.%
, Bostrom, A.%
\BCBL {} Bagnall, A.%
\end{APACrefauthors}%
\unskip\
\newblock
\APACrefYearMonthDay{2021}{{\APACmonth{12}}}{}.
\newblock
{\BBOQ}\APACrefatitle {{{HIVE-COTE}} 2.0: A New Meta Ensemble for Time Series Classification} {{{HIVE-COTE}} 2.0: A new meta ensemble for time series classification}.{\BBCQ}
\newblock
\APACjournalVolNumPages{Machine Learning}{110}{11}{3211--3243,}
\newblock
\begin{APACrefDOI} \doi{10.1007/s10994-021-06057-9} \end{APACrefDOI}
\newblock

\newblock

\PrintBackRefs{\CurrentBib}

\bibitem [\protect \citeauthoryear {%
Middlehurst%
, Sch{\"a}fer%
\BCBL {}\ \BBA {} Bagnall%
}{%
Middlehurst%
, Sch{\"a}fer%
\BCBL {}\ \BBA {} Bagnall%
}{%
{\protect \APACyear {2024}}%
}]{%
middlehurstBakeReduxReview2024}
\APACinsertmetastar {%
middlehurstBakeReduxReview2024}%
\begin{APACrefauthors}%
Middlehurst, M.%
, Sch{\"a}fer, P.%
\BCBL {} Bagnall, A.%
\end{APACrefauthors}%
\unskip\
\newblock
\APACrefYearMonthDay{2024}{{\APACmonth{07}}}{}.
\newblock
{\BBOQ}\APACrefatitle {Bake off Redux: A Review and Experimental Evaluation of Recent Time Series Classification Algorithms} {Bake off redux: A review and experimental evaluation of recent time series classification algorithms}.{\BBCQ}
\newblock
\APACjournalVolNumPages{Data Mining and Knowledge Discovery}{38}{4}{1958--2031,}
\newblock
\begin{APACrefDOI} \doi{10.1007/s10618-024-01022-1} \end{APACrefDOI}
\newblock

\newblock

\PrintBackRefs{\CurrentBib}

\bibitem [\protect \citeauthoryear {%
O'Rourke%
, Trisovic%
\BCBL {}\ \BBA {} Bertsimas%
}{%
O'Rourke%
\ \protect \BOthers {.}}{%
{\protect \APACyear {2026}}%
}]{%
orourkeRocketPFNAccurateTime2026}
\APACinsertmetastar {%
orourkeRocketPFNAccurateTime2026}%
\begin{APACrefauthors}%
O'Rourke, F.M.%
, Trisovic, A.%
\BCBL {} Bertsimas, D.%
\end{APACrefauthors}%
\unskip\
\newblock
\APACrefYearMonthDay{2026}{{\APACmonth{06}}}{}.
\newblock
\APACrefbtitle {{{RocketPFN}}: Accurate Time Series Classification via In-Context Learning.} {{{RocketPFN}}: Accurate time series classification via in-context learning.}
\newblock
\APACrefnote{arXiv preprint}
\PrintBackRefs{\CurrentBib}

\bibitem [\protect \citeauthoryear {%
Pedregosa%
\ \protect \BOthers {.}}{%
Pedregosa%
\ \protect \BOthers {.}}{%
{\protect \APACyear {2011}}%
}]{%
pedregosaScikitlearnMachineLearning2011}
\APACinsertmetastar {%
pedregosaScikitlearnMachineLearning2011}%
\begin{APACrefauthors}%
Pedregosa, F.%
, Varoquaux, G.%
, Gramfort, A.%
, Michel, V.%
, Thirion, B.%
, Grisel, O.%
\BDBL {}Duchesnay, {\'E}.%
\end{APACrefauthors}%
\unskip\
\newblock
\APACrefYearMonthDay{2011}{{\APACmonth{10}}}{}.
\newblock
{\BBOQ}\APACrefatitle {Scikit-Learn: {{Machine Learning}} in {{Python}}} {Scikit-learn: {{Machine Learning}} in {{Python}}}.{\BBCQ}
\newblock
\APACjournalVolNumPages{Journal of Machine Learning Research}{12}{}{2825-2830,}
\newblock

\newblock

\PrintBackRefs{\CurrentBib}

\bibitem [\protect \citeauthoryear {%
Perron%
}{%
Perron%
}{%
{\protect \APACyear {1989}}%
}]{%
perron1989great}
\APACinsertmetastar {%
perron1989great}%
\begin{APACrefauthors}%
Perron, P.%
\end{APACrefauthors}%
\unskip\
\newblock
\APACrefYearMonthDay{1989}{}{}.
\newblock
{\BBOQ}\APACrefatitle {The great crash, the oil price shock, and the unit root hypothesis} {The great crash, the oil price shock, and the unit root hypothesis}.{\BBCQ}
\newblock
\APACjournalVolNumPages{Econometrica: journal of the Econometric Society}{}{}{1361--1401,}
\newblock

\newblock

\PrintBackRefs{\CurrentBib}

\bibitem [\protect \citeauthoryear {%
Platt%
}{%
Platt%
}{%
{\protect \APACyear {1999}}%
}]{%
platt1999probabilistic}
\APACinsertmetastar {%
platt1999probabilistic}%
\begin{APACrefauthors}%
Platt, J.%
\end{APACrefauthors}%
\unskip\
\newblock
\APACrefYearMonthDay{1999}{}{}.
\newblock
{\BBOQ}\APACrefatitle {Probabilistic Outputs for Support Vector Machines and Comparisons to Regularized Likelihood Methods} {Probabilistic outputs for support vector machines and comparisons to regularized likelihood methods}.{\BBCQ}
\newblock
 A.J.~Smola, P.~Bartlett, B.~Sch{\"o}lkopf\BCBL {}\ \BBA {} D.~Schuurmans\ (\BEDS), \APACrefbtitle {Advances in Large Margin Classifiers} {Advances in large margin classifiers}\ (\BPGS\ 61--74).
\newblock
\APACaddressPublisher{}{MIT Press}.
\PrintBackRefs{\CurrentBib}

\bibitem [\protect \citeauthoryear {%
Qu%
, Holzm{\"u}ller%
, Varoquaux%
\BCBL {}\ \BBA {} {Le Morvan}%
}{%
Qu%
\ \protect \BOthers {.}}{%
{\protect \APACyear {2025}}%
}]{%
quTabICLTabularFoundation2025}
\APACinsertmetastar {%
quTabICLTabularFoundation2025}%
\begin{APACrefauthors}%
Qu, J.%
, Holzm{\"u}ller, D.%
, Varoquaux, G.%
\BCBL {} {Le Morvan}, M.%
\end{APACrefauthors}%
\unskip\
\newblock
\APACrefYearMonthDay{2025}{}{}.
\newblock
{\BBOQ}\APACrefatitle {{{TabICL}}: A Tabular Foundation Model for In-Context Learning on Large Data} {{{TabICL}}: A tabular foundation model for in-context learning on large data}.{\BBCQ}
\newblock
 \APACrefbtitle {Proceedings of the 42nd International Conference on Machine Learning} {Proceedings of the 42nd international conference on machine learning}\ (\BVOL~267).
\PrintBackRefs{\CurrentBib}

\bibitem [\protect \citeauthoryear {%
Qu%
, Holzm{\"u}ller%
, Varoquaux%
\BCBL {}\ \BBA {} {Le Morvan}%
}{%
Qu%
\ \protect \BOthers {.}}{%
{\protect \APACyear {2026}}%
}]{%
quTabICLv2BetterFaster2026}
\APACinsertmetastar {%
quTabICLv2BetterFaster2026}%
\begin{APACrefauthors}%
Qu, J.%
, Holzm{\"u}ller, D.%
, Varoquaux, G.%
\BCBL {} {Le Morvan}, M.%
\end{APACrefauthors}%
\unskip\
\newblock
\APACrefYearMonthDay{2026}{{\APACmonth{02}}}{}.
\newblock
\APACrefbtitle {{{TabICLv2}}: A Better, Faster, Scalable, and Open Tabular Foundation Model.} {{{TabICLv2}}: A better, faster, scalable, and open tabular foundation model.}
\newblock
\APACrefnote{arXiv preprint}
\PrintBackRefs{\CurrentBib}

\bibitem [\protect \citeauthoryear {%
Sch{\"a}fer%
}{%
Sch{\"a}fer%
}{%
{\protect \APACyear {2015}}%
}]{%
schaferBOSSConcernedTime2015}
\APACinsertmetastar {%
schaferBOSSConcernedTime2015}%
\begin{APACrefauthors}%
Sch{\"a}fer, P.%
\end{APACrefauthors}%
\unskip\
\newblock
\APACrefYearMonthDay{2015}{{\APACmonth{11}}}{}.
\newblock
{\BBOQ}\APACrefatitle {The {{BOSS}} Is Concerned with Time Series Classification in the Presence of Noise} {The {{BOSS}} is concerned with time series classification in the presence of noise}.{\BBCQ}
\newblock
\APACjournalVolNumPages{Data Mining and Knowledge Discovery}{29}{6}{1505--1530,}
\newblock
\begin{APACrefDOI} \doi{10.1007/s10618-014-0377-7} \end{APACrefDOI}
\newblock

\newblock

\PrintBackRefs{\CurrentBib}

\bibitem [\protect \citeauthoryear {%
Sch{\"a}fer%
\ \BBA {} H{\"o}gqvist%
}{%
Sch{\"a}fer%
\ \BBA {} H{\"o}gqvist%
}{%
{\protect \APACyear {2012}}%
}]{%
schaferSFASymbolicFourier2012}
\APACinsertmetastar {%
schaferSFASymbolicFourier2012}%
\begin{APACrefauthors}%
Sch{\"a}fer, P.%
\BCBT {}\ \BBA {} H{\"o}gqvist, M.%
\end{APACrefauthors}%
\unskip\
\newblock
\APACrefYearMonthDay{2012}{}{}.
\newblock
{\BBOQ}\APACrefatitle {{{SFA}}: A Symbolic Fourier Approximation and Index for Similarity Search in High Dimensional Datasets} {{{SFA}}: A symbolic fourier approximation and index for similarity search in high dimensional datasets}.{\BBCQ}
\newblock
 \APACrefbtitle {Proceedings of the 15th {{International Conference}} on {{Extending Database Technology}} - {{EDBT}} '12} {Proceedings of the 15th {{International Conference}} on {{Extending Database Technology}} - {{EDBT}} '12}\ (\BPG~516).
\newblock
\APACaddressPublisher{Berlin, Germany}{ACM Press}.
\PrintBackRefs{\CurrentBib}

\bibitem [\protect \citeauthoryear {%
Sch{\"a}fer%
\ \BBA {} Leser%
}{%
Sch{\"a}fer%
\ \BBA {} Leser%
}{%
{\protect \APACyear {2017}}%
}]{%
schaferFastAccurateTime2017a}
\APACinsertmetastar {%
schaferFastAccurateTime2017a}%
\begin{APACrefauthors}%
Sch{\"a}fer, P.%
\BCBT {}\ \BBA {} Leser, U.%
\end{APACrefauthors}%
\unskip\
\newblock
\APACrefYearMonthDay{2017}{{\APACmonth{11}}}{}.
\newblock
{\BBOQ}\APACrefatitle {Fast and {{Accurate Time Series Classification}} with {{WEASEL}}} {Fast and {{Accurate Time Series Classification}} with {{WEASEL}}}.{\BBCQ}
\newblock
 \APACrefbtitle {Proceedings of the 2017 {{ACM}} on {{Conference}} on {{Information}} and {{Knowledge Management}}} {Proceedings of the 2017 {{ACM}} on {{Conference}} on {{Information}} and {{Knowledge Management}}}\ (\BPGS\ 637--646).
\newblock
\APACaddressPublisher{New York, NY, USA}{Association for Computing Machinery}.
\PrintBackRefs{\CurrentBib}

\bibitem [\protect \citeauthoryear {%
Sch{\"a}fer%
\ \BBA {} Leser%
}{%
Sch{\"a}fer%
\ \BBA {} Leser%
}{%
{\protect \APACyear {2023}}%
}]{%
schaferWEASEL20Random2023}
\APACinsertmetastar {%
schaferWEASEL20Random2023}%
\begin{APACrefauthors}%
Sch{\"a}fer, P.%
\BCBT {}\ \BBA {} Leser, U.%
\end{APACrefauthors}%
\unskip\
\newblock
\APACrefYearMonthDay{2023}{{\APACmonth{12}}}{}.
\newblock
{\BBOQ}\APACrefatitle {{{WEASEL}} 2.0: A Random Dilated Dictionary Transform for Fast, Accurate and Memory Constrained Time Series Classification} {{{WEASEL}} 2.0: A random dilated dictionary transform for fast, accurate and memory constrained time series classification}.{\BBCQ}
\newblock
\APACjournalVolNumPages{Machine Learning}{112}{12}{4763--4788,}
\newblock
\begin{APACrefDOI} \doi{10.1007/s10994-023-06395-w} \end{APACrefDOI}
\newblock

\newblock

\PrintBackRefs{\CurrentBib}

\bibitem [\protect \citeauthoryear {%
Shifaz%
, Pelletier%
, Petitjean%
\BCBL {}\ \BBA {} Webb%
}{%
Shifaz%
\ \protect \BOthers {.}}{%
{\protect \APACyear {2020}}%
}]{%
shifazTSCHIEF2020}
\APACinsertmetastar {%
shifazTSCHIEF2020}%
\begin{APACrefauthors}%
Shifaz, A.%
, Pelletier, C.%
, Petitjean, F.%
\BCBL {} Webb, G.I.%
\end{APACrefauthors}%
\unskip\
\newblock
\APACrefYearMonthDay{2020}{}{}.
\newblock
{\BBOQ}\APACrefatitle {{{TS-CHIEF}}: A Scalable and Accurate Forest Algorithm for Time Series Classification} {{{TS-CHIEF}}: A scalable and accurate forest algorithm for time series classification}.{\BBCQ}
\newblock
\APACjournalVolNumPages{Data Mining and Knowledge Discovery}{34}{3}{742--775,}
\newblock
\begin{APACrefDOI} \doi{10.1007/s10618-020-00679-8} \end{APACrefDOI}
\newblock

\newblock

\PrintBackRefs{\CurrentBib}

\bibitem [\protect \citeauthoryear {%
Tan%
, Dempster%
, Bergmeir%
\BCBL {}\ \BBA {} Webb%
}{%
Tan%
\ \protect \BOthers {.}}{%
{\protect \APACyear {2022}}%
}]{%
tanMultiRocketMultiplePooling2022}
\APACinsertmetastar {%
tanMultiRocketMultiplePooling2022}%
\begin{APACrefauthors}%
Tan, C.W.%
, Dempster, A.%
, Bergmeir, C.%
\BCBL {} Webb, G.I.%
\end{APACrefauthors}%
\unskip\
\newblock
\APACrefYearMonthDay{2022}{{\APACmonth{09}}}{}.
\newblock
{\BBOQ}\APACrefatitle {{{MultiRocket}}: Multiple Pooling Operators and Transformations for Fast and Effective Time Series Classification} {{{MultiRocket}}: Multiple pooling operators and transformations for fast and effective time series classification}.{\BBCQ}
\newblock
\APACjournalVolNumPages{Data Mining and Knowledge Discovery}{36}{5}{1623--1646,}
\newblock
\begin{APACrefDOI} \doi{10.1007/s10618-022-00844-1} \end{APACrefDOI}
\newblock

\newblock

\PrintBackRefs{\CurrentBib}

\bibitem [\protect \citeauthoryear {%
Tavenard%
\ \protect \BOthers {.}}{%
Tavenard%
\ \protect \BOthers {.}}{%
{\protect \APACyear {2020}}%
}]{%
tavenardTslearnMachineLearning2020}
\APACinsertmetastar {%
tavenardTslearnMachineLearning2020}%
\begin{APACrefauthors}%
Tavenard, R.%
, Faouzi, J.%
, Vandewiele, G.%
, Divo, F.%
, Androz, G.%
, Holtz, C.%
\BDBL {}Woods, E.%
\end{APACrefauthors}%
\unskip\
\newblock
\APACrefYearMonthDay{2020}{}{}.
\newblock
{\BBOQ}\APACrefatitle {Tslearn, {{A Machine Learning Toolkit}} for {{Time Series Data}}} {Tslearn, {{A Machine Learning Toolkit}} for {{Time Series Data}}}.{\BBCQ}
\newblock
\APACjournalVolNumPages{Journal of Machine Learning Research}{21}{118}{1--6,}
\newblock

\newblock

\PrintBackRefs{\CurrentBib}

\bibitem [\protect \citeauthoryear {%
Uribarri%
, Barone%
, Ansuini%
\BCBL {}\ \BBA {} Frans{\'e}n%
}{%
Uribarri%
\ \protect \BOthers {.}}{%
{\protect \APACyear {2024}}%
}]{%
uribarriDetachROCKETSequential2024}
\APACinsertmetastar {%
uribarriDetachROCKETSequential2024}%
\begin{APACrefauthors}%
Uribarri, G.%
, Barone, F.%
, Ansuini, A.%
\BCBL {} Frans{\'e}n, E.%
\end{APACrefauthors}%
\unskip\
\newblock
\APACrefYearMonthDay{2024}{}{}.
\newblock
{\BBOQ}\APACrefatitle {{{Detach-ROCKET}}: Sequential Feature Selection for Time Series Classification with Random Convolutional Kernels} {{{Detach-ROCKET}}: Sequential feature selection for time series classification with random convolutional kernels}.{\BBCQ}
\newblock
\APACjournalVolNumPages{Data Mining and Knowledge Discovery}{38}{6}{3922--3947,}
\newblock
\begin{APACrefDOI} \doi{10.1007/s10618-024-01062-7} \end{APACrefDOI}
\newblock

\newblock

\PrintBackRefs{\CurrentBib}

\bibitem [\protect \citeauthoryear {%
Virtanen%
\ \protect \BOthers {.}}{%
Virtanen%
\ \protect \BOthers {.}}{%
{\protect \APACyear {2020}}%
}]{%
virtanenSciPy10Fundamental2020}
\APACinsertmetastar {%
virtanenSciPy10Fundamental2020}%
\begin{APACrefauthors}%
Virtanen, P.%
, Gommers, R.%
, Oliphant, T.E.%
, Haberland, M.%
, Reddy, T.%
, Cournapeau, D.%
\BDBL {}{SciPy 1.0 Contributors}%
\end{APACrefauthors}%
\unskip\
\newblock
\APACrefYearMonthDay{2020}{{\APACmonth{02}}}{}.
\newblock
{\BBOQ}\APACrefatitle {{{SciPy}} 1.0: Fundamental Algorithms for Scientific Computing in {{Python}}} {{{SciPy}} 1.0: Fundamental algorithms for scientific computing in {{Python}}}.{\BBCQ}
\newblock
\APACjournalVolNumPages{Nature Methods}{17}{3}{261--272,}
\newblock
\begin{APACrefDOI} \doi{10.1038/s41592-019-0686-2} \end{APACrefDOI}
\newblock

\newblock

\PrintBackRefs{\CurrentBib}

\bibitem [\protect \citeauthoryear {%
Wang%
, Wu%
, Wei%
\BCBL {}\ \BBA {} Li%
}{%
Wang%
\ \protect \BOthers {.}}{%
{\protect \APACyear {2025}}%
}]{%
wangKernelGroupingTime2025}
\APACinsertmetastar {%
wangKernelGroupingTime2025}%
\begin{APACrefauthors}%
Wang, P.%
, Wu, J.%
, Wei, Y.%
\BCBL {} Li, T.%
\end{APACrefauthors}%
\unskip\
\newblock
\APACrefYearMonthDay{2025}{{\APACmonth{12}}}{}.
\newblock
{\BBOQ}\APACrefatitle {Kernel Grouping for Time Series Classification with Multiple Transformations and Pooling Operators} {Kernel grouping for time series classification with multiple transformations and pooling operators}.{\BBCQ}
\newblock
\APACjournalVolNumPages{Expert Systems with Applications}{293}{}{128693,}
\newblock
\begin{APACrefDOI} \doi{10.1016/j.eswa.2025.128693} \end{APACrefDOI}
\newblock

\newblock

\PrintBackRefs{\CurrentBib}

\bibitem [\protect \citeauthoryear {%
{W}es {M}c{K}inney%
}{%
{W}es {M}c{K}inney%
}{%
{\protect \APACyear {2010}}%
}]{%
mckinney-proc-scipy-2010}
\APACinsertmetastar {%
mckinney-proc-scipy-2010}%
\begin{APACrefauthors}%
{W}es {M}c{K}inney%
\end{APACrefauthors}%
\unskip\
\newblock
\APACrefYearMonthDay{2010}{}{}.
\newblock
{\BBOQ}\APACrefatitle {{D}ata {S}tructures for {S}tatistical {C}omputing in {P}ython} {{D}ata {S}tructures for {S}tatistical {C}omputing in {P}ython}.{\BBCQ}
\newblock
 {S}t\'efan van~der {W}alt\ \BBA {} {J}arrod {M}illman\ (\BEDS), \APACrefbtitle {{P}roceedings of the 9th {P}ython in {S}cience {C}onference} {{P}roceedings of the 9th {P}ython in {S}cience {C}onference}\ (\BPG~56 - 61).
\PrintBackRefs{\CurrentBib}

\bibitem [\protect \citeauthoryear {%
Xie%
\ \protect \BOthers {.}}{%
Xie%
\ \protect \BOthers {.}}{%
{\protect \APACyear {2025}}%
}]{%
xieCauKerClassificationTime2025}
\APACinsertmetastar {%
xieCauKerClassificationTime2025}%
\begin{APACrefauthors}%
Xie, S.%
, Feofanov, V.%
, Odonnat, A.%
, Zan, L.%
, Alonso, M.%
, Zhang, J.%
\BDBL {}Redko, I.%
\end{APACrefauthors}%
\unskip\
\newblock
\APACrefYearMonthDay{2025}{{\APACmonth{08}}}{}.
\newblock
\APACrefbtitle {{{CauKer}}: Classification Time Series Foundation Models Can Be Pretrained on Synthetic Data.} {{{CauKer}}: Classification time series foundation models can be pretrained on synthetic data.}
\newblock
\APACrefnote{arXiv preprint}
\PrintBackRefs{\CurrentBib}

\bibitem [\protect \citeauthoryear {%
Ye%
\ \BBA {} Keogh%
}{%
Ye%
\ \BBA {} Keogh%
}{%
{\protect \APACyear {2011}}%
}]{%
yeTimeSeriesShapelets2011}
\APACinsertmetastar {%
yeTimeSeriesShapelets2011}%
\begin{APACrefauthors}%
Ye, L.%
\BCBT {}\ \BBA {} Keogh, E.%
\end{APACrefauthors}%
\unskip\
\newblock
\APACrefYearMonthDay{2011}{{\APACmonth{01}}}{}.
\newblock
{\BBOQ}\APACrefatitle {Time Series Shapelets: A Novel Technique That Allows Accurate, Interpretable and Fast Classification} {Time series shapelets: A novel technique that allows accurate, interpretable and fast classification}.{\BBCQ}
\newblock
\APACjournalVolNumPages{Data Mining and Knowledge Discovery}{22}{1}{149--182,}
\newblock
\begin{APACrefDOI} \doi{10.1007/s10618-010-0179-5} \end{APACrefDOI}
\newblock

\newblock

\PrintBackRefs{\CurrentBib}

\end{thebibliography}

\newpage

\begin{appendices}

\section{The 142 data sets of the full benchmark}\label{secA}

Table~\ref{tab:full-benchmark-datasets} lists, for every one of the 142 data sets of the full evaluation benchmark, the same information as Table~\ref{tab:devsubset} reports for the twenty-data-set development subset: the number of training and test instances, the number of classes, the series length, and the mean number of training samples per class.
The twenty development data sets of Table~\ref{tab:devsubset} are included here as well, so this table alone covers the full benchmark.
Table~\ref{tab:devsubset}'s training-coverage column is specific to the development subset's own hyperparameter-optimization procedure and is not reported here, since no such procedure was run on the 122 data sets outside that subset.

\begingroup\footnotesize
\begin{longtable}{lrrrrr}
\caption{The 142 data sets of the full evaluation benchmark (mean samples/class computed on the training set).}
\label{tab:full-benchmark-datasets} \\
\toprule
Data set & Training & Test & Classes & Length & Mean samples/class \\
\midrule
\endfirsthead
\multicolumn{6}{c}{{\tablename\ \thetable{} -- continued from previous page}} \\
\toprule
Data set & Training & Test & Classes & Length & Mean samples/class \\
\midrule
\endhead
\midrule \multicolumn{6}{r}{{Continued on next page}} \\
\endfoot
\bottomrule
\endlastfoot
ACSF1 & 100 & 100 & 10 & 1460 & 10.0 \\
AconityMINIPrinterLarge & 2403 & 1184 & 2 & 300 & 1201.5 \\
AconityMINIPrinterSmall & 589 & 292 & 2 & 300 & 294.5 \\
Adiac & 390 & 391 & 37 & 176 & 10.5 \\
AllGestureWiimoteX & 300 & 700 & 10 & 500 & 30.0 \\
AllGestureWiimoteY & 300 & 700 & 10 & 500 & 30.0 \\
AllGestureWiimoteZ & 300 & 700 & 10 & 500 & 30.0 \\
ArrowHead & 36 & 175 & 3 & 251 & 12.0 \\
AsphaltObstaclesUni & 390 & 391 & 4 & 736 & 97.5 \\
AsphaltPavementTypeUni & 1055 & 1056 & 3 & 2371 & 351.7 \\
AsphaltRegularityUni & 751 & 751 & 2 & 4201 & 375.5 \\
BME & 30 & 150 & 3 & 128 & 10.0 \\
Beef & 30 & 30 & 5 & 470 & 6.0 \\
BeetleFly & 20 & 20 & 2 & 512 & 10.0 \\
BirdChicken & 20 & 20 & 2 & 512 & 10.0 \\
CBF & 30 & 900 & 3 & 128 & 10.0 \\
Car & 60 & 60 & 4 & 577 & 15.0 \\
Chinatown & 20 & 343 & 2 & 24 & 10.0 \\
ChlorineConcentration & 467 & 3840 & 3 & 166 & 155.7 \\
CinCECGTorso & 40 & 1380 & 4 & 1639 & 10.0 \\
Coffee & 28 & 28 & 2 & 286 & 14.0 \\
Colposcopy & 99 & 101 & 6 & 180 & 16.5 \\
Computers & 250 & 250 & 2 & 720 & 125.0 \\
Covid3Month\_disc & 140 & 61 & 3 & 84 & 46.7 \\
CricketX & 390 & 390 & 12 & 300 & 32.5 \\
CricketY & 390 & 390 & 12 & 300 & 32.5 \\
CricketZ & 390 & 390 & 12 & 300 & 32.5 \\
Crop & 7200 & 16800 & 24 & 46 & 300.0 \\
DiatomSizeReduction & 16 & 306 & 4 & 345 & 4.0 \\
DistalPhalanxOutlineAgeGroup & 400 & 139 & 3 & 80 & 133.3 \\
DistalPhalanxOutlineCorrect & 600 & 276 & 2 & 80 & 300.0 \\
DistalPhalanxTW & 400 & 139 & 6 & 80 & 66.7 \\
DodgerLoopDay & 67 & 77 & 7 & 288 & 9.6 \\
DodgerLoopGame & 17 & 127 & 2 & 288 & 8.5 \\
DodgerLoopWeekend & 18 & 126 & 2 & 288 & 9.0 \\
ECG200 & 100 & 100 & 2 & 96 & 50.0 \\
ECG5000 & 500 & 4500 & 5 & 140 & 100.0 \\
ECGFiveDays & 23 & 861 & 2 & 136 & 11.5 \\
EOGHorizontalSignal & 362 & 362 & 12 & 1250 & 30.2 \\
EOGVerticalSignal & 362 & 362 & 12 & 1250 & 30.2 \\
Earthquakes & 322 & 139 & 2 & 512 & 161.0 \\
ElectricDeviceDetection & 624 & 3768 & 2 & 256 & 312.0 \\
ElectricDevices & 8926 & 7711 & 7 & 96 & 1275.1 \\
EthanolLevel & 504 & 500 & 4 & 1751 & 126.0 \\
FaceAll & 560 & 1690 & 14 & 131 & 40.0 \\
FaceFour & 24 & 88 & 4 & 350 & 6.0 \\
FacesUCR & 200 & 2050 & 14 & 131 & 14.3 \\
FiftyWords & 450 & 455 & 50 & 270 & 9.0 \\
Fish & 175 & 175 & 7 & 463 & 25.0 \\
FloodModeling1\_disc & 471 & 202 & 2 & 266 & 235.5 \\
FloodModeling2\_disc & 466 & 201 & 2 & 266 & 233.0 \\
FloodModeling3\_disc & 429 & 184 & 2 & 266 & 214.5 \\
FordA & 3601 & 1320 & 2 & 500 & 1800.5 \\
FordB & 3636 & 810 & 2 & 500 & 1818.0 \\
FreezerRegularTrain & 150 & 2850 & 2 & 301 & 75.0 \\
FreezerSmallTrain & 28 & 2850 & 2 & 301 & 14.0 \\
GestureMidAirD1 & 208 & 130 & 26 & 360 & 8.0 \\
GestureMidAirD2 & 208 & 130 & 26 & 360 & 8.0 \\
GestureMidAirD3 & 208 & 130 & 26 & 360 & 8.0 \\
GesturePebbleZ1 & 132 & 172 & 6 & 455 & 22.0 \\
GesturePebbleZ2 & 146 & 158 & 6 & 455 & 24.3 \\
GunPoint & 50 & 150 & 2 & 150 & 25.0 \\
GunPointAgeSpan & 135 & 316 & 2 & 150 & 67.5 \\
GunPointMaleVersusFemale & 135 & 316 & 2 & 150 & 67.5 \\
GunPointOldVersusYoung & 136 & 315 & 2 & 150 & 68.0 \\
Ham & 109 & 105 & 2 & 431 & 54.5 \\
HandOutlines & 1000 & 370 & 2 & 2709 & 500.0 \\
Haptics & 155 & 308 & 5 & 1092 & 31.0 \\
Herring & 64 & 64 & 2 & 512 & 32.0 \\
HouseTwenty & 40 & 119 & 2 & 2000 & 20.0 \\
InlineSkate & 100 & 550 & 7 & 1882 & 14.3 \\
InsectEPGRegularTrain & 62 & 249 & 3 & 601 & 20.7 \\
InsectEPGSmallTrain & 17 & 249 & 3 & 601 & 5.7 \\
InsectWingbeatSound & 220 & 1980 & 11 & 256 & 20.0 \\
ItalyPowerDemand & 67 & 1029 & 2 & 24 & 33.5 \\
KeplerLightCurves & 920 & 399 & 7 & 4767 & 131.4 \\
LargeKitchenAppliances & 375 & 375 & 3 & 720 & 125.0 \\
Lightning2 & 60 & 61 & 2 & 637 & 30.0 \\
Lightning7 & 70 & 73 & 7 & 319 & 10.0 \\
Mallat & 55 & 2345 & 8 & 1024 & 6.9 \\
Meat & 60 & 60 & 3 & 448 & 20.0 \\
MedicalImages & 381 & 760 & 10 & 99 & 38.1 \\
MelbournePedestrian & 1138 & 2319 & 10 & 24 & 113.8 \\
MiddlePhalanxOutlineAgeGroup & 400 & 154 & 3 & 80 & 133.3 \\
MiddlePhalanxOutlineCorrect & 600 & 291 & 2 & 80 & 300.0 \\
MiddlePhalanxTW & 399 & 154 & 6 & 80 & 66.5 \\
MixedShapesRegularTrain & 500 & 2425 & 5 & 1024 & 100.0 \\
MixedShapesSmallTrain & 100 & 2425 & 5 & 1024 & 20.0 \\
MoteStrain & 20 & 1252 & 2 & 84 & 10.0 \\
NonInvasiveFetalECGThorax1 & 1800 & 1965 & 42 & 750 & 42.9 \\
NonInvasiveFetalECGThorax2 & 1800 & 1965 & 42 & 750 & 42.9 \\
OSULeaf & 200 & 242 & 6 & 427 & 33.3 \\
OliveOil & 30 & 30 & 4 & 570 & 7.5 \\
PLAID & 537 & 537 & 11 & 1345 & 48.8 \\
PhalangesOutlinesCorrect & 1800 & 858 & 2 & 80 & 900.0 \\
PhoneHeartbeatSound & 424 & 182 & 5 & 3053 & 84.8 \\
Phoneme & 214 & 1896 & 39 & 1024 & 5.5 \\
PickupGestureWiimoteZ & 50 & 50 & 10 & 361 & 5.0 \\
PigAirwayPressure & 104 & 208 & 52 & 2000 & 2.0 \\
PigArtPressure & 104 & 208 & 52 & 2000 & 2.0 \\
PigCVP & 104 & 208 & 52 & 2000 & 2.0 \\
Plane & 105 & 105 & 7 & 144 & 15.0 \\
PowerCons & 180 & 180 & 2 & 144 & 90.0 \\
ProximalPhalanxOutlineAgeGroup & 400 & 205 & 3 & 80 & 133.3 \\
ProximalPhalanxOutlineCorrect & 600 & 291 & 2 & 80 & 300.0 \\
ProximalPhalanxTW & 400 & 205 & 6 & 80 & 66.7 \\
RefrigerationDevices & 375 & 375 & 3 & 720 & 125.0 \\
Rock & 20 & 50 & 4 & 2844 & 5.0 \\
ScreenType & 375 & 375 & 3 & 720 & 125.0 \\
SemgHandGenderCh2 & 300 & 600 & 2 & 1500 & 150.0 \\
SemgHandMovementCh2 & 450 & 450 & 6 & 1500 & 75.0 \\
SemgHandSubjectCh2 & 450 & 450 & 5 & 1500 & 90.0 \\
ShakeGestureWiimoteZ & 50 & 50 & 10 & 385 & 5.0 \\
ShapeletSim & 20 & 180 & 2 & 500 & 10.0 \\
ShapesAll & 600 & 600 & 60 & 512 & 10.0 \\
SharePriceIncrease & 965 & 966 & 2 & 60 & 482.5 \\
SmallKitchenAppliances & 375 & 375 & 3 & 720 & 125.0 \\
SmoothSubspace & 150 & 150 & 3 & 15 & 50.0 \\
SonyAIBORobotSurface1 & 20 & 601 & 2 & 70 & 10.0 \\
SonyAIBORobotSurface2 & 27 & 953 & 2 & 65 & 13.5 \\
StarLightCurves & 1000 & 8236 & 3 & 1024 & 333.3 \\
Strawberry & 613 & 370 & 2 & 235 & 306.5 \\
SwedishLeaf & 500 & 625 & 15 & 128 & 33.3 \\
Symbols & 25 & 995 & 6 & 398 & 4.2 \\
SyntheticControl & 300 & 300 & 6 & 60 & 50.0 \\
ToeSegmentation1 & 40 & 228 & 2 & 277 & 20.0 \\
ToeSegmentation2 & 36 & 130 & 2 & 343 & 18.0 \\
Tools & 310 & 134 & 5 & 2926 & 62.0 \\
Trace & 100 & 100 & 4 & 275 & 25.0 \\
TwoLeadECG & 23 & 1139 & 2 & 82 & 11.5 \\
TwoPatterns & 1000 & 4000 & 4 & 128 & 250.0 \\
UMD & 36 & 144 & 3 & 150 & 12.0 \\
UWaveGestureLibraryAll & 896 & 3582 & 8 & 945 & 112.0 \\
UWaveGestureLibraryX & 896 & 3582 & 8 & 315 & 112.0 \\
UWaveGestureLibraryY & 896 & 3582 & 8 & 315 & 112.0 \\
UWaveGestureLibraryZ & 896 & 3582 & 8 & 315 & 112.0 \\
Wafer & 1000 & 6164 & 2 & 152 & 500.0 \\
Wine & 57 & 54 & 2 & 234 & 28.5 \\
WordSynonyms & 267 & 638 & 25 & 270 & 10.7 \\
Worms & 181 & 77 & 5 & 900 & 36.2 \\
WormsTwoClass & 181 & 77 & 2 & 900 & 90.5 \\
Yoga & 300 & 3000 & 2 & 426 & 150.0 \\
\end{longtable}
\endgroup

\section{Six candidate meta-classification methods}\label{secB}

Table~\ref{tab:six-methods-per-dataset} reports the mean accuracy of each of the six candidate meta-classification methods of Section~\ref{sec5subsec1}, for every one of the 142 data sets individually, sorted alphabetically.
These are the per-data-set numbers underlying every aggregate comparison of these six methods on the full benchmark elsewhere in the paper, including the critical difference diagram of Figure~\ref{fig:cd-methods} and Table~\ref{tab:oracle-headroom}.

\begingroup\footnotesize
\begin{longtable}{lcccccc}
\caption{Mean accuracy (over 30 resamples) of the six candidate meta-classification methods, for each of the 142 data sets (HMV = hard majority vote, LR = logistic regression, WHMV = weighted hard majority vote).}
\label{tab:six-methods-per-dataset} \\
\toprule
Data set & HMV & LR & LightGBM & CAWPE & WHMV & TabICLv2 \\
\midrule
\endfirsthead
\multicolumn{7}{c}{{\tablename\ \thetable{} -- continued from previous page}} \\
\toprule
Data set & HMV & LR & LightGBM & CAWPE & WHMV & TabICLv2 \\
\midrule
\endhead
\midrule \multicolumn{7}{r}{{Continued on next page}} \\
\endfoot
\bottomrule
\endlastfoot
ACSF1 & 0.866 & 0.857 & 0.855 & 0.862 & 0.871 & 0.873 \\
AconityMINIPrinterLarge & 0.954 & 0.963 & 0.964 & 0.953 & 0.956 & 0.967 \\
AconityMINIPrinterSmall & 0.975 & 0.978 & 0.975 & 0.975 & 0.975 & 0.978 \\
Adiac & 0.811 & 0.828 & 0.814 & 0.823 & 0.821 & 0.852 \\
AllGestureWiimoteX & 0.745 & 0.758 & 0.755 & 0.778 & 0.771 & 0.780 \\
AllGestureWiimoteY & 0.754 & 0.780 & 0.786 & 0.798 & 0.785 & 0.804 \\
AllGestureWiimoteZ & 0.712 & 0.722 & 0.724 & 0.734 & 0.740 & 0.755 \\
ArrowHead & 0.876 & 0.866 & 0.855 & 0.871 & 0.882 & 0.868 \\
AsphaltObstaclesUni & 0.867 & 0.881 & 0.886 & 0.876 & 0.876 & 0.894 \\
AsphaltPavementTypeUni & 0.902 & 0.941 & 0.933 & 0.901 & 0.913 & 0.940 \\
AsphaltRegularityUni & 0.985 & 0.989 & 0.989 & 0.988 & 0.987 & 0.990 \\
BME & 0.995 & 0.999 & 0.999 & 0.999 & 0.999 & 0.999 \\
Beef & 0.753 & 0.703 & 0.689 & 0.758 & 0.778 & 0.702 \\
BeetleFly & 0.937 & 0.920 & 0.893 & 0.923 & 0.935 & 0.913 \\
BirdChicken & 0.933 & 0.930 & 0.918 & 0.930 & 0.933 & 0.928 \\
CBF & 0.995 & 0.996 & 0.986 & 0.995 & 0.995 & 0.996 \\
Car & 0.909 & 0.906 & 0.921 & 0.916 & 0.922 & 0.925 \\
Chinatown & 0.971 & 0.971 & 0.967 & 0.971 & 0.972 & 0.971 \\
ChlorineConcentration & 0.737 & 0.803 & 0.785 & 0.763 & 0.759 & 0.811 \\
CinCECGTorso & 0.985 & 0.983 & 0.967 & 0.991 & 0.991 & 0.987 \\
Coffee & 1.000 & 0.995 & 0.982 & 0.995 & 1.000 & 0.994 \\
Colposcopy & 0.426 & 0.419 & 0.380 & 0.417 & 0.423 & 0.396 \\
Computers & 0.827 & 0.871 & 0.868 & 0.826 & 0.839 & 0.879 \\
Covid3Month\_disc & 0.629 & 0.597 & 0.602 & 0.620 & 0.638 & 0.599 \\
CricketX & 0.801 & 0.828 & 0.826 & 0.828 & 0.824 & 0.837 \\
CricketY & 0.812 & 0.826 & 0.834 & 0.841 & 0.841 & 0.843 \\
CricketZ & 0.831 & 0.843 & 0.844 & 0.850 & 0.850 & 0.855 \\
Crop & 0.767 & 0.790 & 0.784 & 0.770 & 0.777 & 0.791 \\
DiatomSizeReduction & 0.906 & 0.885 & 0.855 & 0.919 & 0.918 & 0.897 \\
DistalPhalanxOutlineAgeGroup & 0.823 & 0.805 & 0.778 & 0.833 & 0.823 & 0.795 \\
DistalPhalanxOutlineCorrect & 0.833 & 0.842 & 0.837 & 0.836 & 0.836 & 0.848 \\
DistalPhalanxTW & 0.708 & 0.716 & 0.701 & 0.712 & 0.712 & 0.725 \\
DodgerLoopDay & 0.611 & 0.562 & 0.584 & 0.615 & 0.625 & 0.584 \\
DodgerLoopGame & 0.852 & 0.850 & 0.817 & 0.852 & 0.855 & 0.844 \\
DodgerLoopWeekend & 0.985 & 0.984 & 0.972 & 0.985 & 0.985 & 0.983 \\
ECG200 & 0.871 & 0.887 & 0.889 & 0.866 & 0.878 & 0.901 \\
ECG5000 & 0.947 & 0.946 & 0.944 & 0.948 & 0.948 & 0.948 \\
ECGFiveDays & 0.991 & 0.988 & 0.981 & 0.989 & 0.991 & 0.990 \\
EOGHorizontalSignal & 0.812 & 0.833 & 0.839 & 0.840 & 0.841 & 0.850 \\
EOGVerticalSignal & 0.763 & 0.782 & 0.775 & 0.785 & 0.787 & 0.799 \\
Earthquakes & 0.747 & 0.736 & 0.736 & 0.746 & 0.746 & 0.739 \\
ElectricDeviceDetection & 0.904 & 0.904 & 0.895 & 0.904 & 0.904 & 0.906 \\
ElectricDevices & 0.888 & 0.917 & 0.916 & 0.906 & 0.901 & 0.921 \\
EthanolLevel & 0.744 & 0.875 & 0.875 & 0.798 & 0.792 & 0.898 \\
FaceAll & 0.979 & 0.977 & 0.973 & 0.981 & 0.981 & 0.978 \\
FaceFour & 0.920 & 0.935 & 0.884 & 0.939 & 0.947 & 0.956 \\
FacesUCR & 0.950 & 0.959 & 0.949 & 0.965 & 0.965 & 0.963 \\
FiftyWords & 0.753 & 0.777 & 0.744 & 0.792 & 0.788 & 0.782 \\
Fish & 0.972 & 0.974 & 0.979 & 0.978 & 0.980 & 0.975 \\
FloodModeling1\_disc & 0.908 & 0.950 & 0.949 & 0.927 & 0.933 & 0.951 \\
FloodModeling2\_disc & 0.956 & 0.979 & 0.983 & 0.954 & 0.962 & 0.984 \\
FloodModeling3\_disc & 0.868 & 0.936 & 0.937 & 0.909 & 0.905 & 0.945 \\
FordA & 0.960 & 0.981 & 0.981 & 0.963 & 0.961 & 0.982 \\
FordB & 0.932 & 0.942 & 0.939 & 0.934 & 0.934 & 0.942 \\
FreezerRegularTrain & 0.999 & 0.998 & 0.999 & 0.999 & 0.999 & 1.000 \\
FreezerSmallTrain & 0.966 & 0.977 & 0.990 & 0.970 & 0.983 & 0.994 \\
GestureMidAirD1 & 0.756 & 0.722 & 0.716 & 0.756 & 0.757 & 0.764 \\
GestureMidAirD2 & 0.634 & 0.634 & 0.604 & 0.624 & 0.641 & 0.635 \\
GestureMidAirD3 & 0.510 & 0.488 & 0.470 & 0.527 & 0.536 & 0.530 \\
GesturePebbleZ1 & 0.959 & 0.959 & 0.961 & 0.957 & 0.962 & 0.971 \\
GesturePebbleZ2 & 0.963 & 0.963 & 0.966 & 0.968 & 0.970 & 0.974 \\
GunPoint & 0.992 & 0.985 & 0.996 & 0.990 & 0.995 & 0.993 \\
GunPointAgeSpan & 0.993 & 0.994 & 0.994 & 0.993 & 0.994 & 0.994 \\
GunPointMaleVersusFemale & 1.000 & 0.998 & 1.000 & 0.999 & 1.000 & 0.998 \\
GunPointOldVersusYoung & 1.000 & 0.998 & 1.000 & 1.000 & 1.000 & 1.000 \\
Ham & 0.825 & 0.855 & 0.850 & 0.838 & 0.842 & 0.856 \\
HandOutlines & 0.934 & 0.952 & 0.954 & 0.942 & 0.940 & 0.960 \\
Haptics & 0.543 & 0.529 & 0.536 & 0.538 & 0.555 & 0.559 \\
Herring & 0.631 & 0.601 & 0.596 & 0.633 & 0.635 & 0.594 \\
HouseTwenty & 0.984 & 0.983 & 0.968 & 0.986 & 0.985 & 0.983 \\
InlineSkate & 0.511 & 0.542 & 0.531 & 0.544 & 0.568 & 0.598 \\
InsectEPGRegularTrain & 1.000 & 0.998 & 1.000 & 1.000 & 1.000 & 1.000 \\
InsectEPGSmallTrain & 1.000 & 0.990 & 0.989 & 1.000 & 1.000 & 1.000 \\
InsectWingbeatSound & 0.676 & 0.653 & 0.632 & 0.676 & 0.677 & 0.654 \\
ItalyPowerDemand & 0.961 & 0.957 & 0.949 & 0.961 & 0.961 & 0.959 \\
KeplerLightCurves & 0.943 & 0.970 & 0.970 & 0.950 & 0.951 & 0.975 \\
LargeKitchenAppliances & 0.921 & 0.940 & 0.946 & 0.940 & 0.936 & 0.948 \\
Lightning2 & 0.750 & 0.774 & 0.728 & 0.755 & 0.766 & 0.766 \\
Lightning7 & 0.768 & 0.778 & 0.779 & 0.783 & 0.786 & 0.790 \\
Mallat & 0.965 & 0.987 & 0.978 & 0.971 & 0.971 & 0.992 \\
Meat & 0.983 & 0.976 & 0.977 & 0.977 & 0.983 & 0.978 \\
MedicalImages & 0.781 & 0.824 & 0.812 & 0.812 & 0.805 & 0.839 \\
MelbournePedestrian & 0.961 & 0.971 & 0.972 & 0.963 & 0.966 & 0.974 \\
MiddlePhalanxOutlineAgeGroup & 0.710 & 0.667 & 0.625 & 0.681 & 0.710 & 0.696 \\
MiddlePhalanxOutlineCorrect & 0.825 & 0.850 & 0.848 & 0.835 & 0.832 & 0.854 \\
MiddlePhalanxTW & 0.579 & 0.577 & 0.549 & 0.567 & 0.579 & 0.591 \\
MixedShapesRegularTrain & 0.974 & 0.981 & 0.981 & 0.975 & 0.976 & 0.983 \\
MixedShapesSmallTrain & 0.957 & 0.956 & 0.956 & 0.957 & 0.960 & 0.955 \\
MoteStrain & 0.939 & 0.939 & 0.915 & 0.942 & 0.942 & 0.932 \\
NonInvasiveFetalECGThorax1 & 0.942 & 0.964 & 0.961 & 0.951 & 0.948 & 0.968 \\
NonInvasiveFetalECGThorax2 & 0.955 & 0.969 & 0.967 & 0.959 & 0.960 & 0.973 \\
OSULeaf & 0.969 & 0.972 & 0.965 & 0.973 & 0.976 & 0.973 \\
OliveOil & 0.901 & 0.902 & 0.874 & 0.907 & 0.900 & 0.904 \\
PLAID & 0.926 & 0.948 & 0.946 & 0.947 & 0.938 & 0.947 \\
PhalangesOutlinesCorrect & 0.835 & 0.868 & 0.862 & 0.844 & 0.845 & 0.869 \\
PhoneHeartbeatSound & 0.652 & 0.713 & 0.709 & 0.653 & 0.664 & 0.730 \\
Phoneme & 0.368 & 0.321 & 0.314 & 0.377 & 0.397 & 0.356 \\
PickupGestureWiimoteZ & 0.851 & 0.821 & 0.808 & 0.855 & 0.859 & 0.827 \\
PigAirwayPressure & 0.809 & 0.462 & 0.542 & 0.741 & 0.865 & 0.602 \\
PigArtPressure & 0.966 & 0.919 & 0.887 & 0.937 & 0.963 & 0.892 \\
PigCVP & 0.919 & 0.836 & 0.810 & 0.905 & 0.928 & 0.823 \\
Plane & 1.000 & 1.000 & 1.000 & 1.000 & 1.000 & 0.998 \\
PowerCons & 0.975 & 0.988 & 0.990 & 0.974 & 0.978 & 0.993 \\
ProximalPhalanxOutlineAgeGroup & 0.859 & 0.851 & 0.838 & 0.847 & 0.859 & 0.846 \\
ProximalPhalanxOutlineCorrect & 0.896 & 0.906 & 0.904 & 0.903 & 0.898 & 0.906 \\
ProximalPhalanxTW & 0.818 & 0.808 & 0.795 & 0.810 & 0.818 & 0.808 \\
RefrigerationDevices & 0.742 & 0.802 & 0.791 & 0.761 & 0.771 & 0.805 \\
Rock & 0.880 & 0.855 & 0.851 & 0.884 & 0.883 & 0.839 \\
ScreenType & 0.652 & 0.715 & 0.708 & 0.697 & 0.704 & 0.726 \\
SemgHandGenderCh2 & 0.968 & 0.971 & 0.969 & 0.970 & 0.969 & 0.972 \\
SemgHandMovementCh2 & 0.785 & 0.897 & 0.888 & 0.864 & 0.863 & 0.912 \\
SemgHandSubjectCh2 & 0.920 & 0.968 & 0.958 & 0.942 & 0.938 & 0.971 \\
ShakeGestureWiimoteZ & 0.938 & 0.921 & 0.895 & 0.944 & 0.943 & 0.919 \\
ShapeletSim & 1.000 & 1.000 & 1.000 & 1.000 & 1.000 & 1.000 \\
ShapesAll & 0.945 & 0.936 & 0.916 & 0.943 & 0.948 & 0.938 \\
SharePriceIncrease & 0.688 & 0.686 & 0.682 & 0.678 & 0.688 & 0.687 \\
SmallKitchenAppliances & 0.826 & 0.844 & 0.840 & 0.821 & 0.833 & 0.850 \\
SmoothSubspace & 0.994 & 0.994 & 0.995 & 0.996 & 0.995 & 0.997 \\
SonyAIBORobotSurface1 & 0.940 & 0.949 & 0.884 & 0.943 & 0.946 & 0.954 \\
SonyAIBORobotSurface2 & 0.940 & 0.941 & 0.939 & 0.936 & 0.942 & 0.945 \\
StarLightCurves & 0.982 & 0.982 & 0.982 & 0.982 & 0.982 & 0.983 \\
Strawberry & 0.976 & 0.978 & 0.977 & 0.975 & 0.976 & 0.980 \\
SwedishLeaf & 0.964 & 0.968 & 0.967 & 0.969 & 0.966 & 0.970 \\
Symbols & 0.974 & 0.960 & 0.935 & 0.972 & 0.974 & 0.964 \\
SyntheticControl & 0.996 & 0.996 & 0.989 & 0.996 & 0.996 & 0.996 \\
ToeSegmentation1 & 0.948 & 0.945 & 0.936 & 0.947 & 0.952 & 0.946 \\
ToeSegmentation2 & 0.956 & 0.947 & 0.942 & 0.952 & 0.957 & 0.946 \\
Tools & 0.862 & 0.895 & 0.909 & 0.876 & 0.877 & 0.914 \\
Trace & 1.000 & 0.999 & 1.000 & 1.000 & 1.000 & 1.000 \\
TwoLeadECG & 0.996 & 0.995 & 0.995 & 0.996 & 0.998 & 0.996 \\
TwoPatterns & 1.000 & 1.000 & 1.000 & 1.000 & 1.000 & 1.000 \\
UMD & 0.987 & 0.989 & 0.977 & 0.988 & 0.988 & 0.990 \\
UWaveGestureLibraryAll & 0.976 & 0.981 & 0.982 & 0.977 & 0.978 & 0.984 \\
UWaveGestureLibraryX & 0.860 & 0.873 & 0.874 & 0.859 & 0.866 & 0.884 \\
UWaveGestureLibraryY & 0.790 & 0.813 & 0.812 & 0.794 & 0.802 & 0.822 \\
UWaveGestureLibraryZ & 0.803 & 0.820 & 0.820 & 0.808 & 0.812 & 0.834 \\
Wafer & 0.999 & 0.999 & 1.000 & 1.000 & 1.000 & 1.000 \\
Wine & 0.928 & 0.938 & 0.915 & 0.936 & 0.939 & 0.935 \\
WordSynonyms & 0.716 & 0.730 & 0.696 & 0.768 & 0.765 & 0.722 \\
Worms & 0.761 & 0.741 & 0.751 & 0.749 & 0.768 & 0.761 \\
WormsTwoClass & 0.784 & 0.784 & 0.776 & 0.785 & 0.784 & 0.786 \\
Yoga & 0.913 & 0.947 & 0.947 & 0.935 & 0.931 & 0.950 \\
\end{longtable}
\endgroup

\section{TIGER's AUROC and NLL}\label{secC}

As noted in Section~\ref{sec5subsec2}, TIGER's meta-classification stage can defer to weighted hard majority vote, a hard-voting rule with no native probabilistic output.
Table~\ref{tab:multi-metric-142}'s AUROC and NLL for TIGER therefore follow the same convention used elsewhere in the literature for classifiers with no native probabilistic output, such as the Ridge classifiers underlying RDST and WEASEL~2.0.
For these, a one-hot probability vector (1 for the predicted class, 0 for every other class) is used whenever AUROC or NLL must be computed.
Whenever TIGER's adaptive rule selects weighted hard majority vote, we likewise assign a probability of 1 to its predicted class and 0 to every other class.
Whenever it selects TabICLv2, we use TabICLv2's own predicted class probabilities directly.

This convention is not very informative on the 85 data sets on which the adaptive rule always selects weighted hard majority vote, since a one-hot probability vector is, by construction, either perfectly confident and correct (zero loss) or perfectly confident and wrong (unbounded loss), and does not reflect any genuine predictive uncertainty.
We therefore restrict this comparison to the 57 data sets on which the adaptive rule always selects TabICLv2 instead, where TIGER's reported probabilities are genuine throughout and AUROC and NLL are consequently meaningful.
Table~\ref{tab:multi-metric-tabiclv2-subset} repeats the five-metric comparison of Section~\ref{sec5subsec2}, restricted to these 57 data sets.
TIGER obtains the best mean value, and the best mean rank, on every one of the five metrics, including AUROC (0.955) and NLL (0.265), both ahead of HC2 (0.950 and 0.350, respectively), the strongest competitor on these two metrics in the main, full-benchmark comparison (Table~\ref{tab:multi-metric-142}).

\begin{table}[htbp]
\caption{
    Summary performance measures on the 57 data sets where TIGER's adaptive rule always selects TabICLv2.
}
\label{tab:multi-metric-tabiclv2-subset}
\centering
\begin{tabular}{lccccc}
\toprule
Algorithm & ACC & BALACC & AUROC & NLL & F1 \\
\midrule
TIGER & \textbf{0.895 (1)} & \textbf{0.862 (1)} & \textbf{0.955 (1)} & \textbf{0.265 (1)} & \textbf{0.860 (1)} \\
KG-MTP & 0.879 (2) & 0.839 (2) & 0.889 (4) & 4.374 (4) & 0.842 (2) \\
HC2 & 0.878 (3) & 0.825 (4) & 0.950 (2) & 0.350 (2) & -- \\
QUANT & 0.874 (4) & 0.831 (3) & 0.948 (3) & 0.402 (3) & 0.832 (3) \\
RDST & 0.863 (5) & 0.813 (5) & 0.867 (5) & 4.952 (5) & 0.807 (4) \\
WEASEL~2.0 & 0.855 (6) & 0.801 (6) & 0.858 (6) & 5.240 (6) & 0.788 (5) \\
\bottomrule
\end{tabular}
\end{table}

\section{Per-data-set results of TIGER}\label{secD}

Table~\ref{tab:tiger-per-dataset} reports TIGER's own mean ACC, BALACC, AUROC, NLL, and F1, each averaged over the 30 resamples, for every one of the 142 data sets individually, sorted alphabetically.
These are the per-data-set numbers underlying the pooled 142-data-set results of Figure~\ref{fig:cd-literature}, Table~\ref{tab:multi-metric-142}, and Table~\ref{tab:multi-metric-122-heldout}.

\begingroup\footnotesize
\begin{longtable}{lccccc}
\caption{TIGER's mean ACC, BALACC, AUROC, NLL, and F1 (over 30 resamples), for each of the 142 data sets.}
\label{tab:tiger-per-dataset} \\
\toprule
Data set & ACC & BALACC & AUROC & NLL & F1 \\
\midrule
\endfirsthead
\multicolumn{6}{c}{{\tablename\ \thetable{} -- continued from previous page}} \\
\toprule
Data set & ACC & BALACC & AUROC & NLL & F1 \\
\midrule
\endhead
\midrule \multicolumn{6}{r}{{Continued on next page}} \\
\endfoot
\bottomrule
\endlastfoot
ACSF1 & 0.871 & 0.871 & 0.928 & 4.662 & 0.868 \\
AconityMINIPrinterLarge & 0.967 & 0.964 & 0.988 & 0.115 & 0.951 \\
AconityMINIPrinterSmall & 0.978 & 0.966 & 0.982 & 0.101 & 0.960 \\
Adiac & 0.821 & 0.825 & 0.908 & 6.468 & 0.813 \\
AllGestureWiimoteX & 0.771 & 0.771 & 0.873 & 8.266 & 0.768 \\
AllGestureWiimoteY & 0.785 & 0.785 & 0.881 & 7.734 & 0.783 \\
AllGestureWiimoteZ & 0.740 & 0.740 & 0.855 & 9.376 & 0.735 \\
ArrowHead & 0.882 & 0.877 & 0.910 & 4.250 & 0.881 \\
AsphaltObstaclesUni & 0.894 & 0.893 & 0.979 & 0.280 & 0.893 \\
AsphaltPavementTypeUni & 0.940 & 0.933 & 0.992 & 0.144 & 0.940 \\
AsphaltRegularityUni & 0.990 & 0.990 & 1.000 & 0.029 & 0.990 \\
BME & 0.999 & 0.999 & 0.999 & 0.048 & 0.999 \\
Beef & 0.778 & 0.778 & 0.861 & 8.010 & 0.774 \\
BeetleFly & 0.935 & 0.935 & 0.935 & 2.343 & 0.939 \\
BirdChicken & 0.933 & 0.933 & 0.933 & 2.403 & 0.927 \\
CBF & 0.995 & 0.995 & 0.997 & 0.167 & 0.995 \\
Car & 0.922 & 0.930 & 0.949 & 2.803 & 0.923 \\
Chinatown & 0.972 & 0.980 & 0.980 & 1.002 & 0.952 \\
ChlorineConcentration & 0.811 & 0.772 & 0.923 & 0.503 & 0.806 \\
CinCECGTorso & 0.991 & 0.991 & 0.994 & 0.333 & 0.991 \\
Coffee & 1.000 & 1.000 & 1.000 & 0.000 & 1.000 \\
Colposcopy & 0.423 & 0.283 & 0.616 & 20.794 & 0.319 \\
Computers & 0.879 & 0.879 & 0.946 & 0.296 & 0.876 \\
Covid3Month\_disc & 0.638 & 0.550 & 0.687 & 13.039 & 0.592 \\
CricketX & 0.824 & 0.834 & 0.904 & 6.343 & 0.820 \\
CricketY & 0.841 & 0.843 & 0.913 & 5.727 & 0.839 \\
CricketZ & 0.850 & 0.844 & 0.918 & 5.410 & 0.849 \\
Crop & 0.791 & 0.791 & 0.987 & 0.602 & 0.791 \\
DiatomSizeReduction & 0.918 & 0.823 & 0.941 & 2.949 & 0.899 \\
DistalPhalanxOutlineAgeGroup & 0.795 & 0.779 & 0.915 & 0.458 & 0.793 \\
DistalPhalanxOutlineCorrect & 0.848 & 0.843 & 0.929 & 0.339 & 0.816 \\
DistalPhalanxTW & 0.725 & 0.566 & 0.934 & 0.670 & 0.693 \\
DodgerLoopDay & 0.625 & 0.613 & 0.781 & 13.512 & 0.609 \\
DodgerLoopGame & 0.855 & 0.854 & 0.854 & 5.213 & 0.841 \\
DodgerLoopWeekend & 0.985 & 0.990 & 0.990 & 0.524 & 0.974 \\
ECG200 & 0.878 & 0.859 & 0.859 & 4.397 & 0.823 \\
ECG5000 & 0.948 & 0.580 & 0.989 & 0.182 & 0.941 \\
ECGFiveDays & 0.991 & 0.991 & 0.991 & 0.332 & 0.991 \\
EOGHorizontalSignal & 0.841 & 0.841 & 0.913 & 5.725 & 0.838 \\
EOGVerticalSignal & 0.787 & 0.787 & 0.884 & 7.680 & 0.781 \\
Earthquakes & 0.739 & 0.532 & 0.660 & 0.621 & 0.175 \\
ElectricDeviceDetection & 0.906 & 0.748 & 0.914 & 0.251 & 0.609 \\
ElectricDevices & 0.921 & 0.899 & 0.994 & 0.244 & 0.921 \\
EthanolLevel & 0.898 & 0.898 & 0.984 & 0.246 & 0.898 \\
FaceAll & 0.981 & 0.985 & 0.990 & 0.676 & 0.981 \\
FaceFour & 0.947 & 0.953 & 0.964 & 1.925 & 0.946 \\
FacesUCR & 0.965 & 0.941 & 0.981 & 1.274 & 0.964 \\
FiftyWords & 0.788 & 0.668 & 0.888 & 7.652 & 0.759 \\
Fish & 0.980 & 0.980 & 0.988 & 0.714 & 0.980 \\
FloodModeling1\_disc & 0.951 & 0.937 & 0.990 & 0.118 & 0.903 \\
FloodModeling2\_disc & 0.984 & 0.953 & 0.998 & 0.036 & 0.896 \\
FloodModeling3\_disc & 0.945 & 0.935 & 0.988 & 0.137 & 0.913 \\
FordA & 0.982 & 0.982 & 0.999 & 0.048 & 0.982 \\
FordB & 0.942 & 0.942 & 0.986 & 0.147 & 0.942 \\
FreezerRegularTrain & 1.000 & 1.000 & 1.000 & 0.002 & 1.000 \\
FreezerSmallTrain & 0.983 & 0.983 & 0.983 & 0.629 & 0.983 \\
GestureMidAirD1 & 0.757 & 0.757 & 0.874 & 8.743 & 0.746 \\
GestureMidAirD2 & 0.641 & 0.641 & 0.813 & 12.948 & 0.628 \\
GestureMidAirD3 & 0.536 & 0.536 & 0.759 & 16.719 & 0.513 \\
GesturePebbleZ1 & 0.962 & 0.961 & 0.977 & 1.369 & 0.962 \\
GesturePebbleZ2 & 0.970 & 0.970 & 0.982 & 1.095 & 0.970 \\
GunPoint & 0.995 & 0.995 & 0.995 & 0.184 & 0.995 \\
GunPointAgeSpan & 0.994 & 0.994 & 1.000 & 0.014 & 0.994 \\
GunPointMaleVersusFemale & 0.998 & 0.998 & 1.000 & 0.004 & 0.998 \\
GunPointOldVersusYoung & 1.000 & 1.000 & 1.000 & 0.001 & 1.000 \\
Ham & 0.842 & 0.842 & 0.842 & 5.687 & 0.836 \\
HandOutlines & 0.960 & 0.951 & 0.982 & 0.136 & 0.942 \\
Haptics & 0.555 & 0.554 & 0.721 & 16.036 & 0.532 \\
Herring & 0.635 & 0.595 & 0.595 & 13.160 & 0.451 \\
HouseTwenty & 0.985 & 0.983 & 0.983 & 0.535 & 0.982 \\
InlineSkate & 0.568 & 0.586 & 0.745 & 15.575 & 0.566 \\
InsectEPGRegularTrain & 1.000 & 1.000 & 1.000 & 0.000 & 1.000 \\
InsectEPGSmallTrain & 1.000 & 1.000 & 1.000 & 0.000 & 1.000 \\
InsectWingbeatSound & 0.677 & 0.677 & 0.822 & 11.636 & 0.669 \\
ItalyPowerDemand & 0.961 & 0.961 & 0.961 & 1.394 & 0.961 \\
KeplerLightCurves & 0.975 & 0.973 & 0.999 & 0.055 & 0.975 \\
LargeKitchenAppliances & 0.948 & 0.948 & 0.988 & 0.184 & 0.948 \\
Lightning2 & 0.766 & 0.752 & 0.752 & 8.430 & 0.691 \\
Lightning7 & 0.786 & 0.777 & 0.866 & 7.702 & 0.779 \\
Mallat & 0.971 & 0.971 & 0.983 & 1.043 & 0.970 \\
Meat & 0.983 & 0.983 & 0.988 & 0.601 & 0.983 \\
MedicalImages & 0.805 & 0.696 & 0.833 & 7.022 & 0.788 \\
MelbournePedestrian & 0.974 & 0.973 & 0.999 & 0.077 & 0.974 \\
MiddlePhalanxOutlineAgeGroup & 0.696 & 0.649 & 0.772 & 0.777 & 0.679 \\
MiddlePhalanxOutlineCorrect & 0.854 & 0.846 & 0.939 & 0.324 & 0.822 \\
MiddlePhalanxTW & 0.591 & 0.405 & 0.837 & 1.007 & 0.539 \\
MixedShapesRegularTrain & 0.983 & 0.983 & 1.000 & 0.053 & 0.983 \\
MixedShapesSmallTrain & 0.960 & 0.959 & 0.974 & 1.451 & 0.960 \\
MoteStrain & 0.942 & 0.943 & 0.943 & 2.104 & 0.938 \\
NonInvasiveFetalECGThorax1 & 0.948 & 0.946 & 0.973 & 1.884 & 0.947 \\
NonInvasiveFetalECGThorax2 & 0.960 & 0.958 & 0.980 & 1.427 & 0.960 \\
OSULeaf & 0.976 & 0.974 & 0.985 & 0.879 & 0.976 \\
OliveOil & 0.900 & 0.879 & 0.929 & 3.604 & 0.896 \\
PLAID & 0.938 & 0.882 & 0.965 & 2.251 & 0.935 \\
PhalangesOutlinesCorrect & 0.869 & 0.857 & 0.938 & 0.311 & 0.826 \\
PhoneHeartbeatSound & 0.730 & 0.571 & 0.823 & 0.753 & 0.687 \\
Phoneme & 0.397 & 0.218 & 0.678 & 21.751 & 0.354 \\
PickupGestureWiimoteZ & 0.859 & 0.859 & 0.921 & 5.094 & 0.847 \\
PigAirwayPressure & 0.865 & 0.865 & 0.931 & 4.852 & 0.859 \\
PigArtPressure & 0.963 & 0.963 & 0.981 & 1.323 & 0.960 \\
PigCVP & 0.928 & 0.928 & 0.963 & 2.599 & 0.924 \\
Plane & 1.000 & 1.000 & 1.000 & 0.000 & 1.000 \\
PowerCons & 0.993 & 0.993 & 1.000 & 0.016 & 0.993 \\
ProximalPhalanxOutlineAgeGroup & 0.846 & 0.816 & 0.944 & 0.360 & 0.849 \\
ProximalPhalanxOutlineCorrect & 0.906 & 0.879 & 0.956 & 0.237 & 0.844 \\
ProximalPhalanxTW & 0.808 & 0.648 & 0.963 & 0.471 & 0.792 \\
RefrigerationDevices & 0.805 & 0.805 & 0.935 & 0.490 & 0.803 \\
Rock & 0.883 & 0.901 & 0.924 & 4.229 & 0.883 \\
ScreenType & 0.726 & 0.726 & 0.890 & 0.643 & 0.726 \\
SemgHandGenderCh2 & 0.972 & 0.972 & 0.997 & 0.075 & 0.961 \\
SemgHandMovementCh2 & 0.912 & 0.912 & 0.992 & 0.253 & 0.912 \\
SemgHandSubjectCh2 & 0.971 & 0.971 & 0.999 & 0.081 & 0.971 \\
ShakeGestureWiimoteZ & 0.943 & 0.943 & 0.969 & 2.042 & 0.942 \\
ShapeletSim & 1.000 & 1.000 & 1.000 & 0.000 & 1.000 \\
ShapesAll & 0.948 & 0.948 & 0.974 & 1.876 & 0.947 \\
SharePriceIncrease & 0.687 & 0.533 & 0.647 & 0.596 & 0.191 \\
SmallKitchenAppliances & 0.850 & 0.850 & 0.959 & 0.396 & 0.851 \\
SmoothSubspace & 0.995 & 0.995 & 0.996 & 0.192 & 0.995 \\
SonyAIBORobotSurface1 & 0.946 & 0.952 & 0.952 & 1.931 & 0.942 \\
SonyAIBORobotSurface2 & 0.942 & 0.942 & 0.942 & 2.083 & 0.927 \\
StarLightCurves & 0.983 & 0.967 & 0.997 & 0.059 & 0.983 \\
Strawberry & 0.980 & 0.980 & 0.996 & 0.065 & 0.972 \\
SwedishLeaf & 0.966 & 0.967 & 0.982 & 1.232 & 0.966 \\
Symbols & 0.974 & 0.975 & 0.984 & 0.943 & 0.974 \\
SyntheticControl & 0.996 & 0.996 & 0.998 & 0.128 & 0.996 \\
ToeSegmentation1 & 0.952 & 0.952 & 0.952 & 1.739 & 0.949 \\
ToeSegmentation2 & 0.957 & 0.962 & 0.962 & 1.543 & 0.894 \\
Tools & 0.914 & 0.901 & 0.991 & 0.266 & 0.914 \\
Trace & 1.000 & 1.000 & 1.000 & 0.000 & 1.000 \\
TwoLeadECG & 0.998 & 0.998 & 0.998 & 0.088 & 0.998 \\
TwoPatterns & 1.000 & 1.000 & 1.000 & 0.000 & 1.000 \\
UMD & 0.988 & 0.988 & 0.991 & 0.442 & 0.988 \\
UWaveGestureLibraryAll & 0.984 & 0.983 & 1.000 & 0.060 & 0.984 \\
UWaveGestureLibraryX & 0.884 & 0.883 & 0.988 & 0.300 & 0.884 \\
UWaveGestureLibraryY & 0.822 & 0.822 & 0.979 & 0.494 & 0.821 \\
UWaveGestureLibraryZ & 0.834 & 0.835 & 0.982 & 0.415 & 0.833 \\
Wafer & 1.000 & 1.000 & 1.000 & 0.000 & 0.999 \\
Wine & 0.939 & 0.939 & 0.939 & 2.203 & 0.939 \\
WordSynonyms & 0.765 & 0.612 & 0.868 & 8.463 & 0.740 \\
Worms & 0.768 & 0.724 & 0.830 & 8.363 & 0.764 \\
WormsTwoClass & 0.786 & 0.772 & 0.886 & 0.420 & 0.728 \\
Yoga & 0.950 & 0.950 & 0.988 & 0.137 & 0.946 \\
\end{longtable}
\endgroup

\section{Calibration of the Platt-scaled Ridge meta-features}\label{secE}

TIGER's CAWPE meta-classification method needs predicted probabilities.
However, Ridge outputs unbounded signed distances to the learned hyperplanes.
Thus, we need to convert these unbounded scores to probabilities, so that each family's Ridge base learner can be fused with the other base learners' own genuine probabilities.
To do so, we use Platt's calibration procedure \citep{platt1999probabilistic}.
We assess the quality of this Platt (sigmoid) calibration directly: for each of the four families, we calibrate the Ridge block's out-of-fold scores against the corresponding held-out test scores, exactly as CAWPE does at meta-classification time, and pool the resulting (confidence, correctness) pairs across every test row of the full 142-data-set benchmark into a reliability diagram and an Expected Calibration Error (ECE), using 15 equal-width confidence bins.

\begin{figure}[tbp]
    \centering
    \includegraphics[width=0.7\textwidth]{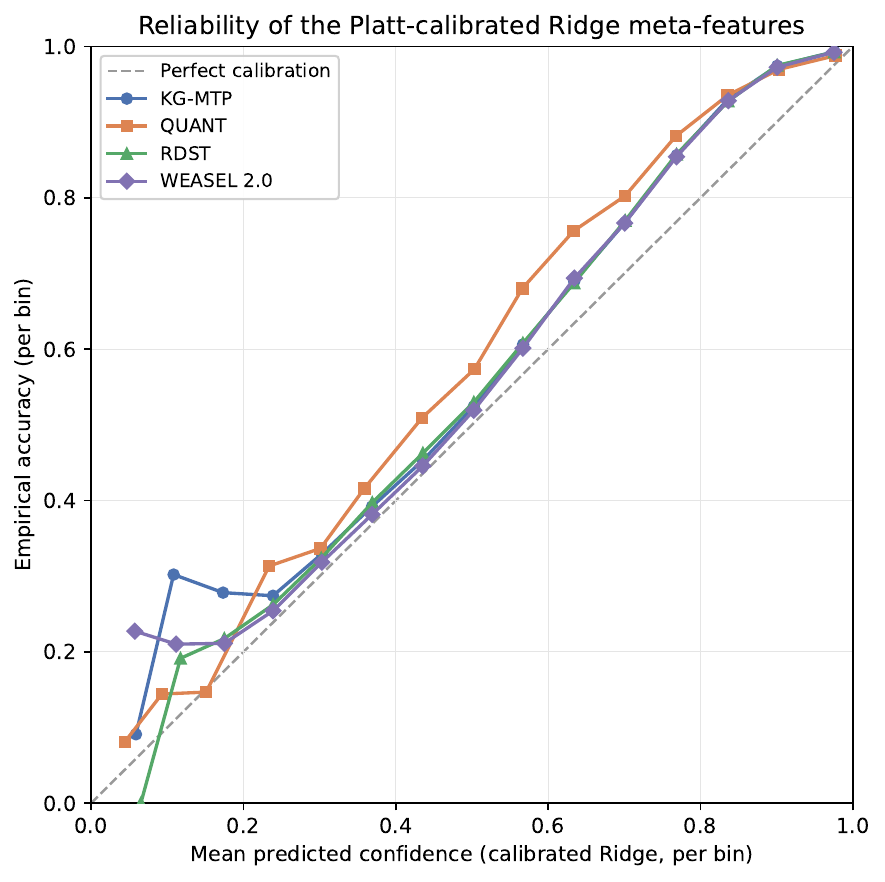}
    \caption{
        Reliability of the Platt-calibrated Ridge meta-features, per representation family.
        The dashed diagonal marks perfect calibration.
    }
    \label{fig:ridge-calibration}
\end{figure}

Figure~\ref{fig:ridge-calibration} shows all four families' reliability curves close to the diagonal across most of the confidence range, and Table~\ref{tab:ridge-calibration-ece} reports the resulting ECE values.
All four families are well calibrated and similar to one another: ECE ranges from 0.044 (WEASEL~2.0) to 0.058 (QUANT), with KG-MTP and RDST in between at 0.044 and 0.045 respectively.
QUANT's calibration, while still reasonably good, is the least reliable of the four, consistent with it also being the weakest in standalone predictive strength (Table~\ref{tab:family-standalone-strength}).

\begin{table}[tbp]
\caption{
    Expected calibration error of the Platt-scaled Ridge meta-features, per representation family (lowest in bold).
}
\label{tab:ridge-calibration-ece}
\centering
\begin{tabular}{lc}
\toprule
Algorithm & Expected Calibration Error \\
\midrule
WEASEL~2.0 & \textbf{0.0440} \\
KG-MTP & 0.0443 \\
RDST & 0.0453 \\
QUANT & 0.0583 \\
\bottomrule
\end{tabular}
\end{table}

\end{appendices}

\end{document}